\documentclass[10pt,journal,compsoc]{IEEEtran}
\usepackage{soul,framed}
\usepackage[utf8]{inputenc}
\usepackage{color}
\usepackage{graphicx,multirow}
\usepackage{amsthm,amsmath,amssymb,bm}
\usepackage[ruled,vlined]{algorithm2e}
\usepackage[dvipsnames]{xcolor}
\usepackage{colortbl}
\usepackage{dirtytalk}
\usepackage{booktabs}
\usepackage{makecell}
\usepackage[usestackEOL]{stackengine}
\usepackage[noadjust]{cite}
\usepackage{stfloats}
\usepackage{caption}
\usepackage{subcaption}
\usepackage[flushleft]{threeparttable}
\usepackage{arydshln}

\newtheorem{prop}{Proposition}
\newtheorem{coro}[]{Corollary}
\newtheorem{lem}[]{Lemma}

\newcommand{\Stress}[1]{\textbf{\textit{#1}}}

\newcommand{\imgStubPDFpage}[4]{\begin{minipage}{#1\textwidth}\begin{center}
 \includegraphics[#2]{#3}\\
#4\end{center}\end{minipage}}
\newcommand{\thickhline}{%
 \noalign {\ifnum 0=`}\fi \hrule height 1pt
 \futurelet \reserved@a \@xhline
}
\newcolumntype{"}{@{\hskip\tabcolsep\vrule width 1pt\hskip\tabcolsep}}

\DeclareMathOperator{\Prob}{\mathbb{P}}

\usepackage[colorlinks=true, linkcolor=OrangeRed, urlcolor=OrangeRed, citecolor=OrangeRed]{hyperref}

\begin{document}
\bstctlcite{IEEEexample:BSTcontrol}

\title{On the Risk of Cancelable Biometrics}

\author{Xingbo~Dong,~\IEEEmembership{Member, IEEE}, Jaewoo Park, Zhe Jin*,~\IEEEmembership{Member, IEEE}, Andrew Beng Jin Teoh,~\IEEEmembership{Senior Member, IEEE}, Massimo Tistarelli,~\IEEEmembership{Senior Member, IEEE}, and KokSheik Wong,~\IEEEmembership{Senior Member, IEEE}
\IEEEcompsocitemizethanks{\IEEEcompsocthanksitem This work was partially done when X. Dong and Z. Jin were affiliated to Monash University. * Corresponding contact: jinzhe@ahu.edu.cn
\IEEEcompsocthanksitem X. Dong and Z. Jin is with School of Artifical Intelligence, Anhui University, 230093, China (e-mail: dong.xingbo@ieee.org,jinzhe@ahu.edu.cn).
\IEEEcompsocthanksitem J. Park and B. J. Teoh are with School of Electrical and Electronic Engineering, Yonsei University, Seoul, South Korea (e-mail: \{julypraise,bjteoh\}@yonsei.ac.kr). 
\IEEEcompsocthanksitem K. Wong is with School of Information Technology, Monash University Malaysia Campus, 47500, Malaysia (e-mail: wong.koksheik@monash.edu).
\IEEEcompsocthanksitem M. Tistarelli is with Computer Vision Laboratory, University of Sassari, Alghero, SS 07041, Italy (e-mail: tista@uniss.it).}
}

\markboth{Journal of \LaTeX\ Class Files,~Vol.~14, No.~8, August~2015}%
{Shell \MakeLowercase{\textit{et al.}}: Bare Demo of IEEEtran.cls for Biometrics Council Journals}






\maketitle

\begin{abstract}
Cancelable biometrics (CB) employs an irreversible transformation to convert the biometric features into transformed templates while preserving the relative distance between two templates for security and privacy protection. However, distance preservation invites unexpected security issues such as pre-image attacks, which are often neglected.
This paper presents a generalized pre-image attack method and its extension version that operates on practical CB systems. We theoretically reveal that distance preservation property is a vulnerability source in the CB schemes. We then propose an empirical information leakage estimation algorithm to access the pre-image attack risk of the CB schemes.
The experiments conducted with six CB schemes designed for the face, iris and fingerprint, demonstrate that the risks originating from the distance computed from two transformed templates significantly compromise the security of CB schemes. Our work reveals the potential risk of existing CB systems theoretically and experimentally.
\end{abstract}



\section{Introduction}
Biometric-based authentication has been widely deployed in identity management systems. However, despite their ease of use, the proliferation of centralized biometric databases has led to significant concerns about the security and privacy of biometric data. Disclosing biometric data may expose private and sensitive information belonging to the user. Moreover, if the compromised biometric data can not be revoked, it remains permanently linked with the user's identity.
Several template protection techniques (BTP) have been developed to address these issues, and cancelable biometrics (CB) is one of them. 

CB utilizes a parametrized, irreversible, and revokable transformation to ensure the security and privacy of the biometric template \cite{patel_cancelable_2015-BTPOVERVIEW,sandhya_biometric_2017-BTPOVERVIEW,chandra_cancelable_2011-BTPOVERVIEW,nandakumar_biometric_2015-BTPOVERVIEW}. If the transformed biometric template (pseudonymous identifier) is compromised, a new template can be generated for the same user just by changing the parameters of the transformation function. A CB system generally consists of a feature extractor, a parametrized transformation function, and a matcher to generate a matching score in the transform domain. The parametrized transformation function can be obtained by choosing an appropriate non-invertible function or salting. The associated parameters can be passwords or user-specific seeded (pseudo) random numbers.

CB needs to comply with the criteria of BTP, namely irreversibility, revocability, and unlikability, identified in ISO24757~\cite{ISOIEC24745}. Revocability requires the system to be able to issue a new protected template to replace the compromised one. Unlinkability, or cross-matching, refers to the impossibility of determining whether two transformed templates, derived from different applications, come from the same subject. Irreversibility means that the retrieval of the original biometric data (or any information associated with the original template) from a stored biometric template is \textit{computationally} unfeasible. In ISO 30136, performance preservation is formally included as another criterion for biometric template protection~\cite{ISOIEC30136}.

However, security attacks, such as the pre-image attacks on which we focus in this paper, are often effective at jeopardizing the BTP schemes. Furthermore, the associate threat is also underestimated. For example, the system security could be compromised if the attacker could recover the original biometric data, either fully or partially, by reverse engineering or a pre-image attack. Subsequently, illegal access can be made based on the pre-image\footnote{Reconstructed template and pre-image are used interchangeably in the following content.}. For instance, \cite{dong-BTAS-2019} demonstrated that a pre-image attack could generate an approximated template (pre-image) and further be utilized to gain illegal access to a system. In this sense, a systematic analysis of the vulnerability of CB schemes under these attacks is urgently needed, which has been overlooked. 

On the other hand, existing works on the pre-image constructions, such as \cite{dong-BTAS-2019,atighehchi2019cryptanalysis,feng2014masquerade22222,feng2014masquerade}, usually assume that the same type cancelable scheme (same transformation function) applies on the pre-image, which is unrealistic. Therefore, a more comprehensive study with \textit{different cancelable schemes} should be carried out. Besides, information leakage is often inevitable when applying cancelable transformations. Yet, quantifying information leakage in a CB scheme remains an open problem.

In this paper, we first present a generic pre-image attack for CB. Then, on top of the pre-image attack, we formulate a practical \textit{cross-transformation attack} that operates on two different CB systems. Furthermore, the risk of the pre-image attack by exploiting the distance preserving property is demonstrated theoretically. Finally, a general framework is also proposed to measure the information leakage of CB schemes from the distance preserving perspective. 

The contributions of this paper can be summarized as
\begin{enumerate}
\item We theoretically reveal that the property of distance preserving is a source of vulnerability in CB schemes. We experimentally demonstrate that the pre-image attacks under Kerckhoffs's assumption are feasible (section \ref{section.attacktheory}).
\item We propose a new security threat based on the pre-image attack, namely the cross-transformation attack; notably, this attack is effective regardless of the CB transformation functions employed (section \ref{section.attack}).
\item We propose to estimate the information leakage from the distance-preserving property based on mutual information. This is critical to BTP yet being studied systematically in literature (section \ref{section.infoleakage}).
\item Thorough experiments on six well-known CB schemes covering the three most popular biometric modalities, i.e., face, iris, and fingerprint, are conducted. The analytical and experimental results indicate that a completely secure BTP scheme remains challenging (section \ref{section.experiment}).
\end{enumerate}

\section{Background\label{section.background}}
\subsection{Review of cancelable biometrics}
\Stress{BioHashing} \cite{teoh_random_2006} is a generic tokenized \textit{two-factor} CB scheme. The user-specific key derived from an external token is mixed with a biometric feature (salting) and binarized.
The $n$-bit BioHash code $\bm{c}$ of a feature vector $X\in \mathbb{R}^N$ is $c_i=Sgn(\sum \bm{x^Tb_i}-\tau)$, where $Sgn(\cdot)$ is the sign function, $\tau$ is an empirically determined threshold, and $\bm{b_i}\in \mathbb{R}^N,i=1,...,n (n\leq N) $ is a random vector.
Since the hash code is a binary string, the Hamming distance is chosen for the similarity metric between pairs of hash codes.
If a biometric template is compromised, the BioHash code can be replaced by a newly generated pseudo-random vector.

\Stress{Bloom Filters} \cite{rathgeb2013bloomfilteriris} can be applied to generate a CB template by mapping the biometric data onto a bit array.
In~\cite{rathgeb2013bloomfilteriris}, Bloom Filters are applied to generate a cancelable template from an iris code.
Specifically, a two-dimensional binary iris code feature, with width $W$ and height $H$, is extracted first. The $W\times H$ iris codes are divided into $K$ blocks of size $l=W/K$, with $\omega \leq H$ bits per column (word size $\omega$).
For each block, all column sequences are projected to the designated locations of the Bloom filter bit array of size $2^\omega$. The final template of size $K\times 2^\omega$ is generated by collecting $K$ different Bloom filters.
Unlinkability is achieved by computing the XOR of the codewords with an application-specific bit vector $T \in \{0,1\}^w$ before mapping the iris code to the Bloom filters.
However, as in \cite{hermans2014bloombecomesdoom} Bloom filters are proved to fail to achieve unlinkability even with a simple attack; an additional structure-preserving feature re-arrangement is proposed in \cite{gomez2016unlinkablebloomfilter}. The re-arrangement can be regarded as a permutation operation with a permutation parameter, which can dissipate the statistical composition but preserve the discrimination of features.

\Stress{Indexing-First-One (IFO) hashing} \cite{lai2017cancelable} is inspired by the min-hashing technique, applied in search engines to detect duplicate web pages \cite{broder2000minhashing}.
The $P$-order Hadamard product and modulo threshold functions are also incorporated to protect privacy.
In the IFO scheme, the iris code $X$ with $n$ columns is permuted columnwise, based on $P$ randomly generated permutation vectors, and the permuted iris code is denoted by $X^{\prime}=\left\{X_{l}^{\prime} | l=1, \ldots, P\right\}$.
Next, the Hadamard product code is generated by elementwise multiplication, and denoted by $\boldsymbol{X}^{P}=\prod_{l=1}^{p}\left(\boldsymbol{X}_{l}^{\prime}\right)$.
In the third step, select the first $K$ elements of each row in the product code $X^{P}$, and record the index value of the first occurrence of the bit '1,' denoted by $C_{X}$.
Lastly, for every $C_{X} \geq K-\tau$, a many-to-one computation is performed by $C_{X}^{\prime}=C_{X} mod (K-\tau)$.
This process is repeated by using $m$ different permutation sets to form an $ n \times m$ IFO hashed code, denoted by $\boldsymbol{C}_{X}^{\prime}=\left\{C_{X i}^{\prime} \in Z^{n} i=1, \ldots, m\right\}$, where $\boldsymbol{C}_{X}^{\prime} \in[0, K-\tau-1]$.
The security of IFO is achieved by combining permutation, a $K$-window operation, the Hadamard product, and a modular threshold function.

\Stress{Index-of-Max (IoM)} \cite{jin2017ranking} is a ranking-based Locality Preserving Sensitive hashing technique, where a cancelable template is generated by collecting the max indices generated by repeated random projections.
Unlike BioHashing, an integer-valued template can be generated and easily converted into a binary vector. Specifically, the product of the feature vector $\bm{x}$ and the randomly generated Gaussian matrices $\bm{W}$ are computed, the index of the max value in the product is recorded as one hash code, and then a template of size $m$ can be generated by repeating this process $m$ times. 

\Stress{Non-linear multi-dimensional spectral hashing (NMDSH)} \cite{dong-IWBF-2019}, initially proposed for face template protection, is an extension of graph-based Hamming embedding~\cite{jin-PRL-2014}, where the distance between feature vectors is compared with the Hamming distance between the corresponding hash codes.
First, the distance between vectors is computed as the affinity matrix $\bm{W}$; then, the hashing problem is reduced to a binary matrix factorization of $\bm{W}$.
The NMDSH algorithm is derived by adding a nonlinearity to the original MSDH algorithm.
Specifically, the one-dimensional eigenfunctions $\phi_{ij}(\bm{x}(i))$ and the corresponding eigenvalues $\lambda_{ij}$ are computed from the training data as shown in (1-2) in \cite{dong-IWBF-2019}, where $\phi_{ij}(\bm{x}(i))$ is the $j$-th eigenfunction of the $i$-th coordinate, and $\lambda_{ij}$ is the corresponding eigenvalue.
Next the eigenvalues $\lambda_{ij}$ are sorted in ascending order, and the top $k$ indices are selected to construct the set $A = \{(i_1,i_1),(i_2,i_2 ), \cdots,(i_k,i_k )\}$.
Then, each data sample $\bm{x}$ from the test dataset is encoded as $y_{ij}(\bm{x}) = \sin (\phi_{ij}(\bm{x}))$ for all $(i,j) \in A$.
The final output is $y = q(\phi_{ij}(\bm{x}))$ where $q(\cdot)$ is a nonlinear softmod function defined as follows: 
\begin{equation}
q(\bm{x})=\frac{2}{1+e^{-8\sin(\alpha\pi\ \bm{x})}} -1,\label{eq::mdshsoftmod}
\end{equation}
where $\alpha$ is the nonlinear rate to be determined empirically. The final output is $y = q(\phi_{ij}(\bm{x}))$. The nonlinearity of $q(\cdot)$ allows the elimination of the correlation between the hash code and the original data. However, a high value of $\alpha$, which induces greater nonlinearity and leads to a large distortion of the projected data, may harm the matching performance.

\Stress{Two-factor Protected Minutia Cylinder-Code (2PMCC)} \cite{ferrara20142pmcc} is a template protection scheme based on Protected Minutia Cylinder-Code (PMCC) \cite{ferrara2012noninvertiblepmcc} and the Minutia Cylinder-Code (MCC) \cite{cappelli2010minutiamcc} for fingerprints. The MCC is a local descriptor of each minutia and encodes spatial and directional relations between the minutia point and its neighborhood within a certain radius.
In PMCC, to achieve non-invertibility, the Karhunen--Loève transformation \cite{fukunaga1993statistical} is applied to project the feature vector extracted from each cylinder, followed by a binarization step to generate the protected template.
In \cite{ferrara2012noninvertiblepmcc}, it is claimed that the PMCC can preserve the distances of two templates, although the original information is not present in the protected template.
However, as the template generated from the PMCC algorithm can not be revoked, a revokable two-factor protection scheme, namely 2PMCC, is proposed in \cite{ferrara20142pmcc}.
The non-invertibility of 2PMCC depends on PMCC, a secret $s$, and a dimension reduction parameter $k$.
The non-invertibility of PMCC has been examined under two attacks \cite{feng2010fingerprintreconstruct}, namely, directly attacking the original template (type-I attack) and attacking a second template generated by different impressions from the same finger (type-II attack).

\subsection{Pre-image attacks}

In \cite{lacharme2013preimagebiohashing}, a pre-image attack for BioHashing based on a genetic algorithm was applied to fingerprint templates. The algorithm reconstructed a feature vector which is an approximation of the original template. Since then, several genetic algorithms have been used to attack cancelable biometric systems, but without a sufficient quantification of the information leakage.

Pagnin et al. \cite{pagnin2014leakage} proved that the information embedded in the reference template can be recovered when the matching is performed using distance measurements, such as the Hamming and the Euclidean distances. This information leakage enables a hill-climbing attack, which, from a template, could recover the original biometric data (e.g., a center search attack), even if it has been encrypted. The presented results were obtained from discrete data, and no quantitative analysis was provided. 


In \cite{feng2014masquerade22222,feng2014masquerade}, a masquerade attack, based on the perceptron and neural-network learning, is applied in the case of both a known and unknown hashing algorithm. In the first case, assuming the attacker can generate the matching score from a set of binary templates and the hashing algorithm is publicly released, the unknown parameters, the random projection matrices, and the BioHashing threshold are estimated. Then, the face image is reconstructed by a hill-climbing attack. In the second case, assuming the hashing algorithm is unknown, and the attacker can acquire a set of face images from different identities, the known feature extractor extracts the real-valued feature vectors. Next, a three-layer MLP is trained to model the hashing and matching processes. Finally, the features are reconstructed by means of a standard hill-climbing algorithm.

In \cite{atighehchi2019cryptanalysis}, two recent CB schemes are cryptanalyzed based on the index-of-max (IoM) hashing function. A constrained optimization algorithm is proposed to generate pre-images of the IoM hash codes. The generated pre-image is used to perform a few attacks. However, the search space constraint can only be applied to specific CB schemes such as the IoM.

In 2019, Chen et al. \cite{chen-SPDSQ-2019} proposed a new biometric hashing, Deep Secure Quantization (DSQ), to issue a pre-image attack. The mutual information quantified the leakage of a CB scheme as $ I(d^-;s^-)$, where $d^-$ and $s^-$ are the normalized distances in the input feature space and transform space, and the symbol $-$ indicates the inter-class matching distance. A hashnet was proposed to achieve $I(d^-;s^-) \leq \delta$, where $\delta$ is an upper-bound constant, thus {\em avoid} the information leakage. However, the DSQ hashing function requires training. Hence it is data-dependent. Besides, as a closed-set protocol (overlap identities from training and testing datasets) is adopted, DSQ may not be suitable for practical usage cases.

Recently, a similarity-based attack, performed with a genetic algorithm under Kerckhoffs's assumption, was proposed in~\cite{dong-BTAS-2019} by reconstructing an approximated feature, i.e., a pre-image, from a CB template. The soundness of the pre-image attack is attributed to the information leakage originating from the template matching. The experimental results confirmed the vulnerability to this attack. More importantly, the pre-image is not necessarily similar or close to the original feature being attacked. The preliminary work in \cite{dong-BTAS-2019} concluded that most CB schemes based on distance preserving transformations might be vulnerable to a pre-image attack. However, a theoretical analysis and the information leakage quantifying was not provided in \cite{dong-BTAS-2019}.

\vspace{-0.2cm}
\section{Attack based on the distance-preserving property\label{section.attack}}
This paper uses calligraphic characters, such as $\mathcal{X}$, to denote a space. In addition, we use uppercase (e.g.,$X$) to indicate a random variable (e.g., in the computation of entropy or probability), and we write lowercase bold letters (e.g., $\bm{x}$ ) to denote a vector.
\vspace{-0.3cm}
\subsection{Formulating the attack}
In the pre-image attack, we assume that the attacker can access the protected template $\bm{y}=f(\bm{x}) \in \mathcal{Y}$ and learn the transformation function $\mathcal{F}$ as well as the parameters of the function (Kerckhoffs’s assumption). The goal is to find the pre-image $\widehat{x}$, which can be expressed as follows:
\begin{equation}
\widehat{\bm{x}} = \operatorname*{arg\,min}_{\widehat{\bm{x}}} d(f(\widehat{\bm{x}}),f(\bm{x})), f \in \mathcal{F}, \label{eq.finalgoal}
\end{equation}
where $d(\cdot)$ denotes the distance function (e.g., Hamming distance), $f(\cdot)$ indicates an algorithm-specific distance preserving function and $f(\bm{x})$ is the compromised template in the database. The pre-image attacks rely on the possibility of breaking a matcher by exceeding a given matching score based on the $\widehat{x}$ without the need to provide an exact copy of the genuine template.

In reality, the attacker may acquire more than one instance of the transformed templates. Given $n$ compromised templates, the goal of the attack can be formulated as
\begin{equation}
\widehat{\bm{x}} = \operatorname*{arg\,min}_{\widehat{\bm{x}}} \frac{1}{n}\sum_{i=1}^n { d(f(\widehat{\bm{x}}),f(\bm{x_i}))}, \label{eq.finalgoal2}
\end{equation}
where $f(\bm{x_i})$ denotes several instances $i=1,2...n$ of the compromised template from the same identity. Hence, the objective function of the GA can be formulated as
\begin{equation}
l =\frac{1}{n}\sum_{i=1}^n { d(f(\widehat{\bm{x}}),f(\bm{x_i}))}, \label{eq.objectivefinal}
\end{equation}
It is worth noting that $n=1$ is equivalent to the single template attack.

To find the best solution for (\ref{eq.finalgoal},\ref{eq.finalgoal2}), various algorithms can be utilized, such as genetic algorithm, hill-climbing, particle swarm and etc. We employ a genetic algorithm (GA) in the pre-image attack in our experiments. Details of GA algorithm are presented in Section \ref{sec.gaappendix}.

\newcolumntype{L}[1]{>{\raggedright\let\newline\\\arraybackslash\hspace{0pt}}m{#1}}
\newcolumntype{C}[1]{>{\centering\let\newline\\\arraybackslash\hspace{0pt}}m{#1}}

\begin{table*}[t!]
\renewcommand{\arraystretch}{1.5}
\centering
\caption{Example settings for Sys C and Sys T}
\label{table::sysCSysTsettings}
\resizebox{.99\linewidth}{!}{
\begin{tabular}{l c C{5cm} C{5cm}} 
\toprule
\multirow{2}{*}{\normalsize Normal/Attempt} & \multicolumn{1}{c}{\normalsize Sys C} & \multicolumn{2}{c}{\normalsize Sys T} \\\cmidrule(lr){2-2} \cmidrule(lr){3-4}
{} & \multicolumn{1}{c}{Normal enroll} & \multicolumn{1}{c}{Normal enroll} & \multicolumn{1}{c}{Attempt~} \\ 
\cmidrule(lr){1-1}\cmidrule(lr){2-2}\cmidrule(lr){3-3}\cmidrule(lr){4-4}
Input Features & Face features $\bm{x}$ of person A & Face features $\bm{x}^\prime$ of person A & $\widehat{\bm{x}}$ \\ 
\arrayrulecolor{black}
Template & $y=f(\bm{x})$ & $y^\prime=f(\bm{x}^\prime) $ & $\widehat{\bm{y}}=f(\widehat{\bm{x}}) $ \\ 

Transformation & BioHashing $f(\cdot)$ & \multicolumn{2}{c}{BioHashing $f(\cdot)$} \\ 

Parameters & Random projection matrix $W$ & \multicolumn{2}{c}{New random projection,~matrix $W^\prime$} \vspace{1mm}\\

Adversary Actions & \Centerstack[l]{Attacker reconstructs a pre-image $\widehat{\bm{x}}$ of $\bm{x}$ based on the\\ parameters, $\bm{y}$ and the knowledge of $f(\cdot)$.} & \multicolumn{2}{l}{\Centerstack[l]{Attempt to access the system using $\widehat{\bm{x}}$ and new parameters, the transformation function \\ can be the same or different.}}\\
\arrayrulecolor{black}\bottomrule
\end{tabular}
}
\vspace{-0.5cm}
\end{table*}

\subsection{System settings}
\begin{figure}[t!]
\centering
\includegraphics[width=0.8\linewidth,trim=11cm 6.5cm 11cm 5.5cm, clip]{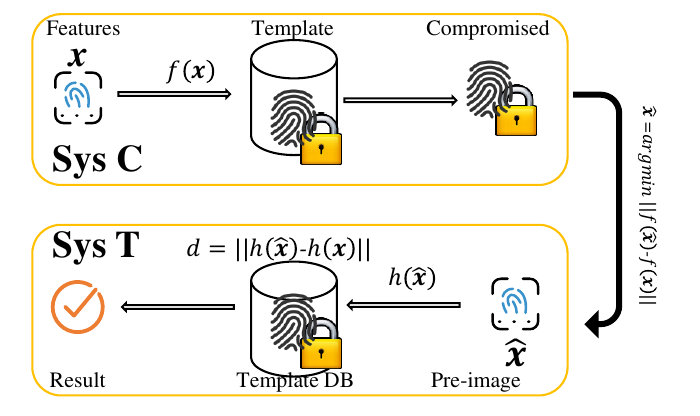} 
\caption{Diagram describes the attack sequence. (Top) The attacker retrieves a template from the compromised system ($Sys~C$) to generate the pre-image. (Bottom) The attacker hacks the target system ($Sys~T$) with the pre-image. The CB functions $f(\cdot)$ and $h(\cdot)$ could be the same, while the feature vector are generated from the same extractor. \label{Figure::attackflow}}
\vspace{-0.5cm}
\end{figure}

To initiate the attack, it is necessary to define the configuration of the two subsystems (Fig.~\ref{Figure::attackflow}).

\Stress{Compromised system ($Sys~C$)}: this is a biometric system protected by a CB algorithm. We assume the worst-case situation where the shared parameters of all users are stolen—for example, a compromised access control system in Company A. 

\Stress{Targeted system ($Sys~T$)}: this is a biometric system subject to a pre-image attack. Each user's pre-image $\bm{\widehat{x}}$ from a transformed template stored in $Sys~C$ was used to break into $Sys~T$. The parameters of the CB function are usually different from $Sys~C$. For example, an attacker tries to attack Company B with the compromised template from Company A. 

As discussed above, the pre-image attack aims to find the best solution for (\ref{eq.finalgoal}).
The attack sequence starts from $Sys~C$ and is carried out to find the best pre-image to break $Sys~T$. Both $Sys~C$ and $Sys~T$ use the same feature extractor.

It is noteworthy that the cancelable transformation functions $f(\cdot)$ and $h(\cdot)$ could be the same function but parametrized differently (for example, Biohashing with different parameters) or even different functions. We refer to this attack scenario as the \textit{\textbf{cross-transformation}} attack.

Compared to the type-I and type-II attacks in \cite{feng2010fingerprintreconstruct}, this approach is dedicated to attacking a CB system concerning revocability and non-invertibility.
The more realistic cases, where different transformation functions/parameters have been applied to other systems, have also been considered (see section \ref{section.cross}). For the attack, Kerckhoffs's assumption is adopted.
A detailed example of $Sys~C$ and $Sys~T$ is presented in Table \ref{table::sysCSysTsettings}. Note that our work does not consider data-dependent transformations (e.g., learning-based cancelable transformation). The user information is usually preserved in those approaches, and the risks are unavoidable \cite{liu2020privacy}. Therefore, we focus on the data-independent transformation functions and assume that there is no information leakage from the transformation functions and their parameters.

\section{Theoretical Risk Analysis of the Distance-preservation\label{section.attacktheory}}
Most CB schemes essentially comply with the notion of a distance-preserving transformation (see \cite{jin2017ranking,lai2017cancelable} for examples), which projects biometric features onto the transform space while preserving the distance between two templates. Therefore, matching two features in the transform space is almost equivalent to performing the same operation in the original feature space. 

In this section, we analyze such cancelable transformation functions by formally defining the degree of distance preservation. We show that a transformation with distance preservation property preserves the recognition performance for the templates. Still, this preservation characteristic also weakens protection against the pre-image attack.

The distance-preserving property of a transformation can be defined in different manners. We consider three types of widely known distance-preservation: locality-preservation, isometry, and locality-sensitivity.

\subsection{On Locality-Preservation \label{sec.onlocality}}
We begin by formalizing the definition of locality-preserving property in a general probabilistic framework.

\noindent
\textbf{Definition.}
Given the original feature space $\mathcal{X}$ and arbitrary random feature vectors $X,X_1,X_2 \in \mathcal{X}$, a transformation $h$ is \textbf{locality-preserving with a degree} $\epsilon$ if 
\begin{multline}
\Prob \left[
d( h(X), h(X_1)) < d(h(X), h(X_2)) \right.
\\
\left.
\mid
d(X, X_1) < d(X, X_2)
\right]
\geq 1 - \epsilon
\end{multline}
and 
\begin{multline}
\Prob \left[
d( X, X_1) \geq d(X, X_2) \right.
\\
\left.
\mid
d(h(X), h(X_1)) \geq d(h(X), h(X_2))
\right]
\geq 1 - \epsilon,
\end{multline}
The degree $\epsilon$ determines the intensity of locality preservation of the transformation $h$. The lower the degree $\epsilon$ is, the more the transformation is locality-preserving.
With $\epsilon =0$, we attain the original definition of the (deterministic) locality preservation \cite{chin1991complexity,chin1994locality}
\begin{multline}
\label{eq::locality}
d(X, X_1) < d(X, X_2)
\\
\implies
d(h(X), h(X_1)) < d(h(X), h(X_2)).
\end{multline}
In this respect, our probabilistic definition can be understood as an extension of the original one. A locality-preserving transformation preserves the relativistic order of distances. In the deterministic case, the distance between features is related to the distance between templates by a strictly increasing function (as depicted in Fig.~\ref{Figure::theory}bcd).
Hence, the transformation makes the recognition performance preserved in the feature space, but unfortunately, it keeps the attacker's performance as well; namely, the success rate of the pre-image attack remains high even after template renewal. We formalize these claims below.

\subsubsection{Performance Preservation of Locality-Preserving Transformation}
\begin{prop}
Given random features $X$ and $X_{in}$ which are within the same class $C_y$, and $X_{out} \notin C_y$ which is an inter-class feature, assume a high recognition performance on the features 
\begin{equation}
\Prob \left[
d(X, X_{in}) < d(X, X_{out})
\right]
\geq p
\end{equation}
with $p >0$.
Then, for a locality-preserving transformation $h$ with a degree $\epsilon$,
\begin{equation}
\mathbb{P} \left[
d(h(X), h(X_{in})) < d(h(X), h(X_{out}))
\right] \geq (1-\epsilon)p.
\end{equation}
\end{prop}

Here, $C_k \subseteq \mathcal{X}$ denotes the set of intra-class features corresponding to the class $k$. The proof of this and subsequent propositions are included in the appendix section \ref{section.attacktheoryappdix}.

In the above proposition, 
$
\Prob \left[
d(X, X_{in}) < d(X, X_{out})
\right]
$
indicates the recognition performance of the original features while
$\mathbb{P} \left[
d(h(X), h(X_{in})) < d(h(X), h(X_{out}))
\right]$ refers to that of the transformed features. The proposition states that a distance-preserving transformation preserves the performance based on the degree of distance-preservation.

\subsubsection{Pre-image Attack on Locally-Preserving Transformation. }
Unfortunately, the transformation also preserves the attack success rate, regardless of template renewal, as shown by the following proposition.

\begin{prop}
\label{prop.cross}
Let $X \in C_y$ and $X_{out} \notin C_y$, and $h$ be locality-preserving with degree $\epsilon$. Let $X_a$ be a pre-image of $h(X)$ with an attack success rate $p$ such that
\begin{equation}
\Prob \left[
d( h(X) , h(X_a) ) < d( h(X), h(X_{out}) )
\right] \geq p.
\end{equation}
Then, for any locality-preserving transformation $f$ with the degree of $\delta$, we have
\begin{equation}
\Prob \left[
d( f(X) , f(X_a) ) < d( f(X), f(X_{out}) )
\right] \geq (1-\epsilon) (1-\delta) p.
\end{equation}
\end{prop}

The proposition states that, once we breach the system for the initial transformation $h$ with the pre-image $\bm{x}_a$, the same pre-image can successfully attack the replaced system with a probability $(1 - \delta) (1-\epsilon) p$. Thus, the proposition implicates that a high accuracy performance (namely, low values of $\epsilon$ and $\delta$) inevitably leads to a high attack success probability.

\subsubsection{Information Leakage of Locality-Preserving Transformation}

Preserving the distance relationship by locality enables one to infer the original feature from the transformed template. This is formally verified by the following.
\begin{prop}
\label{prop.locality_leakage}
Let $X$ be a random feature of discrete range with $n$ values. Assume $h$ is locality-preserving with a sufficiently small degree $\epsilon$ such that $1 - \epsilon > 1/e$ and $\epsilon < 1/e$. Then,
\begin{equation}
I(X, h(X)) \geq H(X) - \left[ 
\epsilon + 2(n-1)\frac{\sqrt{\epsilon}}{e}
\right]
\end{equation}
where the entropy $H(X)$ is a constant and a maximum of the leakage given a certain biometric modality; $H(X) \geq I(X,h(X))$.
\end{prop}

Due to the lower bound of the leakage given in above proposition, the leakage is increased if the transformation is more locality-preserving (namely, the higher the probability $p$). 

Under practical consideration, however, estimation of the information leaked from the features is intractable due to the curse of the high-dimensionality of the features. Fortunately, the mutual relationship between the feature $X$ and its template $h(X)$ can be relaxed and represented by the relationship between their corresponding feature distance $S=d(X_1, X_2)$ and template distance $T=d(h(X_1), h(X_2))$. In particular, the degree $\epsilon$ of distance preservation is correspondent to the leakage $I(S,T)$ just as it is correspondent to the leakage $I(X, h(X))$:

\begin{prop}
\label{prop.leakage_translate}
Let $X_1, X_2$ be discrete random features with at most $n$ values. Let $S=d(X_1, X_2)$ and $T = d(h(X_1), h(X_2))$ where $h$ is locality-preserving with degree $\epsilon$. Then, 
\begin{equation}
I(S,T) \geq H(S) - n^2 \left[
\epsilon + \frac{2(n^2-1) \sqrt{\epsilon}}{e}
\right]
\end{equation}
where $H(S)$ is a constant and a maximum of the leakage $I(S,T)$ given a certain biometric modality; $H(S) \geq I(S,T)$.
\end{prop}

\subsection{On Isometric Distance-Preservation}
\medskip
\noindent
\textbf{Definition.}
The notion of distance-preserving transformation can be formalized in a more strict, geometric context. Namely,
the distance-preservation with a degree of $\epsilon$ can be geometrically characterized by\footnote{We slightly abuse the $\epsilon$, the $\epsilon$ here is different from section \ref{sec.onlocality} by definition.}
\begin{equation}
\label{eq::isometry}
\epsilon = \inf \{ \epsilon : \lvert d( h(\bm{x}_1), h(\bm{x}_2) ) - d(\bm{x}_1, \bm{x}_2) \rvert < \epsilon, \forall \bm{x}_1, \bm{x}_2 \in \mathcal{X} \}
\end{equation}
where $\bm{x}_1$ and $\bm{x}_2$ denote arbitrary feature vectors in the feature space $\mathcal{X}$. 
Following the Hausdorff approximation, we define \textbf{isometric distance-preserving} transformation $h$ as an $\epsilon$-isometry that satisfies the Eq.~\eqref{eq::isometry} with $\epsilon {<} \infty$.
The isometric distance preservation with $\epsilon {=} 0$ implicates the locality preservation defined in Eq.~\eqref{eq::locality}. Moreover, with small $\epsilon$, the feature distance is related to the transformed distance by a linear increasing function (Fig~\ref{Figure::theory}d). Hence, isometric distance preservation can be understood as a more specific, strict condition than the locality preservation. 

As in the locality-preserving transformation, an isometric distance-preserving transformation preserves the performance of original features at the cost of vulnerability to pre-image attack. Moreover, high isometric distance preservation inevitably results in high information leakage. Our observations are presented formally as below:

\subsubsection{Performance Preservation of Isometric Distance-Preserving Transformation}
For any inter-class index pair $(k, l)$, the set distance metric $d(C_k, C_l)$ on the sets $C_k$ and $C_l$ can be defined by
\begin{equation}
d(C_k, C_l) 
= \inf \{
d(\bm{x}_k, \bm{x}_l) : \bm{x}_k \in C_k, \bm{x}_l \in C_l
\}.
\end{equation}
Consider the family of isometric distance-preserving transformations
\begin{equation}
\label{eq.hfamily}
\mathcal{H}_\epsilon = 
\begin{aligned}
\{h : h \text{ whose isometry degree} \geq \epsilon\}
\end{aligned}
\end{equation}
with small $\epsilon>0$.
Any transformation in $\mathcal{H}_\epsilon$ preserves the performance of original features as stated by the below proposition:

\begin{prop} 
For any inter-class index pair $(k, l)$, if $d(C_k, C_l) > \epsilon$, then $h(C_k)$ and $h(C_l)$ are linearly separated for any $h \in \mathcal{H}_\epsilon$.
\end{prop}

\subsubsection{Pre-image Attack on Isometric Distance-Preserving Transformation}
Unfortunately, like locality-preservation, a high degree of isometric distance-preservation allows attackers to attack any CB system easily. This is formalized in the following proposition:

\begin{prop}
Given any feature $\bm{x} \in \mathcal{X}$, and let $\epsilon > 0$ be a constant for a given $h \in \mathcal{H}_\epsilon$. Assume a pre-image $\bm{x}_a$ satisfies that
\begin{equation}
d( h(\bm{x}), h(\bm{x}_a) ) < \tau
\end{equation}
for a fixed $\tau > 0$.
then 
\begin{equation}
d( f(\bm{x}), f(\bm{x}_a) ) < \tau_0 + 2 \epsilon
\end{equation}
for any $f \in \mathcal{H}_\epsilon$.
Here $\tau_0$ is a value satisfies $d(h(\bm{x}), h(\bm{x}_a)) < \tau_0 < \tau$ and independent of $\epsilon$. 
If $\epsilon < \tau - \tau_0$, moreover, then
\begin{equation}
d( f(\bm{x}), f(\bm{x}_a) ) < \tau.
\end{equation}
\end{prop}

In addition, isometric distance-preserving transformations are prone to information leakage; a high degree of isometry enables the attacker to reveal the original feature. The following proposition indicates this:

\begin{prop}
\label{prop.isoleakage}
For $h \in \mathcal{H}_\epsilon$,
\begin{equation}
I(X, h(X)) \geq c + d \log 1/(2\epsilon)
\end{equation}
where $c=H(X)$ is a constant and $d$ is the feature dimension such that $X \in \mathcal{X} \subseteq \mathbb{R}^d$. $I(\cdot, \cdot)$ denotes the mutual information, and $H(X)$ the entropy of the random feature $X$.
\end{prop}
\begin{figure*}[t!]
\centering
\includegraphics[width=0.8\linewidth,trim=0cm 9cm 0cm 0cm, clip]{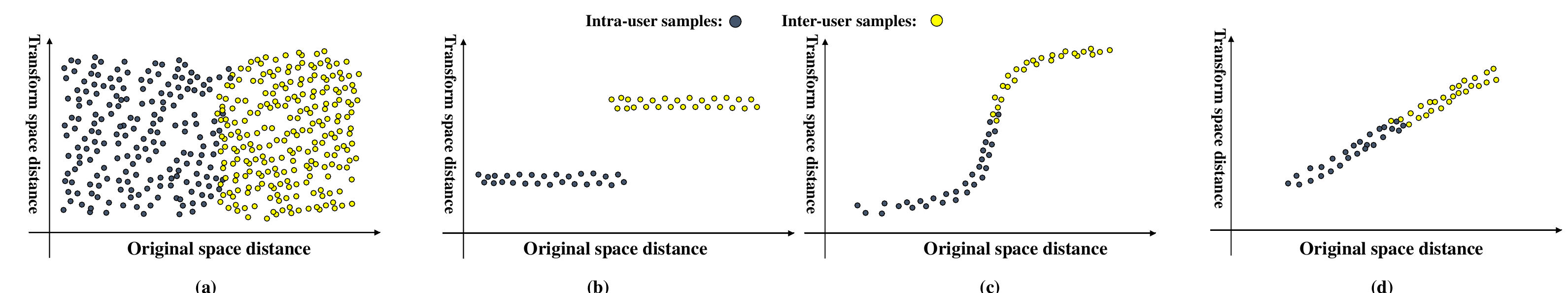} 
\caption{Illustrations of possible distance preservation grounded CB transformations. (a) achieves the best security but zero biometric utility. Representatives of (b) are learning-based CB schemes such as DSQ; it achieves an excellent performance but is subject to model inversion attacks. (c) preserves performance but has high information leakage. (d) preserves performance but has the worst information leakage.}
\label{Figure::theory}
\end{figure*}

\subsection{Locality-Sensitive Hashing}

\medskip
\noindent
Now, we consider distance-preservation based on locality-sensitive hashing. 

\noindent
\textbf{Definition.} A transformation $h$ is a \textbf{locality-sensitive hashing (LSH) with degree} $\epsilon{>}0$ if there are thresholds $\tau, r>0$ such that for any probability $p \geq 0$
\begin{equation}
\Prob \left[
d( h(X), h(X_{in}) ) \leq \tau
\right]
\geq (1-\epsilon) p 
\end{equation}
if $\Prob[ d( X, X_{in} ) < r ] \geq p$, and 
\begin{equation}
\Prob \left[
d( h(X), h(X_{out}) ) > \tau
\right]
\geq (1-\epsilon) p 
\end{equation}
if $\Prob[ d( X, X_{out} ) > r ] \geq p$.
From now on, we term such a transformation by $\epsilon$-\textbf{LSH}. With $\tau =0$, $\epsilon=1$, and $p=1$, we recover the original definition of LSH given in \cite{indyk1998approximate}. The degree $\epsilon$ indicates how well the transformation $h$ follows the LSH property. The thresholds $\tau$ and $r$, on the other hand, are data-dependent values and are pre-selected based on the distribution of features.

\medskip
\noindent
\textbf{Performance Preservation.}
By definition, a $\epsilon$-LSH transformation preserves the recognition of original features. Precisely, if $\epsilon$ is smaller, then the LSH transformation better holds the performance.

\medskip
\noindent
\textbf{Pre-image Attack.}
Like other distance-preservation properties, the LSH property also makes the CB system vulnerable to the pre-image attack. The following proposition verifies this:

\begin{prop}
\label{prop.lshattack}
Assume a high attack success rate of the pre-image $X_a$ attack
\begin{equation}
\Prob \left[
d( h(X), h(X_a) ) \leq \tau 
\right]
\geq p
\end{equation}
on the initial transformation $h$.
Then, for any other $\delta$-LSH transformation $f$, a high attack success rate is posed even after template renewal by $f$:
\begin{equation}
\Prob \left[
d( f(X), f(X_a) ) \leq \tau 
\right]
\geq \frac{p - \epsilon}{1 - \epsilon}(1 - \delta).
\end{equation}
\end{prop}

The proposition states that if an attack algorithm can generate a pre-image $X_a$ whose template is considered to be in the same class of $X$ with the probability $p$, then even after template renewal by another LSH $f$, the renewed template of the same pre-image is regarded to be in the same class of $X$ with the probability $(p-\epsilon)(1-\delta)/(1-\epsilon)$. Thus, renewing the template would not be much helpful for protection against similarity attacks by pre-image. 

\begin{table*}[t]
\centering
\begin{threeparttable}
\caption{There is an exact correspondence between Locality-Sensitive Hashing, accuracy performance, and attack success rate after template renewal. On the other hand, the information leakage corresponds to locality-preservation.}
\label{table::theory}
\begin{tabular}{p{1.4cm}c:p{1.3cm} p{1.5cm} p{1.5cm}: p{1.5cm} p{1.5cm} p{1.7cm} }
\toprule
Scenario Visualization & Algorithm Examples & Locality Sensitive Hashing& Performance of templates & Pre-image Attack Risk$\ddagger$ & Locality Preserving & Isometric Distance Preserving& {Info. Leakage from Templates}\\
\midrule
Fig.~\ref{Figure::theory}a & MD5/SHA256 & No & No & No & No & No & No \\ 
Fig.~\ref{Figure::theory}b & DSQ\cite{chen-SPDSQ-2019} & Yes & Good & High & Low & Low & Low$\dagger$ \\ 
Fig.~\ref{Figure::theory}c & NMDSH\cite{dong-IWBF-2019} & Yes & Good & High & High & High & High \\ 
Fig.~\ref{Figure::theory}d & BioHashing \cite{teoh_random_2006} & Yes & Good & High & Very High & Very High & Very High \\ 
\bottomrule
\end{tabular}
\begin{tablenotes}
\small
\item $\dagger$ Learning-based CB schemes enjoy lower information leakage \cite{chen-SPDSQ-2019} but are subjected to model inversion attack risks \cite{liu2020privacy}. 
\item $\ddagger$ here indicates the pre-image attack risk after template renewal, and we assume a high attack success rate of the pre-image attack on the initial transformation.
\end{tablenotes}
\end{threeparttable}
\vspace{-0.5cm}
\end{table*}

\noindent
\textbf{Information Leakage.} Unlike other distance-preserving properties, information leakage $I(X, h(X))$ is not linked to the LSH property. Particularly, there are LSH transformations with low information leakage (Fig.~\ref{Figure::theory}b) and ones with high information leakage (Fig.~\ref{Figure::theory}cd). The leakage can depend on the specific thresholds of the LSH transformation other than the degree of LSH.

\subsection{Remarks of theoretical analysis}
Performance preservation is mandatory for CB transformations. Based on the distance correlation, such transformations can be summarised into four scenarios as illustrated in Fig. \ref{Figure::theory}, Fig. \ref{Figure::distanceco} and Table~\ref{table::theory}. 

Fig. \ref{Figure::theory}a indicates a perfect secure transformation function. Typical realizations are cryptography hashing, such as MD5 and SHA256. However, this scenario does not preserve performance at all due to complete mixed mated and non-mated samples. 

The scenario in Fig. \ref{Figure::theory}b suggests low information leakage attributed to the non-linear distance correlation. On the other hand, the high separation of intra-user and inter-user samples in the transformed space implies enhanced performance over the original one. Representative examples are data-dependence (learning-based) CB schemes such as DSQ \cite{chen-SPDSQ-2019}. However, data-dependence CB schemes are vulnerable to inversion attacks \cite{liu2020privacy}. 

Fig. \ref{Figure::theory}c indicates non-linear data-independent transformations such as NMDSH \cite{dong-IWBF-2019}. Such a transformation offers decent performance preservation. However, the pre-image attack risk remains high due to its distance-preserving property.

Fig. \ref{Figure::theory}d preserves the distance between features optimally due to linear distance correlation. However, this transformation has the worst information leakage and, thus, the highest pre-image attack risk. 

Unifying existing CB schemes under a general distance preservation framework is challenging. However, we highlight that most CB schemes comply with LSH as it guarantees performance preservation. However, as indicated by Proposition \ref{prop.lshattack}, the LSH-based transformation makes the CB schemes vulnerable to the pre-image attack. 

Though all scenarios from Fig. \ref{Figure::theory}b-d comply with the LSH, the information leakage level differs for each scenario and is weakly related to the pre-image attack risk. Indeed, the information leakage is more related to locality and isometric distance preserving. For example, Fig. \ref{Figure::theory}b achieves the lowest degree of locality preserving and isometric distance preserving, Fig. \ref{Figure::theory}c shows higher, and Fig. \ref{Figure::theory}d achieves the best with respect to the degree of locality preserving and isometric distance preserving. Based on the Proposition \ref{prop.locality_leakage} and Proposition \ref{prop.isoleakage}, Fig. \ref{Figure::theory} b,c, and d suffer from the lowest, high, and highest information leakage, respectively.

In a nutshell, we provide a detailed analysis of information leakage from three distance preservation notions, i.e., LSH, locality preserving, and isometric distance preserving. This new perspective may inspire CB scheme design. Remarkably, the information leakage of each CB scheme should be analyzed through the locality and isometric distance preservation. Besides that, keeping the accuracy performance by distance preservation inevitably astray the CB schemes vulnerable to pre-image attack.

\section{Estimation of the information leakage \label{section.infoleakage}}
By measuring the entropy, the amount of information in the original feature $X$ (attacker’s initial uncertainty), the amount of information leaked to the protected template $Y$, and the amount of unleaked information about $X$ (attacker’s remaining uncertainty) can be computed \cite{smith2009foundations}. Subsequently, the mutual information $I(X,Y)$ between $X$ and $Y$ can be exploited to evaluate the information learned from a known $Y$. The information leakage at the feature level is theoretically sound. However, due to the difficulty of estimating the entropy of continuous biometric features, it is intractable to calculate the mutual information directly at the feature level.

Fortunately, Proposition \ref{prop.leakage_translate} suggests that the information leakage $I(X,Y)$ estimation from features can be relaxed to $I(S,T)$ estimation. Here $S$ is the distance between features, and $T$ is the distance between templates.
Therefore, we propose to estimate the information leakage from the distances statistics.

Let $S$ be a random distance ranged in a set $\{s_1, \dots, s_I\}$ of quantized distances between features. Similarly, $T$ is ranged in quantized distances between templates $\{t_1, \dots, t_J\}$. 
The information leakage is defined by
\begin{equation}
\begin{split}
I(S;T) =\sum_{i=1}^{I} \sum_{j=1}^{J} \bm{D}(j|i) q(i) \log \frac{\bm{\Phi}(i|j)}{\bm{q}(i)}. \label{eq.mutual_convex}
\end{split}
\end{equation}
where $\Phi(i|j) = \Prob(S = s_i | T=t_j)$ and and $q(i) = \Prob(S=s_i)$. The transition matrix $\bm{D}$ is defined by $\bm{D}(j|i) = \Prob(T=t_j | S=s_i)$.

For any given CB scheme with fixed $\bm{D}(j|i)$, (\ref{eq.mutual_convex}) is a convex function of the input distribution $\bm{q}$, hence there exists a maximum leakage ($ML$) defined as \cite{blahut1972computation,arimoto1972algorithm}: 
\begin{equation}
\begin{split}
\lambda_{max} \triangleq \max_{\bm{q}} I(\mathcal{S};\mathcal{T}).
\end{split}
\end{equation}

To solve the above problem, we adopt the Blahut–Arimoto algorithm \cite{blahut1972computation,arimoto1972algorithm} to compute the information-theoretic maximum leakage of a transformation function. Blahut–Arimoto algorithm and the evaluation details are presented in Appendix Section \ref{appx.quantify}.

\section{Experimental results\label{section.experiment}}
In this section, we first demonstrate the proposed pre-image attack against several state-of-art CB schemes.
In this step, $Sys~C$ and $Sys~T$ use the same CB scheme/same transformation function, but with different parameters, hence we refer to this case as the \emph{single-transformation attack}.
We first launch the attack with single and multiple compromised templates.
Then the leakage and the security level of the considered state-of-art CB schemes are evaluated.
Next we validate the performance of the attack in the case of a cross-transformation function, i.e., reconstruct the feature vector from $Sys~C$ with one CB scheme, and attack $Sys~T$ with another type of CB scheme; we refer to this case as the \emph{cross-transformation attack}.

\begin{table*}
\centering
\caption{Setups of the cancelable biometric schemes.\label{table::cancelablesexperiemnt}}
\begin{tabular}{lllll} 
\toprule
Method & Dataset & Biometric trait & Feature size & CB template size \\ 
\midrule
BioHashing & LFW & Deep facial feature & 512 real-valued vector & 16–500~bits \\
Spectral hashing & LFW & Deep facial feature & 512 real-valued vector & 10–256~bits \\
IoM hashing & LFW & Deep facial feature & 512 real-valued vector & 16–500~bits \\
Bloom Filter & Casia-V4 Interval & IrisCode & 10240~bits & $2^6*2^{8,9,10}$~bits \\
IFO & Casia-V4 Interval & IrisCode & 10240~bits & 200–800~bits \\
2PMCC & FVC2002 DB2 & Fingerprint~MCC$_16$ & 16*16*5 real-valued~number for each~minutiae point & $k=16,32,64,128$~bits \\
\bottomrule
\end{tabular}
\vspace{-0.5cm}
\end{table*}

To explore the capability of the pre-image attack on BioHashing, IoM hashing, and MDSH, the Live Faces in the Wild (LFW) \cite{huang-2008-LFW} face dataset has been exploited to compute the score distribution. The 158 subjects with more than ten images were selected, and the first ten images were chosen to build a new subset (LFW10). The facial features were extracted by the InsightFace (ArcFace) \cite{deng2018arcface} deep convolutional neural network. To determine the capability of pre-image attacks on Bloom Filtering and IFO hashing schemes, the algorithm described in \cite{rathgeb2013bloomfilteriris,gomez2016unlinkablebloomfilter} was applied to the left eye images in the CASIA-v4-Iris-Interval\footnote{http://www.cbsr.ia.ac.cn/china/Iris Databases CH.asp} to generate the IrisCodes \cite{daugman2006iriscode}. Furthermore, the fingerprint dataset FVC2002 DB2-A \cite{maltoni2009handbookfvc} is used for 2PMCC. The default parameters of MCC SDK\footnote{http://biolab.csr.unibo.it/mccsdk.html} have been adopted, i.e., $R=75,N_S=16$, and $N_D=5$. 

To initiate the GA, the population size in the GA was set to 200 individuals and the crossover fraction to 0.9, while the mutation ratio was set to 0.01. The mutation is performed by adding a random number taken from a Gaussian distribution with a mean of 0 to each entry of the parent vector. In this experiment, each user's feature vector is reconstructed based on varying numbers of compromised templates (\# of templates) in $Sys~C$. Then the pre-image generates the template with the transformation function in $Sys~T$. Finally, the newly generated template is compared with the real template of the same person in $Sys~T$ to cause the \Stress{mated-imposter score}, where the successful attack rate (SAR) is calculated as the proportion of impostors that are falsely declared to match the template of the same user at the given normal EER threshold. Here, SAR can be regarded as the false match rate under attack, denoted by $FMR_{attack}$, and a higher SAR implies that the scheme is more vulnerable to attack and vice versa. The setup of each CB scheme is presented in Table~\ref{table::cancelablesexperiemnt}.

\subsection{Single-transformation attack}

\subsubsection{Attack on Biohashing and IoM}
The effectiveness of the pre-image attack on BioHashing templates $f(\bm{x})$ with different bit sizes $l$ is first evaluated. The EER and FMR of a BioHashing system under normal conditions are collected. The system \textbf{T}hreshold $\theta$ is determined and fixed concerning the \textbf{E}ER, denoted as \Stress{ET}. Next, the pre-image attack is launched with a compromised template from $Sys~C$. The pre-image $f(\bm{\widehat{x}})$ is compared with $f(\bm{x})$ in $Sys~T$ to generate the \Stress{mated-imposter scores}. The \Stress{mated} and \Stress{non-mated} scores are generated by comparing templates from the same user and different users in a normal situation. Th FMR at a normal system threshold $\theta$ concerning the EER, denoted as FMR@ET, is computed. The attack SAR at a normal system threshold $\theta$ in $Sys~C$ is also computed, and so is $\Delta$ FMR = SAR - FMR@ET (seeTable~\ref{table::biohashingresult}). 

\begin{table}[t!]
\centering
\caption{Evaluation of the pre-image attack on BioHashing and IoM (\%). \label{table::biohashingresult}}
\resizebox{.99\linewidth}{!}{
\begin{tabular}{cccccccccc} 
\toprule
& \multicolumn{4}{c}{Biohashing} & \multicolumn{4}{c}{IoM} \\
& \multicolumn{2}{c}{Normal} & \multicolumn{2}{c}{Attack} & \multicolumn{2}{c}{Normal} & \multicolumn{2}{c}{Attack} \\
$l$ & EER & FMR@ET & SAR & $\Delta$ FMR & EER & FMR@ET & SAR & $\Delta$ FMR \\
\cmidrule(lr){1-1}\cmidrule(lr){2-3}\cmidrule(lr){4-5}\cmidrule(lr){6-7}\cmidrule(lr){8-9}
16 & 19.95 & 24.12 & 28.28 & 4.16 & 22.67 & 23.88 & 22.94 & -0.94 \\
32 & 12.76 & 12.48 & 14.22 & 1.74 & 16.37 & 9.64 & 10.44 & 0.8 \\
64 & 7.80 & 7.11 & 11.35 & 4.24 & 10.74 & 9.5 & 11.8 & 2.3 \\
100 & 6.56 & 6.81 & 14.33 & 7.52 & 8.64 & 9.27 & 14.37 & 5.1 \\
200 & 5.49 & 5.43 & 28.78 & 23.35 & 6.49 & 6.76 & 16.33 & 9.57 \\
300 & 5.34 & 5.18 & 51.62 & 46.44 & 5.59 & 5.3 & 20.04 & 14.74 \\
400 & 5.48 & 5.57 & 72.20 & 66.63 & 5.55 & 5.4 & 29.06 & 23.66 \\
500 & 5.29 & 5.22 & 85.54 & 80.32 & 5.57 & 5.66 & 41.01 & 35.35 \\
\bottomrule
\end{tabular}
}
\vspace{-0.5cm}
\end{table}

Under normal conditions, as the length of the BioHash code becomes longer, the accuracy performance in terms of EER and FMR@ET becomes better (See the column `Normal' in Table~\ref{table::biohashingresult}). However, the risk from a pre-image attack shows a different result. A more extended BioHash code will lead to higher SAR (See the column `Attack' in Table~\ref{table::biohashingresult}). This is not surprising, as a longer BioHash code suggests the presence of more information, thus leading to better accuracy performance. However, more information implies lower attack complexity.

On the other hand, smaller $l$ may provide more robust resistance to the pre-image attack but will lead to a higher risk with the false accept attack (see section 26.6.1.1 in \cite{boulgouris2009biometrics}), which can break the system by exploiting authentication attempts without knowing the algorithm. In the scenario of the false-accept attack, assuming that the attacker can generate a sufficient number of biometric samples, the attacker may try to pass the authentication by simulating authentication attempts. The attack's success rate depends on the system's FMR, which means $\frac{1}{FMR}$ authentication attempts can be expected to break the system. As a result, it suggests thata higher false match rate accompanies a short hash code; the best performance achieved is 5.22\% of FMR@ET with 500 bits of Biohashing hash code. The results suggest that both long and short BioHash codes are not recommended. 

In reality, the attacker can likely break more than one system. Hence, we consider a case that allows the attacker to initiate the attack with multiple compromised templates. Multiple compromised templates ranging from 1 to 5 in $Sys~C$ are selected for each user to launch the attack. Intuitively, multiple instances of compromised templates provide more information for the attack to construct a better pre-image, and therefore, our result in Table~\ref{table::biohashingresultmulti} does support our hypothesis. Specifically, SAR increases from 80\% to 95\% when the number of compromised templates increases from 1 to 5 for Biohashing.

\begin{table*}[t!]
\centering
\caption{Evaluation of the pre-image attack with multiple compromised templates (\%). \label{table::biohashingresultmulti}}
\begin{tabular}{c ccc ccc ccc ccc} 
\toprule
\multirow{3}{*}{\# of templates} & \multicolumn{3}{c}{Biohashing ($l=500$)} & \multicolumn{3}{c}{NMDSH ($l=500$)} & \multicolumn{3}{c}{IoM ($l=500$)} & \multicolumn{3}{c}{2PMCC ($k=64$)} \\
& Normal & \multicolumn{2}{c}{Attack} & Normal & \multicolumn{2}{c}{Attack} & Normal & \multicolumn{2}{c}{Attack} & Normal & \multicolumn{2}{c}{Attack} \\
& FMR@ET & SAR &$\Delta$FMR & FMR@ET & SAR &$\Delta$FMR & FMR@ET & SAR &$\Delta$FMR & FMR@ET & SAR &$\Delta$FMR \\ 
\cmidrule(lr){1-1}\cmidrule(lr){2-4}\cmidrule(lr){5-7}\cmidrule(lr){8-10}\cmidrule(lr){11-13}
1 & \multirow{5}{*}{5.21} & 80.25 & 75.04 & \multirow{5}{*}{5.69} & 23.27 & 17.58 & \multirow{5}{*}{5.57} & 34.43 & 28.86 & \multirow{5}{*}{3.13} & 28.87 & 25.74 \\
2 & & 91.14 & 85.93 & & 24.99 & 19.3 & & 52.34 & 46.77 & & - & - \\
3 & & 94.30 & 89.09 & & 30.31 & 24.74 & & 55.32 & 49.74 & & 56.5 & 53.37 \\
4 & & 95.95 & 90.74 & & 32.71 & 27.02 & & 58.54 & 52.97 & & - & - \\
5 & & 94.37 & 89.15 & & 30.81 & 25.12 & & 62.66 & 57.09 & & 77.75 & 74.62 \\
\bottomrule
\end{tabular}
\end{table*}

\begin{figure*}[t!]
\setlength{\abovecaptionskip}{8pt} 
\setlength{\belowcaptionskip}{-14pt} 
\begin{center}
\imgStubPDFpage{0.24}{page=21,width=1\linewidth,trim=10.6cm 5cm 10.6cm 5cm,clip}{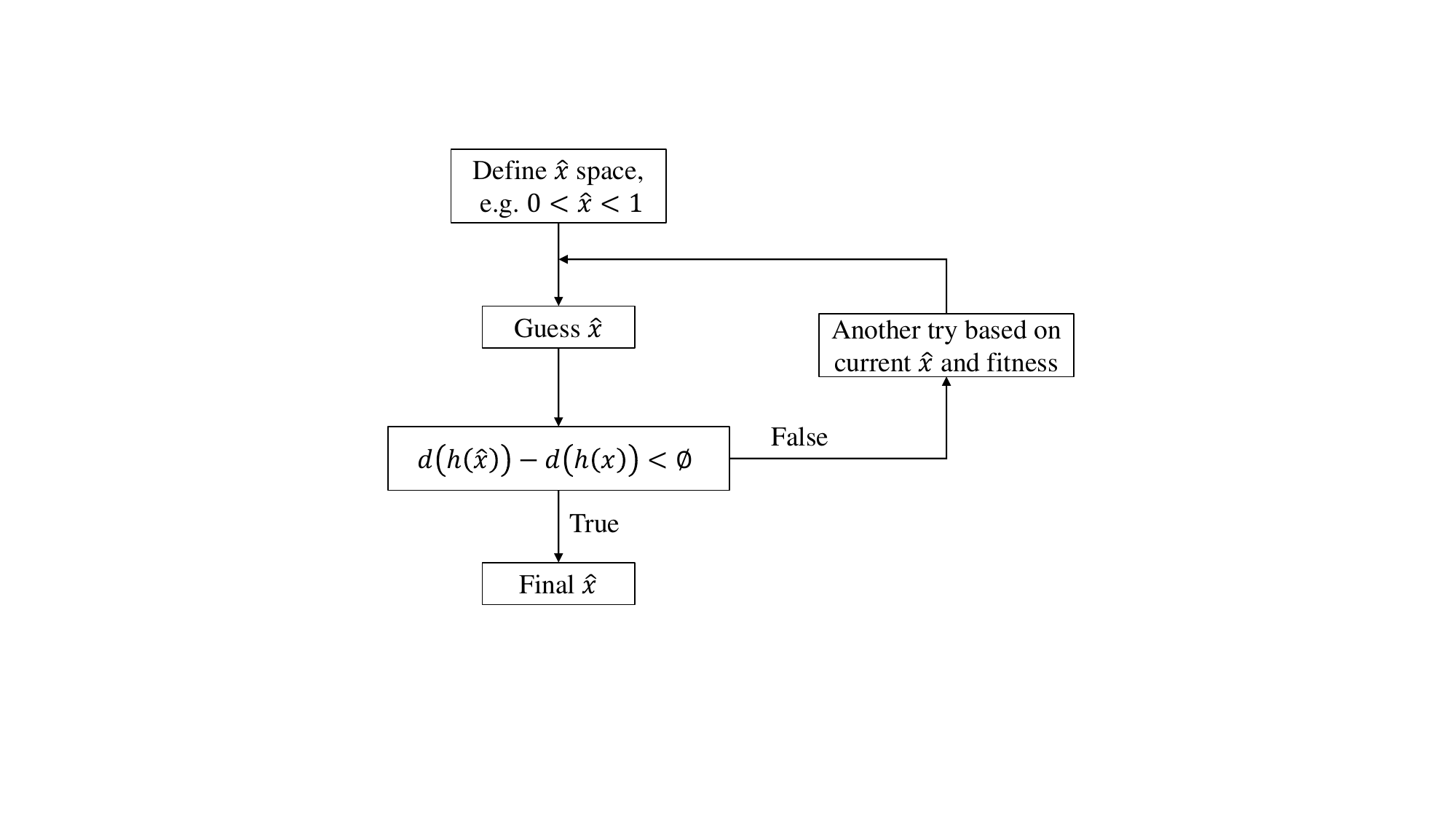}{(a) BioHashing} 
\imgStubPDFpage{0.24}{page=20,width=1\linewidth,trim=10.6cm 5cm 10.6cm 5cm,clip}{pdf/diagram.pdf}{(b) IoM Hashing}
\imgStubPDFpage{0.24}{page=22,width=1\linewidth,trim=10.6cm 5cm 10.6cm 5cm,clip}{pdf/diagram.pdf}{(c) NMDSH}
\imgStubPDFpage{0.24}{page=35,width=1\linewidth,trim=10.6cm 5cm 10.6cm 5cm,clip}{pdf/diagram.pdf}{(d) IFO Hashing}
\caption{Maximum leakage and accuracy performance against hash code length. Information leakage and accuracy are inversely correlated to the hash code length in BioHashing, NMDSH, IoM and IFO hashing. \label{Fig.maxleakagegroup}}
\end{center}
\vspace{-1.0em}
\end{figure*}

IoM hashing is slightly different from BioHashing, and its core idea is to project the feature vectors onto a subspace by ranking. As for IoM hashing, we also perform an evaluation based on the same protocol as we did for BioHashing. In our experiment, the parameter $q$ of the IoM hashing is fixed at 16, while $l$ ranges from 16 to 500. 

As shown in Table~\ref{table::biohashingresult}, IoM hashing also offers similar characteristics to those observed for BioHashing. Specifically, when $l$ increases, SAR increases significantly. The simulation results for multiple compromised templates in Table~\ref{table::biohashingresultmulti} suggest that the availability of more compromised instances will lead to higher risks. 

As discussed in section \ref{section.infoleakage}, we measure $\lambda_{max}$, i.e., the maximum leakage, from the distance correlation of a CB transformation function against bit length $l$ to explore their relationship. From Fig.~\ref{Fig.maxleakagegroup}(a), we observe that the leakage and accuracy performance increase with $l$ for Biohashing. This shows that good accuracy does not imply good security. As shown in Fig.~\ref{Fig.maxleakagegroup} (b), IoM hashing are similar to Biohashing in terms of leakage and accuracy performance.

\subsubsection{Attack on nonlinear multi-dimensional spectral hashing}
The NMDSH consists of spectral hashing,and ``softmod'' function inspired from \cite{chen-SPDSQ-2019}. The EER and FMR are evaluated for NMDSH based on the same protocol as in the BioHashing evaluation, except we set $\alpha=\{0.1,0.3,0.5,0.7,0.9\}$ in (\ref{eq::mdshsoftmod}). Lastly, SAR and $\Delta FMR$ are presented (Table~\ref{mdshlength}). 

The result of NMDSH shows a similar trend as that of BioHashing, i.e., a longer hash code results in more information leakage (Table~\ref{mdshlength}). For example, as $l$ increases from 10 to 256, $\Delta$ FMR increases significantly from 1.35 to 63.7. This is reasonable because a longer hash code is supposed to capture more information. However, if higher accuracy is required, then $l$ should be longer, yet this contradicts security demand. 
The attack performance under different distortion rates $\alpha$ is also tabulated in Table~\ref{mdshlength}. It is observed that more vital distortion can weaken the attack performance as $\Delta$ FMR drops from 63.7 to 3.47 when $\alpha$ increases from 0.1 to 0.9. 

\begin{table}[t!]
\centering
\caption{Evaluation of the pre-image attack on NMDSH under different $l$ (\%).\label{mdshlength}}
\begin{tabular}{ccccccc} 
\toprule
& & & \multicolumn{2}{c}{Normal} & \multicolumn{2}{c}{Attack} \\
$\alpha$ & $l$ & $\theta$ & EER & FMR@ET & SAR & $\Delta$ FMR \\ 
\cmidrule(lr){1-3}\cmidrule(lr){4-5}\cmidrule(lr){6-7}
0.10 & 10 & 0.60 & 22.02 & 17.80 & 19.15 & 1.35 \\
0.10 & 50 & 0.60 & 8.15 & 11.68 & 16.33 & 4.65 \\
0.10 & 128 & 0.58 & 5.70 & 5.60 & 22.28 & 16.68 \\
0.10 & 256 & 0.56 & 5.43 & 5.41 & 69.11 & 63.7 \\
\hline
0.30 & 256 & 0.56 & 5.41 & 5.36 & 65.44 & 60.08 \\
0.50 & 256 & 0.55 & 6.35 & 6.35 & 33.42 & 27.07 \\
0.70 & 256 & 0.53 & 17.33 & 17.80 & 20.06 & 2.98 \\
0.90 & 256 & 0.52 & 24.93 & 24.50 & 27.97 & 3.47 \\
\bottomrule
\end{tabular}
\end{table}

More instances of the compromised template are proven to increase the attack performance, as shown in Table~\ref{table::biohashingresultmulti}. For example, the SAR increases from 23.27 to 32.71 when the number of compromised templates increases from 1 to 4. 

The results, which include $\lambda_{max}$ and EER of NMDSH, are shown in Fig.~\ref{Fig.maxleakagegroup}(c). The leakage and the accuracy performance offer similar characteristics to those for Biohashing, consistent with the results obtained for the above attack experiments.

\subsubsection{Attack on IFO hashing}
To evaluate the accuracy of the generated IrisCode, the left eye images in CASIA-v4-Iris-Interval were used in our experiments. To standardize the analysis, the matching protocol in \cite{lai2017cancelable} was adopted for both IFO hashing and Bloom filter hashing. Specifically, a subject with at least seven samples is selected, forming 868 (124 subjects * 7 samples) iris samples. To generate the mated similarity scores, each iris template is matched to the other iris templates of the same person ( ${7 \choose 2} *124=2604$ mated comparisons). In contrast, non-mated similarity scores are generated by comparing the template with all other templates from different iris samples of different people (${124 \choose 2} *7*7=373674$). The EER accuracy of the IrisCode is computed based on the Hamming distances during matching. 
The original EER of IrisCode is 0.38\% with shifted $\pm$ 16 bits.

This paper implements the IFO hashing as described in~\cite{lai2017cancelable}. In the IFO scheme, the four principal parameters are the window size $K$, the number of permutations $m$, $P$ number of random generated permutation vectors, and security threshold $\tau$. To evaluate the attack success rate on IFO, we fixed $K=100$, $P=3$, and $\tau=0$, and then launched the attack with $m=\{200,400,800\}$. As shown in Table \ref{table::iforesult}, IFO is also vulnerable to attack. The SAR increases from 65.99\% to 81.45\% as $m$ increases to 800. As shown in Fig.~\ref{Fig.maxleakagegroup}(d), there is also a trade-off between the leakage $\lambda_{max}$ and the EER of IFO. 



\begin{table}
\centering
\caption{Evaluation of the pre-image attack on IFO (\%) with $P=3$, $\tau=0$. }
\label{table::iforesult}
\begin{tabular}{cccccc}
\toprule
\multicolumn{1}{c}{} & \multicolumn{3}{c}{Normal} & \multicolumn{2}{c}{Attack} \\
$m$ & $\theta$ & EER & FMR & SAR & $\Delta$ FMR \\ 
\cmidrule(lr){1-2}\cmidrule(lr){3-4}\cmidrule(lr){5-6}
200 & 0.06 & 10.97 & 10.45 & 65.99 & 55.54 \\ 
400 & 0.06 & 10.19 & 10.04 & 81.32 & 71.27 \\ 
800 & 0.06 & 9.83 & 9.56 & 81.45 & 71.89 \\
\bottomrule
\end{tabular}
\vspace{-0.5cm}
\end{table}

\subsubsection{Attack on the Bloom Filter}
This section discusses the implementation and evaluation of an improved version of the Bloom filter reported in~\cite{gomez2016unlinkablebloomfilter}. Compared with its ancestor \cite{rathgeb2013bloomfilteriris}, a feature permutation is employed in the improved Bloom filter, which can achieve unlinkability. Note that the Bloom filter requires two parameters and keys: the word size $\omega$ and the block size $l$, the xor key, and the permutation key (perm). In the present paper, the word size $w$ is fixed and equal to 10, while the parameter $l$ ranges from 4 to $6\left(2^{4}, 2^{5}, 2^{6}\right)$. Since xor operations and permutation keys are used in Bloom filtering, three settings are considered to evaluate the attack performance. Specifically, in $Sys~C$, we assume that all users possess the common $x$ or key (global key) and the common application-wise permutation key. After reconstructing the pre-images from $Sys~C$, they are exploited for attempting to access $Sys~T$ with three different settings: i) the same xor and permutation keys as in $Sys~C$; ii) the same xor key as in $Sys~C$ but with a different permutation key, and iii) a different xor and a different permutation key. The same attacks were also performed when there were multiple compromised templates. The results are presented in Table~\ref{table::bfresult}.

As can be observed from Table~\ref{table::bfresult}, the performance of the Bloom filtering degrades when $Sys~T$ employs the same xor and permutation keys as in $Sys~C$. The results presented with the same xor and permutation keys also demonstrate that having multiple compromised templates can help attackers to increase the attack success rate. However, the system can significantly resist the attack with a new permutation key and xor key. The SAR can only achieve $8.33\%$ and $8.6\%$ when $l=4$. This suggests that security can be increased by adding an extra permutation key and xor key. However, extra multiple keys requires the effort of storing those keys securely. 

\begin{table}
\centering
\caption{Evaluation of pre-image attack on the Bloom filter with $\omega=10$.}
\label{table::bfresult}
\resizebox{.95\linewidth}{!}{
\begin{tabular}{ccccccccl} 
\toprule
& \multicolumn{4}{c}{Normal} & \multicolumn{3}{c}{Attack} & \multicolumn{1}{c}{\multirow{3}{*}{$\lambda_{max}$}} \\ 
\multirow{2}{*}{$l$} & \multirow{2}{*}{\begin{tabular}[c]{@{}c@{}}\# of \\tamp.\end{tabular}} & \multirow{2}{*}{$\theta$} & \multirow{2}{*}{EER} & \multirow{2}{*}{\begin{tabular}[c]{@{}c@{}}FMR\\@ET\end{tabular}} & \begin{tabular}[c]{@{}c@{}}Same xor\\Same perm\end{tabular} & \begin{tabular}[c]{@{}c@{}}Same xor\\New perm\end{tabular} & \begin{tabular}[c]{@{}c@{}}New xor\\New perm\end{tabular} & \multicolumn{1}{c}{} \\
& & & & & \multicolumn{3}{c}{SAR} & \multicolumn{1}{c}{} \\ 
\cmidrule(lr){1-2}\cmidrule(lr){3-5}\cmidrule(lr){6-8}\cmidrule(lr){9-9}
\multirow{5}{*}{4} & 1 & \multirow{5}{*}{0.16} & \multirow{5}{*}{11.95} & \multirow{5}{*}{12.82} & 51.71 & 8.33 & 8.60 & \multirow{5}{*}{3.12} \\
& 2 & & & & 62.62 & 5.32 & 7.90 & \\
& 3 & & & & 67.39 & 10.08 & 11.49 & \\
& 4 & & & & 71.64 & 10.48 & 9.14 & \\
& 5 & & & & 71.27 & 7.66 & 9.27 & \\ 
\cmidrule(lr){1-2}\cmidrule(lr){3-5}\cmidrule(lr){6-8}\cmidrule(lr){9-9}
\multirow{5}{*}{5} & 1 & \multirow{5}{*}{0.23} & \multirow{5}{*}{13.42} & \multirow{5}{*}{13.27} & 46.98 & 5.65 & 7.26 & \multirow{5}{*}{3.07~~} \\
& 2 & & & & 56.90 & 4.19 & 6.29 & \\
& 3 & & & & 60.03 & 7.06 & 9.68 & \\
& 4 & & & & 64.92 & 5.91 & 8.33 & \\
& 5 & & & & 63.91 & 8.06 & 6.85 & \\ 
\cmidrule(lr){1-2}\cmidrule(lr){3-5}\cmidrule(lr){6-8}\cmidrule(lr){9-9}
\multirow{5}{*}{6} & 1 & \multirow{5}{*}{0.35} & \multirow{5}{*}{14.77} & \multirow{5}{*}{15.46} & 36.36 & 0.13 & 1.48 & \multirow{5}{*}{2.88} \\
& 2 & & & & 41.21 & 0.00 & 0.16 & \\
& 3 & & & & 47.03 & 0.00 & 0.00 & \\
& 4 & & & & 50.74 & 0.27 & 0.81 & \\
& 5 & & & & 50.81 & 0.40 & 0.40 & \\
\bottomrule
\end{tabular}}
\vspace{-0.5cm}
\end{table}

The results presented for $\lambda_{\max }$ also agree with the computed SAR in the pre-image attack scenario. As shown in Table~\ref{table::bfresult}, the leakage decreases from 3.12 to 2.88 as $l$ increases from 4 to $6$. This corresponds to a drop of the SAR from 8.6 to 1.48 when different xor and permutation keys are used, and the attacker holds only one compromised template. It is noteworthy that generally, the SAR is less than the FMR@ET, which implies that the false acceptance constitutes a higher risk than the proposed attack.

\subsubsection{Attack on 2PMCC}
There are two parameters in 2PMCC, namely, $k$ for the KL-transform and $c$ for the dimension reduction. In our experiment, for simplicity we assume $c=k$, which means there is no dimension reduction in 2PMCC, and $k \in \{16, 32, 64, 128\}$. We also assume that the secret $s$ in 2PMCC is set to an identical global parameter, which corresponds to the worstcase of a stolen parameter. An open-source minutiae extractor, FingerJetFX OSE \, cite{fingerjetfx2011openFingerJetFX}, has been used to extract the ISO minutiae templates for all fingerprints. FVC2002 DB2-A has been used for the simulation. FVC2002 DB2-A consists of 100 users with eight samples per user. The original accuracy performance in EER is calculated with the official protocol\footnote{http://bias.csr.unibo.it/fvc2002/protocol.asp}.

For the pre-image attack simulation, we set the generated fingerprint image to a square with a width and height of 350 with 50 minutiae points because the fingerprints in FVC2002 DB2-A have an average of 52 minutiae points. Under a pre-image attack, the first fingerprint template of each user is used as the compromised template to reconstruct the original minutiae points. Then, an additional two or four templates are used for the multiple compromised templates scenario to imitate the attack. The fingerprint minutiae pre-image generates a 2PMCC template with the secret in the target system; then, the pre-image is compared with their corresponding target templates. Hence $8 \times 100 = 800$ matching scores are collected for the computation of the SAR. 

Table~\ref{table::2pmccresult} tabulates the attack performance on 2PMCC with one compromised template for different $k$ and $c$. It is evidenced that the attack can achieve 25.6\% for $k=c=128$ and the original EER=2.05\%. We find that SAR under pre-image attack does not increase w.r.t. $k$ and $c$. However, $\Delta$FMR remains stable.
In Table~\ref{table::2pmccresult} we can also see that the maximum leakage decreases as $k,c$ increases. This is consistent with the SAR under pre-image attack and proves that the proposed measurement of the information leakage is reasonable. 

\begin{table}[t!]
\centering
\caption{Evaluation of the pre-image attack on 2PMCC (\%). \label{table::2pmccresult}}
\begin{tabular}{ccccccc} 
\toprule
& \multicolumn{3}{c}{Normal} & \multicolumn{2}{c}{Attack} & \multirow{2}{*}{$\lambda_{max}$} \\
$k,c$ & \begin{tabular}[c]{@{}c@{}}Threshold\\ $\theta$\end{tabular} & EER & \begin{tabular}[c]{@{}c@{}}FMR\\ @ET\end{tabular} & SAR & $\Delta $ FMR & \\
\cmidrule(lr){1-1}\cmidrule(lr){2-4}\cmidrule(lr){5-6}\cmidrule(lr){7-7}
16 & 0.86 & 15.16 & 15.39 & 40.50 & 25.11 & 3.02 \\
32 & 0.77 & 6.74 & 7.05 & 32.00 & 24.95 & 2.88 \\
64 & 0.70 & 3.11 & 3.09 & 28.37 & 25.28 & 2.62 \\
128 & 0.65 & 2.05 & 1.93 & 25.62 & 23.69 & 2.61 \\
\bottomrule
\end{tabular}
\vspace{-0.5cm}
\end{table}

The attack with multiple compromised templates is shown in Table~\ref{table::biohashingresultmulti}. Similar to other CB schemes, the results suggest that having multiple compromised templates can help to achieve better performance. For example, the SAR increases from 28.87\% to 77.75\% when the number of templates increases from 1 to 5.


\subsection{Cross-transformation attack\label{section.cross}}
The experiment above is conducted on a specific transformation function, while in reality, different systems may use different transformation functions.
In other words, theprevious experiment assumes that $ Sys~C$ and $Sys~T$ use the same CB schemes. Hence the $f(\cdot)$ and $h(\cdot)$ are from the same CB algorithm families.

As discussed in section \ref{section.attacktheory}, the attack success rate is correlated to the distance-preservation property regardless of template renewal. To validate this conclusion empirically, we assume a more complicated situation where the transformation function in $Sys~C$ (e.g., BioHashing) is different from that of $Sys~T$ (e.g., IoM hashing).
Thus, $f(\cdot)$ and $h(\cdot)$ are from different CB algorithm families. 
Again, we assume the attacker knows the parameters and the transformation function. 
We want to address the question, ``Can the attacker gain access to the $Sys~T$ with a \textit{cross-transformation} attack?''
In this section, we demonstrate the \textit{cross-transformation} attack on the above-discussed transformation functions on the face and iris datasets. Note that this section does not adopt fingerprint modality as limited open-sourced cancellable biometric schemes can be obtained, and various feature formats are used by difference cancelable schemes.

This subsection generates the face pre-images from different $Sys~C$ protected by BioHashing, IoM, and NMDSH. Then the pre-images are used to attempt $Sys~T$ with different transformation functions. For example, given a transformation function BioHashing and its parameters/keys in $Sys~C$, a feature vector is first reconstructed from the compromised template. Then $Sys~T$ is attempted to evaluate the attack performance, while the $Sys~T$ uses IoM to protect the biometric features. 

In this experiment, the length $l$ of BioHashing, IoM, NMDSH is set to 500, while the $\alpha$ of NMDSH is set to 0.5. Table~\ref{table::crossattack} tabulates the SAR of the \textit{cross-transformation} attack. The result suggests that even when $Sys~C$ and $Sys~T$ use different transformation functions, it is still possible to attain a large SAR, which leads to a high-security risk for the system. We also found that NMDSH shows better resistance against the attack when $\alpha$ is set to 0.5.

\begin{table}[t!]
\centering
\caption{The SAR (\%) under \textit{cross-transformation} attack on face ($l=500$).\label{table::crossattack}}
\resizebox{\linewidth}{!}{%
\begin{tabular}{ccccc}
\toprule
& & \multicolumn{3}{c}{Compromised Sys C} \\ 
& & Biohashing & IoM hashing& NMDSH$_{0.5}$ \\ 
\midrule
\multirow{3}{*}{Target Sys T} & Biohashing & 84.05 & 32.96 & 8.39 \\
& IoM hashing& 43.87 & 24.53 & 8.54 \\ 
& NMDSH$_{0.5}$ & 32.33 & 20.37 & 23.27 \\
\bottomrule
\end{tabular}
}
\end{table}

\begin{table}[t!]
\centering
\caption{The SAR (\%) under cross-transformation attack on IrisCode.}
\label{table::crossattackiriscode}
\begin{tabular}{cccc} 
\toprule
& & \multicolumn{2}{c}{Compromised Sys C} \\ 
& & IFO & Bloom filter \\ 
\midrule
\multirow{2}{*}{Target Sys T} & IFO & 81.32 & 0.26 \\ 
& Bloom filter & 18.68 & 8.60 \\
\bottomrule
\end{tabular}
\vspace{-0.5cm}
\end{table}

\begin{table*}
\centering
\caption{Summary of the single-transformation attack on various cancelable schemes.\label{table::summary}}
\label{table::summary}
\begin{tabular}{cllccc} 
\toprule
Modality & Features & CB schemes & \begin{tabular}[c]{@{}c@{}}Original\\ EER\end{tabular} & \begin{tabular}[c]{@{}c@{}}Attack\\ SAR\end{tabular} & Time cost \\ 
\midrule
Face & InsightFace~deep features~on LFW10 & BioHashing ($l=500$) & 5.29\% & 85.54\% & 17 s \\
Face & InsightFace~deep features~on LFW10 & IoM hashing ($l=500$) & 5.57\% & 41.01\% & 14 s \\
Face & InsightFace~deep features~on LFW10 & NMDSH ($\alpha=0.1,l=250$) & 5.43\% & 69.11\% & 13mins \\
Fingerprint & MCC (minutiae points) on FVC2002~DB2 & 2PMCC ($k=c=64$) & 3.11\% & 28.37\% & 21mins \\
Iris & IrisCode on~Casia-V4-Interval & BloomFilter ($l=2^4,\omega=10$) & 11.95\% & 8.6\% & 33mins \\
Iris & IrisCode on~Casia-V4-Interval & IFO ($m=800, \tau=50$) & 9.83\% & 81.45\% & 9min \\
\bottomrule
\end{tabular}
\vspace{-0.5cm}
\end{table*}

As with the cross-transformation attack on faces, the IrisCode is also reconstructed from different $Sys~C$ protected by the IFO and Bloom filter. Then the reconstructed IrisCode is used to attempt $Sys~T$ with different transformation functions. The result is shown in Table \ref{table::crossattackiriscode}.

It can be found that the IrisCode reconstructed from IFO hashing can easily bypass any $Sys~T$ protected by the Bloom filter or IFO, as the SAR reaches 18.68\% and 81.32\%, respectively. This proves that IFO is not strong at concealing the information in the template. However, the upgraded Bloom filter shows strong resistance to the proposed attack, as the SAR only achieves 0.26\% and 8.60\% in IFO or a Bloom filter protected $Sys~T$, respectively.

\section{Discussion and Conclusion\label{section.discussion}}
Table~\ref{table::summary} provided a summary of the performance of an attack on different cancelable biometric (CB) schemes. From the results, we can conclude that most CB schemes suffer from the distance preserving property. It was also worth highlighting that the attack can achieve good performance with the \textit{cross-transformation} attack with the same feature extractor.

The success of the pre-image attack is attributed to the distance preserving property from the matcher of the CB scheme. By incorporating the Blahut–Arimoto algorithm, we proposed an estimation method for information leakage. Our quantification results also proved that the proposed attack could breach the security of CB schemes.

While most CB schemes showed good accuracy performance, the analysis and results suggested otherwise. Specifically, our results showed that higher accuracy might lead to a severe security breach, and also note that lower accuracy did not necessarily imply better security. The security level of a given CB system should consider multiple factors, such as the false match rate and the information leakage via single and multiple compromised templates.

The pre-image attack requires the parameters of the transformation function to be known. Also, it requires a massive number of query access to the target CB transformation function (see section \ref{section.attack} and \ref{sec.timecostappendix}). However, massive querying makes the pre-image attack less realistic because each query access may be costly in practice. Therefore, limiting the number of attempts per user session and securely storing the parameters \cite{murakami2019cancelable,takahashi2011parameter} is essential to mitigate the attack. 

On the other hand, our analysis is based on the strong assumption that a high attack success rate can be achieved for the initial CB transformation function. Thus, one weak CB system in practice will lead to high attack risks for all CB systems regardless of template renewal. In conclusion, the attack risk and information leakage presented in this paper remain an open problem for CB schemes based on distance preserving. We highlight that CB algorithms may need to work together with other protection mechanisms to achieve a secure authentication system. 
\section*{Acknowledgement}
This work was partially funded by the University of Sassari Fondo di Ateneo per la Ricerca 2020 and 2021; in part by the Italian Ministry for Research Special Research Project SPADA.

\bibliographystyle{IEEEtran} 
\bibliography{IEEEabrv,sample_edited}

\begin{thebibliography}{10}
\expandafter\ifx\csname url\endcsname\relax
  \def\url#1{\texttt{#1}}\fi
\expandafter\ifx\csname urlprefix\endcsname\relax\def\urlprefix{URL }\fi
\expandafter\ifx\csname href\endcsname\relax
  \def\href#1#2{#2} \def\path#1{#1}\fi

\bibitem{patel_cancelable_2015-BTPOVERVIEW}
V.~M. Patel, N.~K. Ratha, R.~Chellappa, Cancelable {Biometrics}: {A} review,
  IEEE Signal Processing Magazine 32~(5) (2015) 54--65.
\newblock \href {https://doi.org/10.1109/MSP.2015.2434151}
  {\path{doi:10.1109/MSP.2015.2434151}}.

\bibitem{sandhya_biometric_2017-BTPOVERVIEW}
M.~Sandhya, M.~V. Prasad, Biometric template protection: A systematic
  literature review of approaches and modalities, in: Biometric Security and
  Privacy, Springer-Verlag, 2017, pp. 323--370.

\bibitem{chandra_cancelable_2011-BTPOVERVIEW}
E.~Chandra, K.~Kanagalakshmi, Cancelable biometric template generation and
  protection schemes: {A} review, in: 2011 3rd {International} {Conference} on
  {Electronics} {Computer} {Technology}, Vol.~5, 2011, pp. 15--20.
\newblock \href {https://doi.org/10.1109/ICECTECH.2011.5941948}
  {\path{doi:10.1109/ICECTECH.2011.5941948}}.

\bibitem{nandakumar_biometric_2015-BTPOVERVIEW}
K.~Nandakumar, A.~K. Jain, Biometric {Template} {Protection}: {Bridging} the
  performance gap between theory and practice, IEEE Signal Processing Magazine
  32~(5) (2015) 88--100.
\newblock \href {https://doi.org/10.1109/MSP.2015.2427849}
  {\path{doi:10.1109/MSP.2015.2427849}}.

\bibitem{ISOIEC24745}
{ISO/IEC24745:2011:Information technology — Security techniques — Biometric
  information protection}, Standard, International Organization for
  Standardization (Mar. 2011).

\bibitem{ISOIEC30136}
{ISO/IEC30136:2018(E): Information technology —Performance testing of
  biometric template protection schemes}, Standard, International Organization
  for Standardization (Mar. 2018).

\bibitem{gomez2017generalunlink}
M.~Gomez-Barrero, J.~Galbally, C.~Rathgeb, C.~Busch, General framework to
  evaluate unlinkability in biometric template protection systems, IEEE
  Transactions on Information Forensics and Security 13~(6) (2017) 1406--1420.

\bibitem{dong-BTAS-2019}
X.~Dong, Z.~Jin, A.~T.~B. Jin, A genetic algorithm enabled similarity-based
  attack on cancellable biometrics, in: 10th IEEE International Conference on
  Biometrics: Theory, Applications and Systems (BTAS), IEEE, 2019.

\bibitem{teoh_random_2006}
A.~B.~J. Teoh, A.~Goh, D.~C.~L. Ngo, Random {Multispace} {Quantization} as an
  {Analytic} {Mechanism} for {BioHashing} of {Biometric} and {Random}
  {Identity} {Inputs}, IEEE Transactions on Pattern Analysis and Machine
  Intelligence 28~(12) (2006) 1892--1901.
\newblock \href {https://doi.org/10.1109/TPAMI.2006.250}
  {\path{doi:10.1109/TPAMI.2006.250}}.

\bibitem{rathgeb2013bloomfilteriris}
C.~Rathgeb, F.~Breitinger, C.~Busch, Alignment-free cancelable iris biometric
  templates based on adaptive bloom filters, in: 2013 International Conference
  on Biometrics (ICB), IEEE, 2013, pp. 1--8.

\bibitem{hermans2014bloombecomesdoom}
J.~Hermans, B.~Mennink, R.~Peeters, When a {B}loom filter is a doom filter:
  Security assessment of a novel iris biometric template protection system, in:
  2014 International Conference of the Biometrics Special Interest Group
  (BIOSIG), IEEE, 2014, pp. 1--6.

\bibitem{gomez2016unlinkablebloomfilter}
M.~Gomez-Barrero, C.~Rathgeb, J.~Galbally, C.~Busch, J.~Fierrez, Unlinkable and
  irreversible biometric template protection based on {B}loom filters,
  Information Sciences 370 (2016) 18--32.

\bibitem{lai2017cancelable}
Y.-L. Lai, Z.~Jin, A.~B.~J. Teoh, B.-M. Goi, W.-S. Yap, T.-Y. Chai, C.~Rathgeb,
  Cancellable iris template generation based on {Indexing-First-O}ne hashing,
  Pattern Recognition 64 (2017) 105--117.

\bibitem{broder2000minhashing}
A.~Z. Broder, M.~Charikar, A.~M. Frieze, M.~Mitzenmacher, Min-wise independent
  permutations, Journal of Computer and System Sciences 60~(3) (2000) 630--659.

\bibitem{jin2017ranking}
Z.~Jin, J.~Y. Hwang, Y.-L. Lai, S.~Kim, A.~B.~J. Teoh, Ranking-based locality
  sensitive hashing-enabled cancelable biometrics: Index-of-max hashing, IEEE
  Transactions on Information Forensics and Security 13~(2) (2017) 393--407.

\bibitem{dong-IWBF-2019}
X.~Dong, K.~Wong, Z.~Jin, J.-l. Dugelay, A cancellable face template scheme
  based on nonlinear multi-dimension spectral hashing, in: 2019 7th
  International Workshop on Biometrics and Forensics (IWBF), IEEE, 2019, pp.
  1--6.

\bibitem{jin-PRL-2014}
Z.~Jin, M.-H. Lim, A.~B.~J. Teoh, B.-M. Goi, A non-invertible {Randomized}
  {Graph}-based {Hamming} {Embedding} for generating cancelable fingerprint
  template, Pattern Recognition Letters 42 (2014) 137--147.
\newblock \href {https://doi.org/10.1016/j.patrec.2014.02.011}
  {\path{doi:10.1016/j.patrec.2014.02.011}}.

\bibitem{ferrara20142pmcc}
M.~Ferrara, D.~Maltoni, R.~Cappelli, A two-factor protection scheme for {MCC}
  fingerprint templates, in: 2014 International Conference of the Biometrics
  Special Interest Group (BIOSIG), IEEE, 2014, pp. 1--8.

\bibitem{ferrara2012noninvertiblepmcc}
M.~Ferrara, D.~Maltoni, R.~Cappelli, Noninvertible minutia cylinder-code
  representation, IEEE Transactions on Information Forensics and Security 7~(6)
  (2012) 1727--1737.

\bibitem{cappelli2010minutiamcc}
R.~Cappelli, M.~Ferrara, D.~Maltoni, Minutia cylinder-code: A new
  representation and matching technique for fingerprint recognition, IEEE
  Transactions on Pattern Analysis and Machine Intelligence 32~(12) (2010)
  2128--2141.

\bibitem{fukunaga1993statistical}
K.~Fukunaga, Statistical pattern recognition, in: Handbook of Pattern
  Recognition and Computer Vision, World Scientific, 1993, pp. 33--60.

\bibitem{feng2010fingerprintreconstruct}
J.~Feng, A.~K. Jain, Fingerprint reconstruction: From minutiae to phase, IEEE
  Transactions on Pattern Analysis and Machine Intelligence 33~(2) (2010)
  209--223.

\bibitem{ratha2007fingerprint1}
N.~K. Ratha, S.~Chikkerur, J.~H. Connell, R.~M. Bolle, Generating cancelable
  fingerprint templates, IEEE Transactions on Pattern Analysis and Machine
  Intelligence 29~(4) (2007) 561--572.

\bibitem{ratha2006fingerprint2}
N.~Ratha, J.~Connell, R.~M. Bolle, S.~Chikkerur, Cancelable biometrics: A case
  study in fingerprints, in: 18th International Conference on Pattern
  Recognition (ICPR'06), Vol.~4, IEEE, 2006, pp. 370--373.

\bibitem{quan2008cracking}
F.~Quan, S.~Fei, C.~Anni, Z.~Feifei, Cracking cancelable fingerprint template
  of {R}atha, in: 2008 International Symposium on Computer Science and
  Computational Technology, Vol.~2, IEEE, 2008, pp. 572--575.

\bibitem{lacharme2013preimagebiohashing}
P.~Lacharme, E.~Cherrier, C.~Rosenberger, Preimage attack on biohashing, in:
  2013 International Conference on Security and Cryptography (SECRYPT), IEEE,
  2013, pp. 1--8.

\bibitem{lee2009inverse}
Y.~Lee, Y.~Chung, K.~Moon, Inverse operation and preimage attack on biohashing,
  in: 2009 IEEE Workshop on Computational Intelligence in Biometrics: Theory,
  Algorithms, and Applications, IEEE, 2009, pp. 92--97.

\bibitem{pagnin2014leakage}
E.~Pagnin, C.~Dimitrakakis, A.~Abidin, A.~Mitrokotsa, On the leakage of
  information in biometric authentication, in: International Conference in
  Cryptology in India, Springer-Verlag, 2014, pp. 265--280.

\bibitem{hattori2012provablyHattori}
M.~Hattori, N.~Matsuda, T.~Ito, Y.~Shibata, K.~Takashima, T.~Yoneda,
  Provably-secure cancelable biometrics using 2-{DNF} evaluation, Information
  and Media Technologies 7~(2) (2012) 749--760.

\bibitem{izu2014spoofing}
T.~Izu, Y.~Sakemi, M.~Takenaka, N.~Torii, A spoofing attack against a
  cancelable biometric authentication scheme, in: 2014 IEEE 28th International
  Conference on Advanced Information Networking and Applications, IEEE, 2014,
  pp. 234--239.

\bibitem{feng2014masquerade22222}
Y.~C. Feng, P.~C. Yuen, Vulnerabilities in binary face template, in: 2012 IEEE
  Computer Society Conference on Computer Vision and Pattern Recognition
  Workshops, IEEE, 2012, pp. 105--110.

\bibitem{feng2014masquerade}
Y.~C. Feng, M.-H. Lim, P.~C. Yuen, Masquerade attack on transform-based
  binary-template protection based on perceptron learning, Pattern Recognition
  47~(9) (2014) 3019--3033.

\bibitem{atighehchi2019cryptanalysis}
L.~{Ghammam}, K.~{Karabina}, P.~{Lacharme}, K.~{Atighehchi}, A cryptanalysis of
  two cancelable biometric schemes based on {Index-of-M}ax hashing, IEEE
  Transactions on Information Forensics and Security 15 (2020) 2869--2880.

\bibitem{chen-SPDSQ-2019}
Y.~Chen, Y.~Wo, R.~Xie, C.~Wu, G.~Han, Deep {Secure} {Quantization}: {On}
  secure biometric hashing against similarity-based attacks, Signal Processing
  154 (2019) 314--323.

\bibitem{smith2009foundations}
G.~Smith, On the foundations of quantitative information flow, in:
  International Conference on Foundations of Software Science and Computational
  Structures, Springer-Verlag, 2009, pp. 288--302.

\bibitem{blahut1972computation}
R.~Blahut, Computation of channel capacity and rate-distortion functions, IEEE
  Transactions on Information Theory 18~(4) (1972) 460--473.

\bibitem{arimoto1972algorithm}
S.~Arimoto, An algorithm for computing the capacity of arbitrary discrete
  memoryless channels, IEEE Transactions on Information Theory 18~(1) (1972)
  14--20.

\bibitem{huang-2008-LFW}
G.~B. Huang, M.~Mattar, T.~Berg, E.~Learned-Miller, Labeled faces in the wild:
  A database forstudying face recognition in unconstrained environments, in:
  Workshop on Faces in 'Real-Life' Images: Detection, alignment, and
  recognition, 2008.

\bibitem{deng2018arcface}
J.~Deng, J.~Guo, N.~Xue, S.~Zafeiriou, Arcface: Additive angular margin loss
  for deep face recognition, arXiv preprint arXiv:1801.07698 (2018).

\bibitem{daugman2006iriscode}
J.~Daugman, Probing the uniqueness and randomness of {IrisCodes}: Results from
  200 billion iris pair comparisons, Proceedings of the IEEE 94~(11) (2006)
  1927--1935.

\bibitem{maltoni2009handbookfvc}
D.~Maltoni, D.~Maio, A.~K. Jain, S.~Prabhakar, Handbook of Fingerprint
  Recognition, Springer-Verlag, 2009.

\bibitem{boulgouris2009biometrics}
N.~V. Boulgouris, K.~N. Plataniotis, E.~Micheli-Tzanakou, Biometrics: Theory,
  Methods, and Applications, Vol.~9, Wiley, 2009.

\bibitem{fingerjetfx2011openFingerJetFX}
D.~Inc., Fingerjetfx, open source edition,
  http://digitalpersona.com/fingerjetfx/ (2011).

\bibitem{murakami2019cancelable}
T.~Murakami, R.~Fujita, T.~Ohki, Y.~Kaga, M.~Fujio, K.~Takahashi, Cancelable
  permutation-based indexing for secure and efficient biometric identification,
  IEEE Access 7 (2019) 45563--45582.

\bibitem{takahashi2011parameter}
K.~Takahashi, S.~Hirata, Parameter management schemes for cancelable
  biometrics, in: 2011 IEEE Workshop on Computational Intelligence in
  Biometrics and Identity Management (CIBIM), IEEE, 2011, pp. 145--151.

\end{thebibliography}


\begin{thebibliography}{10}
\providecommand{\url}[1]{#1}
\csname url@samestyle\endcsname
\providecommand{\newblock}{\relax}
\providecommand{\bibinfo}[2]{#2}
\providecommand{\BIBentrySTDinterwordspacing}{\spaceskip=0pt\relax}
\providecommand{\BIBentryALTinterwordstretchfactor}{4}
\providecommand{\BIBentryALTinterwordspacing}{\spaceskip=\fontdimen2\font plus
\BIBentryALTinterwordstretchfactor\fontdimen3\font minus
  \fontdimen4\font\relax}
\providecommand{\BIBforeignlanguage}[2]{{%
\expandafter\ifx\csname l@#1\endcsname\relax
\typeout{** WARNING: IEEEtran.bst: No hyphenation pattern has been}%
\typeout{** loaded for the language `#1'. Using the pattern for}%
\typeout{** the default language instead.}%
\else
\language=\csname l@#1\endcsname
\fi
#2}}
\providecommand{\BIBdecl}{\relax}
\BIBdecl

\bibitem{patel_cancelable_2015-BTPOVERVIEW}
V.~M. Patel, N.~K. Ratha, and R.~Chellappa, ``Cancelable {Biometrics}: {A}
  review,'' \emph{IEEE Signal Processing Magazine}, vol.~32, no.~5, pp. 54--65,
  Sep. 2015.

\bibitem{sandhya_biometric_2017-BTPOVERVIEW}
M.~Sandhya and M.~V. Prasad, ``Biometric template protection: A systematic
  literature review of approaches and modalities,'' in \emph{Biometric Security
  and Privacy}.\hskip 1em plus 0.5em minus 0.4em\relax Springer-Verlag, 2017,
  pp. 323--370.

\bibitem{chandra_cancelable_2011-BTPOVERVIEW}
E.~Chandra and K.~Kanagalakshmi, ``Cancelable biometric template generation and
  protection schemes: {A} review,'' in \emph{2011 3rd {International}
  {Conference} on {Electronics} {Computer} {Technology}}, vol.~5, Apr. 2011,
  pp. 15--20.

\bibitem{nandakumar_biometric_2015-BTPOVERVIEW}
K.~Nandakumar and A.~K. Jain, ``Biometric {Template} {Protection}: {Bridging}
  the performance gap between theory and practice,'' \emph{IEEE Signal
  Processing Magazine}, vol.~32, no.~5, pp. 88--100, Sep. 2015.

\bibitem{ISOIEC24745}
``{ISO/IEC24745:2011:Information technology — Security techniques —
  Biometric information protection},'' International Organization for
  Standardization, Standard, Mar. 2011.

\bibitem{ISOIEC30136}
``{ISO/IEC30136:2018(E): Information technology —Performance testing of
  biometric template protection schemes},'' International Organization for
  Standardization, Standard, Mar. 2018.

\bibitem{dong-BTAS-2019}
X.~Dong, Z.~Jin, and A.~T.~B. Jin, ``A genetic algorithm enabled
  similarity-based attack on cancellable biometrics,'' in \emph{10th IEEE
  International Conference on Biometrics: Theory, Applications and Systems
  (BTAS)}.\hskip 1em plus 0.5em minus 0.4em\relax IEEE, 2019.

\bibitem{atighehchi2019cryptanalysis}
L.~{Ghammam}, K.~{Karabina}, P.~{Lacharme}, and K.~{Atighehchi}, ``A
  cryptanalysis of two cancelable biometric schemes based on {Index-of-M}ax
  hashing,'' \emph{IEEE Transactions on Information Forensics and Security},
  vol.~15, pp. 2869--2880, 2020.

\bibitem{feng2014masquerade22222}
Y.~C. Feng and P.~C. Yuen, ``Vulnerabilities in binary face template,'' in
  \emph{2012 IEEE Computer Society Conference on Computer Vision and Pattern
  Recognition Workshops}.\hskip 1em plus 0.5em minus 0.4em\relax IEEE, 2012,
  pp. 105--110.

\bibitem{feng2014masquerade}
Y.~C. Feng, M.-H. Lim, and P.~C. Yuen, ``Masquerade attack on transform-based
  binary-template protection based on perceptron learning,'' \emph{Pattern
  Recognition}, vol.~47, no.~9, pp. 3019--3033, 2014.

\bibitem{teoh_random_2006}
A.~B.~J. Teoh, A.~Goh, and D.~C.~L. Ngo, ``Random {Multispace} {Quantization}
  as an {Analytic} {Mechanism} for {BioHashing} of {Biometric} and {Random}
  {Identity} {Inputs},'' \emph{IEEE Transactions on Pattern Analysis and
  Machine Intelligence}, vol.~28, no.~12, pp. 1892--1901, Dec. 2006.

\bibitem{rathgeb2013bloomfilteriris}
C.~Rathgeb, F.~Breitinger, and C.~Busch, ``Alignment-free cancelable iris
  biometric templates based on adaptive bloom filters,'' in \emph{2013
  International Conference on Biometrics (ICB)}.\hskip 1em plus 0.5em minus
  0.4em\relax IEEE, 2013, pp. 1--8.

\bibitem{hermans2014bloombecomesdoom}
J.~Hermans, B.~Mennink, and R.~Peeters, ``When a {B}loom filter is a doom
  filter: Security assessment of a novel iris biometric template protection
  system,'' in \emph{2014 International Conference of the Biometrics Special
  Interest Group (BIOSIG)}.\hskip 1em plus 0.5em minus 0.4em\relax IEEE, 2014,
  pp. 1--6.

\bibitem{gomez2016unlinkablebloomfilter}
M.~Gomez-Barrero, C.~Rathgeb, J.~Galbally, C.~Busch, and J.~Fierrez,
  ``Unlinkable and irreversible biometric template protection based on {B}loom
  filters,'' \emph{Information Sciences}, vol. 370, pp. 18--32, 2016.

\bibitem{lai2017cancelable}
Y.-L. Lai, Z.~Jin, A.~B.~J. Teoh, B.-M. Goi, W.-S. Yap, T.-Y. Chai, and
  C.~Rathgeb, ``Cancellable iris template generation based on
  {Indexing-First-O}ne hashing,'' \emph{Pattern Recognition}, vol.~64, pp.
  105--117, 2017.

\bibitem{broder2000minhashing}
A.~Z. Broder, M.~Charikar, A.~M. Frieze, and M.~Mitzenmacher, ``Min-wise
  independent permutations,'' \emph{Journal of Computer and System Sciences},
  vol.~60, no.~3, pp. 630--659, 2000.

\bibitem{jin2017ranking}
Z.~Jin, J.~Y. Hwang, Y.-L. Lai, S.~Kim, and A.~B.~J. Teoh, ``Ranking-based
  locality sensitive hashing-enabled cancelable biometrics: Index-of-max
  hashing,'' \emph{IEEE Transactions on Information Forensics and Security},
  vol.~13, no.~2, pp. 393--407, 2017.

\bibitem{dong-IWBF-2019}
X.~Dong, K.~Wong, Z.~Jin, and J.-l. Dugelay, ``A cancellable face template
  scheme based on nonlinear multi-dimension spectral hashing,'' in \emph{2019
  7th International Workshop on Biometrics and Forensics (IWBF)}.\hskip 1em
  plus 0.5em minus 0.4em\relax IEEE, 2019, pp. 1--6.

\bibitem{jin-PRL-2014}
Z.~Jin, M.-H. Lim, A.~B.~J. Teoh, and B.-M. Goi, ``A non-invertible
  {Randomized} {Graph}-based {Hamming} {Embedding} for generating cancelable
  fingerprint template,'' \emph{Pattern Recognition Letters}, vol.~42, pp.
  137--147, Jun. 2014.

\bibitem{ferrara20142pmcc}
M.~Ferrara, D.~Maltoni, and R.~Cappelli, ``A two-factor protection scheme for
  {MCC} fingerprint templates,'' in \emph{2014 International Conference of the
  Biometrics Special Interest Group (BIOSIG)}.\hskip 1em plus 0.5em minus
  0.4em\relax IEEE, 2014, pp. 1--8.

\bibitem{ferrara2012noninvertiblepmcc}
M.~Ferrara, D.~Maltoni, and R.~Cappelli, ``Noninvertible minutia cylinder-code
  representation,'' \emph{IEEE Transactions on Information Forensics and
  Security}, vol.~7, no.~6, pp. 1727--1737, 2012.

\bibitem{cappelli2010minutiamcc}
R.~Cappelli, M.~Ferrara, and D.~Maltoni, ``Minutia cylinder-code: A new
  representation and matching technique for fingerprint recognition,''
  \emph{IEEE Transactions on Pattern Analysis and Machine Intelligence},
  vol.~32, no.~12, pp. 2128--2141, 2010.

\bibitem{fukunaga1993statistical}
K.~Fukunaga, ``Statistical pattern recognition,'' in \emph{Handbook of Pattern
  Recognition and Computer Vision}.\hskip 1em plus 0.5em minus 0.4em\relax
  World Scientific, 1993, pp. 33--60.

\bibitem{feng2010fingerprintreconstruct}
J.~Feng and A.~K. Jain, ``Fingerprint reconstruction: From minutiae to phase,''
  \emph{IEEE Transactions on Pattern Analysis and Machine Intelligence},
  vol.~33, no.~2, pp. 209--223, 2010.

\bibitem{lacharme2013preimagebiohashing}
P.~Lacharme, E.~Cherrier, and C.~Rosenberger, ``Preimage attack on
  biohashing,'' in \emph{2013 International Conference on Security and
  Cryptography (SECRYPT)}.\hskip 1em plus 0.5em minus 0.4em\relax IEEE, 2013,
  pp. 1--8.

\bibitem{pagnin2014leakage}
E.~Pagnin, C.~Dimitrakakis, A.~Abidin, and A.~Mitrokotsa, ``On the leakage of
  information in biometric authentication,'' in \emph{International Conference
  in Cryptology in India}.\hskip 1em plus 0.5em minus 0.4em\relax
  Springer-Verlag, 2014, pp. 265--280.

\bibitem{chen-SPDSQ-2019}
Y.~Chen, Y.~Wo, R.~Xie, C.~Wu, and G.~Han, ``Deep {Secure} {Quantization}: {On}
  secure biometric hashing against similarity-based attacks,'' \emph{Signal
  Processing}, vol. 154, pp. 314--323, Jan. 2019.

\bibitem{liu2020privacy}
X.~Liu, L.~Xie, Y.~Wang, J.~Zou, J.~Xiong, Z.~Ying, and A.~V. Vasilakos,
  ``Privacy and security issues in deep learning: A survey,'' \emph{IEEE
  Access}, vol.~9, pp. 4566--4593, 2020.

\bibitem{chin1991complexity}
D.~A. Chin, ``Complexity issues in general purpose parallel computing,'' Ph.D.
  dissertation, University of Oxford, 1991.

\bibitem{chin1994locality}
A.~Chin, ``Locality-preserving hash functions for general purpose parallel
  computation,'' \emph{Algorithmica}, vol.~12, no.~2, pp. 170--181, 1994.

\bibitem{indyk1998approximate}
P.~Indyk and R.~Motwani, ``Approximate nearest neighbors: towards removing the
  curse of dimensionality,'' in \emph{Proceedings of the thirtieth annual ACM
  symposium on Theory of computing}, 1998, pp. 604--613.

\bibitem{smith2009foundations}
G.~Smith, ``On the foundations of quantitative information flow,'' in
  \emph{International Conference on Foundations of Software Science and
  Computational Structures}.\hskip 1em plus 0.5em minus 0.4em\relax
  Springer-Verlag, 2009, pp. 288--302.

\bibitem{blahut1972computation}
R.~Blahut, ``Computation of channel capacity and rate-distortion functions,''
  \emph{IEEE Transactions on Information Theory}, vol.~18, no.~4, pp. 460--473,
  1972.

\bibitem{arimoto1972algorithm}
S.~Arimoto, ``An algorithm for computing the capacity of arbitrary discrete
  memoryless channels,'' \emph{IEEE Transactions on Information Theory},
  vol.~18, no.~1, pp. 14--20, 1972.

\bibitem{huang-2008-LFW}
G.~B. Huang, M.~Mattar, T.~Berg, and E.~Learned-Miller, ``Labeled faces in the
  wild: A database forstudying face recognition in unconstrained
  environments,'' in \emph{Workshop on Faces in 'Real-Life' Images: Detection,
  alignment, and recognition}, 2008.

\bibitem{deng2018arcface}
J.~Deng, J.~Guo, N.~Xue, and S.~Zafeiriou, ``Arcface: Additive angular margin
  loss for deep face recognition,'' \emph{arXiv preprint arXiv:1801.07698},
  2018.

\bibitem{daugman2006iriscode}
J.~Daugman, ``Probing the uniqueness and randomness of {IrisCodes}: Results
  from 200 billion iris pair comparisons,'' \emph{Proceedings of the IEEE},
  vol.~94, no.~11, pp. 1927--1935, 2006.

\bibitem{maltoni2009handbookfvc}
D.~Maltoni, D.~Maio, A.~K. Jain, and S.~Prabhakar, \emph{Handbook of
  Fingerprint Recognition}.\hskip 1em plus 0.5em minus 0.4em\relax
  Springer-Verlag, 2009.

\bibitem{boulgouris2009biometrics}
N.~V. Boulgouris, K.~N. Plataniotis, and E.~Micheli-Tzanakou, \emph{Biometrics:
  Theory, Methods, and Applications}.\hskip 1em plus 0.5em minus 0.4em\relax
  Wiley, 2009, vol.~9.

\bibitem{murakami2019cancelable}
T.~Murakami, R.~Fujita, T.~Ohki, Y.~Kaga, M.~Fujio, and K.~Takahashi,
  ``Cancelable permutation-based indexing for secure and efficient biometric
  identification,'' \emph{IEEE Access}, vol.~7, pp. 45\,563--45\,582, 2019.

\bibitem{takahashi2011parameter}
K.~Takahashi and S.~Hirata, ``Parameter management schemes for cancelable
  biometrics,'' in \emph{2011 IEEE Workshop on Computational Intelligence in
  Biometrics and Identity Management (CIBIM)}.\hskip 1em plus 0.5em minus
  0.4em\relax IEEE, 2011, pp. 145--151.

\end{thebibliography}
\vfill

\clearpage

\setcounter{page}{1}
\setcounter{section}{0}
\setcounter{table}{0}
\setcounter{figure}{0}
\setcounter{prop}{0}
\setcounter{coro}{0}
\setcounter{lem}{0}
\setcounter{define}{0}
\setcounter{equation}{0}

\twocolumn[ 
 \begin{@twocolumnfalse}
\begin{center}
{\Huge{Appendix}}
 \end{center}
\end{@twocolumnfalse}
]

\renewcommand{\thefigure}{A\arabic{figure}} 
\renewcommand{\thetable}{A\arabic{table}} 
\renewcommand{\thealgocf}{A\arabic{algocf}}
\renewcommand{\thesection}{A\arabic{section}} 
\renewcommand{\thesubsection}{A\arabic{subsection}}

\section{Pre-image Attack algorithm based on GA \label{sec.gaappendix}}

\begin{figure}[t!]
\centering
\includegraphics[page=32,width=1\linewidth,trim=3cm 5cm 4cm 3cm, clip]{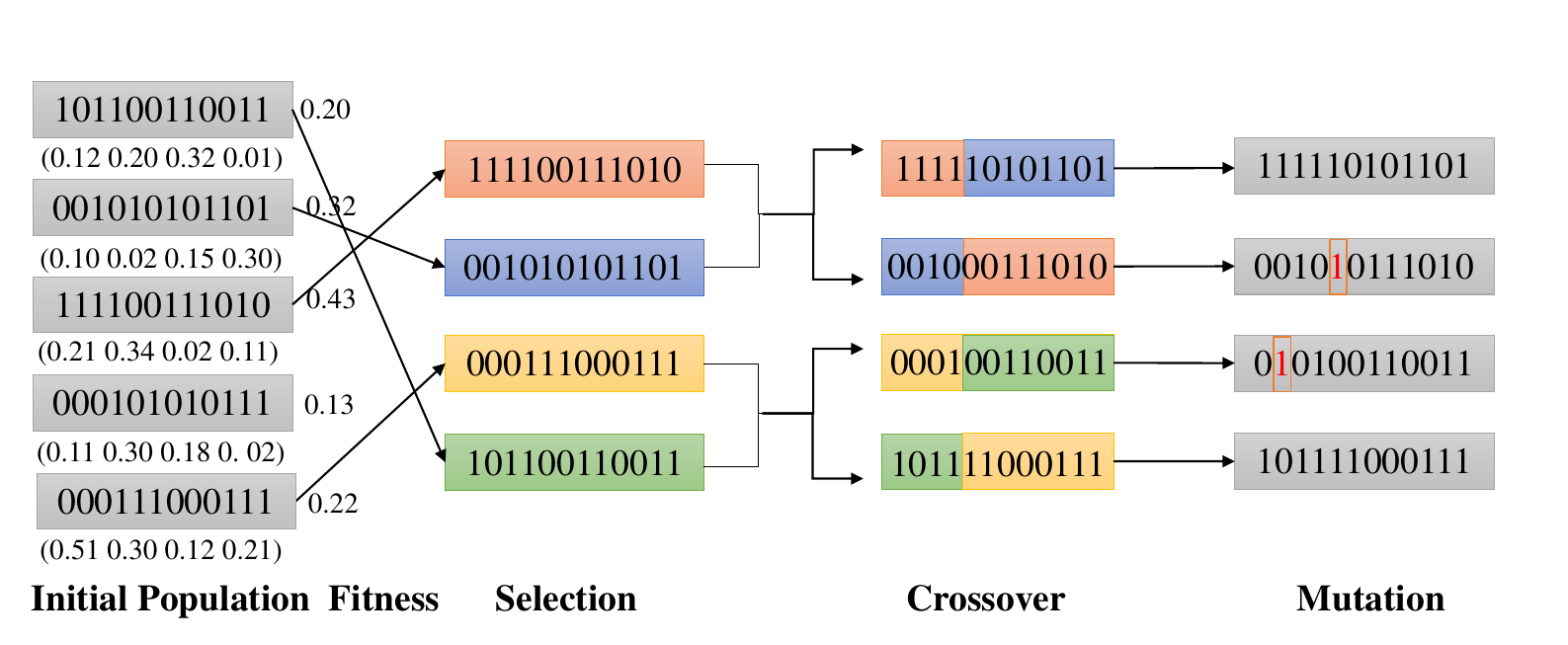} 
\caption{GA evolution process flow. The initial population has 5 individuals, the value in the bracket is the decoded floating number of the feature, 4 individuals are selected according to the rank of objective value, crossover is performed on the selected individuals, then followed by mutation to generate the next generation of the population. \label{Figure::gadetails}}
\end{figure}

The pre-image attacks exploit the information leakage from the distance preservation between the original biometric input feature space and the transformed feature space. Pre-image attacks aim to approximate the original template (pre-image) based on an initial guess of the template instance and the objective function. 

To find the best pre-image, a genetic algorithm (GA) is employed in the pre-image attack. The GA simulates the genetic mechanism of the biological world, including mutations, crossover, and selection. To find an optimal solution, a GA can drive the evolution of the population of solutions by repeatedly modifying individual solutions, i.e., generating children through selection, crossover, and mutations. The first population is initialized randomly in the defined solution space; then, these operations are used to generate the next generation. Finally, the objective function is used to evaluate how optimal the individual in the population is (see Fig. \ref{Figure::gadetails}). 
A GA repeats this process until the value of an objective function converges. The attack sequence is described in Algorithm \ref{algo.SimilaritybasedAttackGA}.

\begin{algorithm}[h]
\SetAlgoLined
\begin{flushleft}
\textbf{INPUT:} Cancelable transformation $f(\cdot)$, compromised template $f(\bm{x})$, objective function $l$\\
\textbf{OUTPUT:} Pre-image $\bm{\widehat{x}}$
\end{flushleft}
Encoding the solution space, i.e., the space of $\bm{\widehat{x}}$, \;
Initiate iteration index of generation $\ell=0$, constraint tolerance $\delta$, max generations $\epsilon$\;
Evaluate initial individuals by objective function (\ref{eq.objectivefinal})\;
\Repeat{$|l(\ell+1)-l(\ell)|<\delta$ or $\ell>\epsilon$}{
Selection of the best-fit individuals for the next generation\;
Generate a new child by crossover and mutation operations\;
Evaluate new individuals by the objective function $l$ \;
Replace the least-fit population with new individuals\;
$\ell=\ell+1$\;
}
\caption{Pre-image Attack based on GA}\label{algo.SimilaritybasedAttackGA}
\end{algorithm}

\section{Proof details for Section \ref{section.attacktheory}\label{section.attacktheoryappdix}}
\subsection{On Locality-Preservation}

\medskip
\noindent
\textbf{Definition.}
We begin by formalizing the definition of locality-preserving property in a general probabilistic framework. 
A transformation $h$ is \textbf{locality-preserving with a degree} $\epsilon$ if 
\begin{multline}
\Prob \left[
d( h(X), h(X_1)) < d(h(X), h(X_2)) \right.
\\
\left.
\mid
d(X, X_1) < d(X, X_2)
\right]
\geq 1 - \epsilon
\end{multline}
and 
\begin{multline}
\Prob \left[
d( X, X_1) \geq d(X, X_2) \right.
\\
\left.
\mid
d(h(X), h(X_1)) \geq d(h(X), h(X_2))
\right]
\geq 1 - \epsilon.
\end{multline}
The degree $\epsilon$ determines the degree of locality preservation of the transformation $h$. The lower the degree $\epsilon$ is, the more the transformation is locality-preserving.
With $\epsilon =0$, we attain the original definition of the (deterministic) locality preservation \cite{chin1991complexity,chin1994locality}
\begin{multline}
\label{eq::localitya}
d(X, X_1) < d(X, X_2)
\\
\implies
d(h(X), h(X_1)) < d(h(X), h(X_2)).
\end{multline}
In this respect, our probabilistic definition can be understood as an extension of the original one. A locality-preserving transformation preserves the relativistic order of distances. In the deterministic case, the distance between features is related to the distance between templates by a strictly increasing function (as depicted in Fig.~\ref{Figure::theory}bcd).
Hence, the transformation makes the recognition performance preserved in the template space, but unfortunately, it keeps the attacker's performance as well; in this sense, the success rate of the pre-image attack remains high even after template renewal. We formalize these claims below.

\subsubsection{Performance Preservation of Locality-Preserving Transformation}
\begin{prop}
Assume a high recognition performance on the features 
\begin{equation}
\Prob \left[
d(X, X_{in}) < d(X, X_{out})
\right]
\geq p
\end{equation}
with $p >0$
under the condition where the random features $X,X_{in}$ are within the same class $C_y$ while $X_{out} \notin C_y$ is an inter-class feature.
Then, for a locality-preserving transformation $h$ with a degree $\epsilon$,
\begin{equation}
\mathbb{P} \left[
d(h(X), h(X_{in})) < d(h(X), h(X_{out}))
\right] \geq (1-\epsilon)p
\end{equation}
conditioned on $X, X_{in} \in C_y$ and $X_{out} \notin C_y$.
\end{prop}

Here, $C_k \subseteq \mathcal{X}$ denotes the set of intra-class features corresponding to the class $k$.

\begin{proof}
Assume the condition $X, X_{in} \in C_y$ and $X_{out} \notin C_y$ here throughout. Let $A$ and $B$ denote the events
\begin{alignat}{2}
A & \equiv d(X, X_{in}) < d(X, X_{out})\\
B & \equiv d(h(X), h(X_{in})) < d(h(X), h(X_{out})).
\end{alignat}
By the law of total probability, 
\begin{alignat}{2}
\Prob(B) & = \Prob(B|A) \Prob(A) + \Prob(B|A^c) \Prob(A^c) \\
& \geq \Prob(B|A) \Prob(A) \\
& \geq p (1-\epsilon),
\end{alignat}
concluding the proof.
\end{proof}

In the above proposition, 
$
\Prob \left[
d(X, X_{in}) < d(X, X_{out})
\right]
$
indicates the recognition performance of the original features while
$\mathbb{P} \left[
d(h(X), h(X_{in})) < d(h(X), h(X_{out}))
\right]$ refers to that of the transformed features. The proposition states that a distance-preserving transformation preserves the performance based on the degree of distance-preservation.

\subsubsection{Pre-image Attack on Locally-Preserving Transformation. }
Unfortunately, the distance-preservation also preserves the attack success rate, regardless of template renewal, as shown by the following proposition.

\begin{prop}
\label{prop.cross}
Let $\mathcal{A}$ denote an attack algorithm that generates pre-image $X_a = \mathcal{A}(h(X))$ given $h(X)$ for a distance-preserving transformation $h$ with a degree $\epsilon$. Assume a high attack success rate 
\begin{equation}
\Prob \left[
d( h(X) , h(X_a) ) < d( h(X), h(X_{out}) )
\right] \geq p
\end{equation}
conditioned on $X \in C_y$ and $X_{out} \notin C_y$, where $\epsilon >0$ is small. Then, for any locality-preserving transformation $f$ with the degree of $\delta$, we have
\begin{equation}
\Prob \left[
d( f(X) , f(X_a) ) < d( f(X), f(X_{out}) )
\right] \geq (1-\epsilon) (1-\delta) p 
\end{equation}
conditioned on $X \in C_y$ and $X_{out} \notin C_y$.
\end{prop}

\begin{proof}
Assume we are conditioned on $X \in C_y$ and $X_{out} \notin C_y$.
Let $f$ be an arbitrary distance-preserving transformation with the probability $q$. Denote the events by
\begin{alignat}{2}
A & \equiv d(X, X_a) < d(X, X_{out}) \\
B & \equiv d(h(X), h(X_a)) < d(h(X), h(X_{out})) \\
C & \equiv d( f(X) , f(X_a) ) < d( f(X), f(X_{out}) ) 
\end{alignat}
and
\begin{alignat}{2}
A' & \equiv d(X, X_a) \leq d(X, X_{out}) \\
B' & \equiv d(h(X), h(X_a)) \leq d(h(X), h(X_{out})).
\end{alignat}
We note that $\Prob(A) = \Prob(A')$ since the probability of equality is zero in the space of continuum $\mathbb{R}^{\geq 0}$. Thus, we use this property freely. 

Our main objective in this proof is to show that $\Prob(C) \geq (1 - \epsilon)(1 - \delta) p$. To see this, observe 
\begin{alignat}{2}
\Prob(A)& = \Prob(A') \\
& \geq \Prob(A' | B') \Prob(B') \\
& \geq \Prob(A' | B') \Prob(B) \\
& \geq (1- \epsilon)p.
\end{alignat}
Here, the second inequality is due to the law of total probability,
the third inequality is due to $B \subseteq B'$,
and the fourth inequality holds by the locality-preservation of $h$. Then, further, observe that
\begin{alignat}{2}
P(C) & \geq P(C | A) P(A) \\
& \geq (1 - \delta) (1-\epsilon) p
\end{alignat}
where the first inequality is by the law of total probability while the second is due to the locality-preservation of $f$.
\end{proof}

The proposition states that, once we breach the system for the initial transformation $h$ with the pre-image $x_a$, the same pre-image can successfully attack the replaced system with a probability $(1 - \delta) (1-\epsilon) p$.

\subsubsection{Information Leakage of Locality-Preserving Transformation}

Preserving the distance relationship by locality enables one to infer the original feature from the transformed template. This is formally verified by the following.
\begin{prop}
\label{prop.locality_leakage}
Let $X$ be a random feature of discrete range with $n$ values. Assume $h$ is locality-preserving with a sufficiently small degree $\epsilon$ such that $1 - \epsilon > 1/e$ and $\epsilon < 1/e$. Then,
\begin{equation}
I(X, h(X)) \geq H(X) - \left[ 
\epsilon + 2(n-1)\frac{\sqrt{\epsilon}}{e}
\right]
\end{equation}
where the entropy $H(X)$ is a constant and a maximum of the leakage given a certain biometric modality; $H(X) \geq I(X,h(X))$.
\end{prop}

To show this, we need a technical lemma.
\begin{lem}
\label{lem.locality_leakage}
Fix $p > 0$ and $0 < \alpha <1$. Then, $- p \log p \leq \frac{p^\alpha}{e(1-\alpha)}$.
\end{lem}
\begin{proof}
Due to the convexity, we have $- p \log p < 1/e$ for any $p$. Replacing $p$ by $p^{1-\alpha}$, we obtain
\begin{equation}
-(1 - \alpha) p^{1-\alpha} \log p
= -p^{1-\alpha} \log (p^{1-\alpha}) \leq \frac{1}{e}.
\end{equation}
Since $0< \alpha < 1$, the desired follows by division by $p^\alpha$ and $1-\alpha$.
\end{proof}

\begin{proof}[Proof of Proposition \ref{prop.locality_leakage}]
With $Y= h(X)$, it suffices to find an upper bound of $H(X | Y)$, which is
\begin{equation}
H(X|Y) = 
\sum_{i=1}^n p(y_i) \sum_{j=1}^n -p(x_j|y_i) \log p(x_j|y_i)
\end{equation}
where $p(y) = \Prob(Y = y)$ and $p(x | y) = \Prob(X = x| Y = y)$. For fixed $y$ with $p(y) > 0$, choose $x$ such that $h(x) = y$ 
\begin{alignat}{2}
& p(x|y) \\
& = \Prob \left[ 
d(X, x) \leq 0 \mid d( h(X), y ) \leq 0 
\right] \\
& = \Prob \left[ 
d(X, x) \leq d(x,x) \mid d( h(X), h(x) ) \leq d( h(x), h(x))
\right] \\
& \geq 1 - \epsilon
\end{alignat}
due to the locality-preserving property of $h$. Then, for any $x' \neq x$, then $p(x'|y) < \epsilon$. Then, we can separate the summands of the conditional entropy as follows:
\begin{multline}
H(X|Y) \\
= \sum p(y) \left[
-p(x|y)\log p(x|y) + \sum_{x' \neq x} - p(x'|y) \log p(x'|y)
\right].
\end{multline}
Observe
\begin{alignat}{2}
-p(x|y) \log p(x|y) & \leq - (1-\epsilon) \log (1- \epsilon) \\
& \leq 1 - (1-\epsilon),
\end{alignat}
where the first inequality holds since $g(q) = -q \log q$ is a decreasing function of $q$ for $1 /e \leq q \leq 1$, and the second inequality holds due to the logarithmic inequality $1 - 1/q \leq \log q$. On the other hand, note that 
\begin{alignat}{2}
-p(x'|y) \log p(x'|y) & \leq - \epsilon \log \epsilon \\
& \leq 2 \sqrt{\epsilon}/e,
\end{alignat}
where the first inequality holds due to the increasing nature of $g(q) = -q \log q$ with $0 \leq q \leq 1/e$, and the second holds due to Lemma \ref{lem.locality_leakage} with $\alpha = 1/2$.
Therefore, the inner sum of $H(X|Y)$ is less than or equal to $\epsilon + (n-1)2\sqrt{\epsilon}/e$, completing the proof.
\end{proof}

Due to the lower bound of the leakage given in the above proposition, the leakage is increased if the transformation is more locality-preserving (namely, the higher the probability $p$). 

Under practical consideration, however, estimation of the information leaked from the features is intractable due to the curse of the high-dimensionality of the features. Fortunately, the mutual relationship between the feature $X$ and its template $h(X)$ can be equally represented by the relationship between their corresponding feature distance $S=d(X_1, X_2)$ and template distance $T=d(h(X_1), h(X_2))$. In particular, the degree $\epsilon$ of distance preservation is correspondent to the leakage $I(S,T)$ just as it is correspondent to the leakage $I(X, h(X))$:

\begin{prop}
\label{prop.leakage_translatea}
Let $X_1, X_2$ be discrete random features with at most $n$ values. Let $S=d(X_1, X_2)$ and $T = d(h(X_1), h(X_2))$ where $h$ is locality-preserving with degree $\epsilon$. Then, 
\begin{equation}
I(S,T) \geq H(S) - n^2 \left[
\epsilon + \frac{2(n^2-1) \sqrt{\epsilon}}{e}
\right]
\end{equation}
where $H(s)$ is a constant and a maximum of the leakage $I(S,T)$ given a certain biometric modality; $H(S) \geq I(S,T)$.
\end{prop}

\begin{proof}
Define $g$ such that $g(s) = d(h(x_1), h(x_2))$ for $s=d(x_1,x_2)$. Then, $T = g(S)$, and by the definition of locality-preserving property, we have
\begin{equation}
\Prob \left[ g(S) > g(s) \mid S > s \right] \geq 1-\epsilon 
\end{equation}
for any $s \in \mathbb{R}$. Note that
\begin{multline}
H(S|T) \\
=
\sum_t \Prob(T=t) 
\sum_s - \Prob(S =s | T= t) \log \Prob (S =s | T=t).
\end{multline}
For a fixed $t$, choose $s$ such that $g(s)=t$. Now we consider such $s$ and other $s' \neq s$. Observe
\begin{equation}
\Prob(S =s | T=t) 
= 1 - \Prob(S>s | T=t) - \Prob(S < s | T=t)
\end{equation}
where
\begin{equation}
\Prob (S > s | T =t) 
= 
\frac{
\Prob(T=t|S>s)\Prob(S>s)
}{
\Prob(T=t)
}
\end{equation}
by Bayes' rule. We can upper-bound this term since
\begin{equation}
\Prob(T=t|S>s) 
\leq 1 - \Prob(T>t|S>s) \leq \epsilon.
\end{equation} 
Namely, $\Prob(S>s|T=t) \leq \epsilon \Prob(S>s)/\Prob(T=t)$, and likewise for $\Prob(S<s|T=t)$. Hence, 
\begin{equation}
\Prob(S=s|T=t) \geq 
1 - \epsilon \frac{\Prob(S \neq s)}{\Prob(T=t)}
\end{equation}
and, therefore,
\begin{equation}
\Prob(S = s'|T=t) < \epsilon \frac{\Prob(S \neq s)}{\Prob(T=t)}. 
\end{equation}
Thus, if $\epsilon$ is sufficiently small, then
\begin{multline}
\sum_s -\Prob(S=s|T=t) \log \Prob(S=s|T=t)
\\
\leq 
\frac{
\epsilon + 2(n^2-1)\sqrt{\epsilon}/e
}
{
\Prob(T=t)
}
\end{multline}
due to logarithmic inequality, Lemma \ref{lem.locality_leakage}, and the fact that the function $-x \log x$ is an increasing function with $x$ close to 0 while being a decreasing function with $x$ close to 1.
Note that $n^2$ arises since there are at most $n^2$ number of summands in $\sum_s$ and $\sum_t$.
We obtain the desired proposition by substituting the obtained to $H(S|T)$.
\end{proof}

\subsection{On Isometric Distance-Preservation}

\medskip
\noindent
\textbf{Definition.}
The notion of distance-preserving transformation can be formalized in a more strict, geometric context. Namely,
the distance-preservation with a degree of $\epsilon$ can be geometrically characterized by\footnote{We slightly abuse the $\epsilon$, the $\epsilon$ here is different from section \ref{sec.onlocality} by definition.}
\begin{equation}
\label{eq::isometrya}
\epsilon = \inf \{ \epsilon : \lvert d( h(\bm{x}_1), h(\bm{x}_2) ) - d(\bm{x}_1, \bm{x}_2) \rvert < \epsilon, \forall \bm{x}_1, \bm{x}_2 \in \mathcal{X} \}.
\end{equation}
Following the Hausdorff approximation, we define \textbf{isometric distance-preserving} transformation $h$ as an $\epsilon$-isometry that satisfies the Eq.~\eqref{eq::isometrya}.
The isometric distance preservation with $\epsilon {=} 0$ implicates the locality preservation defined in Eq.~\eqref{eq::locality}. Moreover, with small $\epsilon$, the feature distance is related to the transformed distance by a linearly increasing function (Fig~\ref{Figure::theory}d). Hence, isometric distance preservation can be understood as a more specific, strict condition than locality preservation. 

As in locality-preserving transformation, an isometric distance-preserving transformation preserves the performance of original features at the cost of vulnerability to similarity attack. Moreover, high isometric distance preservation inevitably results in high information leakage. Our observations are presented formally as below:

\subsubsection{Performance Preservation of Isometric Distance-Preserving Transformation}
Consider the family of isometric distance-preserving transformations
\begin{equation}
\label{eq.hfamily}
\mathcal{H}_\epsilon = 
\begin{aligned}
\{h : h \text{ whose isometry degree} \geq \epsilon\}
\end{aligned}
\end{equation}
with small $\epsilon>0$.
Any transformation in $\mathcal{H}_\epsilon$ preserves the performance of original features as stated by the below proposition:

\begin{prop} 
For any inter-class index pair $(k, l)$, if $d(C_k, C_l) > \epsilon$, then $h(C_k)$ and $h(C_l)$ are linearly separated for any $h \in \mathcal{H}_\epsilon$.
\end{prop}
Here, the set distance metric $d(C_k, C_l)$ on the sets $C_k$ and $C_l$ is defined by
\begin{equation}
d(C_k, C_l) 
= \inf \{
d(\bm{x}_k, \bm{x}_l) : \bm{x}_k \in C_k, \bm{x}_l \in C_l
\}.
\end{equation}

\begin{proof}
Without loss of generality, assume there are only two classes, $C_1$, and $C_2$. 
Let $y_1 \in h(C_1)$ and $y_2 \in h(C_2)$ be arbitrary. Then, $y_1 = h(\bm{x}_1)$ and $y_2 = h(\bm{x}_2)$ for some $\bm{x}_1 \in C_1$ and $\bm{x}_2 \in C_2$. Now, due to $h \in \mathcal{H}_\epsilon$ as stated in (\ref{eq.hfamily}), we have
\begin{equation}
\lvert d( h(\bm{x}_1), h(\bm{x}_2) ) - d(\bm{x}_1, \bm{x}_2) \rvert < \epsilon,
\end{equation}
which implies
\begin{equation}
- \epsilon < d( h(\bm{x}_1), h(\bm{x}_2) ) - d(\bm{x}_1, \bm{x}_2).
\end{equation}
Thus,
\begin{equation}
d(y_1, y_2) = d(h(\bm{x}_1),h(\bm{x}_2)) > d(\bm{x}_1, \bm{x}_2) - \epsilon > \epsilon - \epsilon =0.
\end{equation}
Since $y_1$ and $y_2$ were arbitrary, we have $d(h(C_1), h(C_2)) \geq 0$, demonstrating the linear separability between $h(C_1)$ and $h(C_2)$.
\end{proof}

\subsubsection{Pre-image Attack on Isometric Distance-Preserving Transformation}
Unfortunately, just as in the case of locality-preservation, a high degree of isometric distance-preservation allows the attacker to have a high attack success rate for any cancelable biometric system. This is formalized in the following proposition:

\begin{prop}
Given any feature $x \in \mathcal{X}$, and let $\epsilon > 0$ be a constant for a given $h \in \mathcal{H}_\epsilon$. Assume a pre-image $\bm{x}_a$ satisfies that
\begin{equation}
d( h(\bm{x}), h(\bm{x}_a) ) < \tau
\end{equation}
for a fixed $\tau > 0$.
then 
\begin{equation}
d( f(\bm{x}), f(\bm{x}_a) ) < \tau_0 + 2 \epsilon
\end{equation}
for any $f \in \mathcal{H}_\epsilon$.
Here $\tau_0$ is a value satisfies $d(h(\bm{x}), h(\bm{x}_a)) < \tau_0 < \tau$ and independent of $\epsilon$. 
If $\epsilon < \tau - \tau_0$, moreover, then
\begin{equation}
d( f(\bm{x}), f(\bm{x}_a) ) < \tau.
\end{equation}
\end{prop}

\begin{proof}
Choose $\tau_0$ such that $0< \tau_0 < \tau$. Then, construct the pre-image $\bm{x}_a = \mathcal{A}(\bm{x}, h)$ by $\mathcal{A}$ such that
\begin{equation}
d(h(\bm{x}), h(\bm{x}_a)) < \tau_0 < \tau.
\end{equation}
The high performance of the cancelable transformation $h \in \mathcal{H}_\epsilon$ allows the attacker to find a pre-image that is similar to the original feature; namely,
\begin{equation}
\begin{split}
d(\bm{x}_a, x) & \leq \lvert d(\bm{x}_a, x) - d( h(\bm{x}_a), h(\bm{x}) ) \rvert + d( h(\bm{x}_a), h(\bm{x}) ) \\
& \leq \epsilon + \tau_0 . 
\end{split}
\end{equation}
This enables the attacker to attack any cancelable transformation by the pre-image. To see this, assume $f \in \mathcal{H}_\epsilon$ is another cancelable transformation function to be attacked. Then, it is followed by
\begin{equation}
\begin{split}
d(f(\bm{x}_a), f(\bm{x}))
&\leq
\lvert
d( f(\bm{x}_a), f(\bm{x}) ) - d(\bm{x}_a, x) 
\rvert
+d(\bm{x}_a, x) \\
&\leq \epsilon + (\epsilon + \tau_0).
\end{split}
\end{equation}
Since $\tau_0$ is independent of $\epsilon$, we may decrease $\epsilon$ such that $\epsilon < (\tau - \tau_0)/2$. Then, 
\begin{equation}
\begin{split}
d(f(\bm{x}_a), f(\bm{x})) & < \tau_0 + 2\epsilon \\
& < \tau_0 + (\tau - \tau_0) = \tau, 
\end{split}
\end{equation}
concluding the proof.
\end{proof}

\begin{coro}
For a feature vector $x \in \mathcal{X}$, and $\epsilon > 0$, let $h \in \mathcal{H}_\epsilon$. If
\begin{equation}
d(h(\bm{x}), h(\bm{x}_a)) = 0,
\end{equation}
then
\begin{equation}
d(f(\bm{x}), f(\bm{x}_a)) < 2\epsilon,
\end{equation}
for any $f \in \mathcal{H}_\epsilon$.
\end{coro}

In addition, isometric distance-preserving transformations involve the risk of information leakage; a high degree of isometry enables the attacker to retrieve the original feature. following nextproposition indicates this:

\begin{prop}
For $h \in \mathcal{H}_\epsilon$,
\begin{equation}
I(X, h(X)) \geq c + d \log 1/(2\epsilon)
\end{equation}
where $c=H(X)$ is a constant and $d$ is the feature dimension such that $X \in \mathcal{X} \subseteq \mathbb{R}^d$, $I(\cdot)$ denotes the mutual information and $H(X)$ denotes the entropy of the random variable $X$.
\end{prop}

\begin{proof}
Let $Y = h(X)$. Note that $I(X, Y) = H(X) - H(X|Y)$ where $H(X)$ is a constant. Thus, it suffices to find an upper bound of 
\begin{equation}
H(X|Y) = \int_y p(y) H(X | y) dy. 
\end{equation}
We claim that for each fixed $\bm{y}$
\begin{equation}
H(X|\bm{y}) \leq d \log 2\epsilon.
\end{equation}
To see this, let $\bm{x}_1, \bm{x}_2 \in h^{-1}(\bm{y})$. Then,
\begin{equation}
d(\bm{x}_1, \bm{x}_2) = | d(\bm{x}_1, \bm{x}_2) - d( h(\bm{x}_1), h(\bm{x}_2) ) | < \epsilon
\end{equation}
due to the degree $\epsilon$ of distance-preservation and $d(h(\bm{x}_1), h(\bm{x}_2)) = 0$. Therefore, as the distance between any two points in $h^{-1}(\bm{y})$ is bounded by $\epsilon$, the volume of $h^{-1}(\bm{y})$ is bounded by
\begin{equation}
\text{Vol}( h^{-1}(\bm{y}) ) \leq \text{Vol}( R ) \leq (2\epsilon)^d
\end{equation}
where $R$ is a $d$-dimensional rectangle that encloses $h^{-1}(\bm{y})$, and its volume is $(2\epsilon)^d$.
Therefore, by Jensen's inequality, the entropy $H(X|\bm{y})$ is bounded by the maximum entropy of a uniform distribution over $R$
\begin{equation}
H(X|\bm{y}) \leq \log (2 \epsilon)^d = d \log 2 \epsilon,
\end{equation}
concluding the proof.
\end{proof}

\subsection{Locality-Sensitive Hashing}

\medskip
\noindent
Now, we consider distance-preservation based on locality-sensitive hashing. 

\textbf{Definition.} A transformation $h$ is a \textbf{locality-sensitive hashing (LSH) with degree} $\epsilon{>}0$ if there are thresholds $\tau, r>0$ such that for any probability $p \geq 0$
\begin{equation}
\Prob \left[
d( h(X), h(X_{in}) ) \leq \tau
\right]
\geq (1-\epsilon) p 
\end{equation}
if $\Prob[ d( X, X_{in} ) < r ] \geq p$, and 
\begin{equation}
\Prob \left[
d( h(X), h(X_{out}) ) > \tau
\right]
\geq (1-\epsilon) p 
\end{equation}
if $\Prob[ d( X, X_{out} ) > r ] \geq p$.
From now on, we term such a transformation by $\epsilon$-\textbf{LSH}. With $\tau =0$, $\epsilon=1$, and $p=1$, we recover the original definition of LSH given in \cite{indyk1998approximate}. The degree $\epsilon$ indicates how well the transformation $h$ follows the LSH property. The thresholds $\tau$ and $r$, on the other hand, are data-dependent values and are pre-selected based on the distribution of features.

\medskip
\noindent
\textbf{Performance Preservation.}
By definition, a $\epsilon$-LSH transformation preserves the recognition of original features. Precisely, if $\epsilon$ is smaller, then the LSH transformation better holds the performance.

\medskip
\noindent
\textbf{Pre-image Attack.}
As in other distance-preservation properties, the LSH property also makes the cancelable system vulnerable to the pre-image attack. The following proposition verifies this:

\begin{prop}
Assume a high attack success rate of the pre-image $X_a$ attack
\begin{equation}
\Prob \left[
d( h(X), h(X_a) ) \leq \tau 
\right]
\geq p
\end{equation}
on the initial transformation $h$.
Then, for any other $\delta$-LSH transformation $f$, a high attack success rate is posed even after template renewal by $f$:
\begin{equation}
\Prob \left[
d( f(X), f(X_a) ) \leq \tau 
\right]
\geq \frac{p - \epsilon}{1 - \epsilon}(1 - \delta).
\end{equation}
\end{prop}
\begin{proof}
To prove by contradiction, assume
\begin{equation}
\Prob \left[
d( f(X), f(X_a) ) \leq \tau 
\right]
< q
\end{equation}
where $q = (p - \epsilon)(1 - \delta)/(1 - \epsilon)$.
Then,
\begin{equation}
\Prob \left[
d( X, X_a ) < r
\right]
< q/(1 - \delta)
\end{equation}
by the definition of $\delta$-LSH transformation $f$. The above inequality is equivalent to
\begin{equation}
\Prob \left[
d( X, X_a ) > r
\right]
=
\Prob \left[
d( X, X_a ) \geq r
\right]
\geq 1 - \frac{q}{1-\delta}
\end{equation}
where the equality is due to the continuum of the space $\mathbb{R}$. Then, by the property of LSH of $f$, we have
\begin{equation}
\Prob \left[
d( h(X), h(X_a) ) > \tau
\right]
\geq (1-\epsilon)\left(
1 - \frac{q}{1-\delta}
\right),
\end{equation}
which is equivalent to 
\begin{equation}
\Prob \left[
d( h(X), h(X_a) ) \leq \tau
\right]
< 1 - (1-\epsilon)\left(
1 - \frac{q}{1-\delta}
\right).
\end{equation}
Now, setting $q$ as given, we obtain the above probability less than $p$, a contradiction. This completes the proof.
\end{proof}

The proposition states that if an attack algorithm can generate a pre-image $X_a$ whose template is considered to be in the same class of $X$ with the probability $p$, then even after template renewal by another LSH $f$, the replaced template of the same pre-image is regarded to be in the same class of $X$ with the probability $(p-\epsilon)(1-\delta)/(1-\epsilon)$. Thus, template renewal would not be much helpful for protection against similarity attacks by pre-image. 

\noindent
\textbf{Information Leakage.} Unlike other distance-preserving properties, information leakage $I(X, h(X))$ is not linked to the LSH property. Particularly, there are LSH transformations with low information leakage (Fig.~\ref{Figure::theory}b) and ones with high information leakage (Fig.~\ref{Figure::theory}cd). Particularly, the leakage can depend on the specific thresholds of the LSH transformation other than the degree of LSH. 

\section{Visualisation of the correlation between original space distances and transform space distances}
\begin{figure*}[t!]
\centering
\subfloat[BioHashing]{\includegraphics[width=0.32\linewidth]{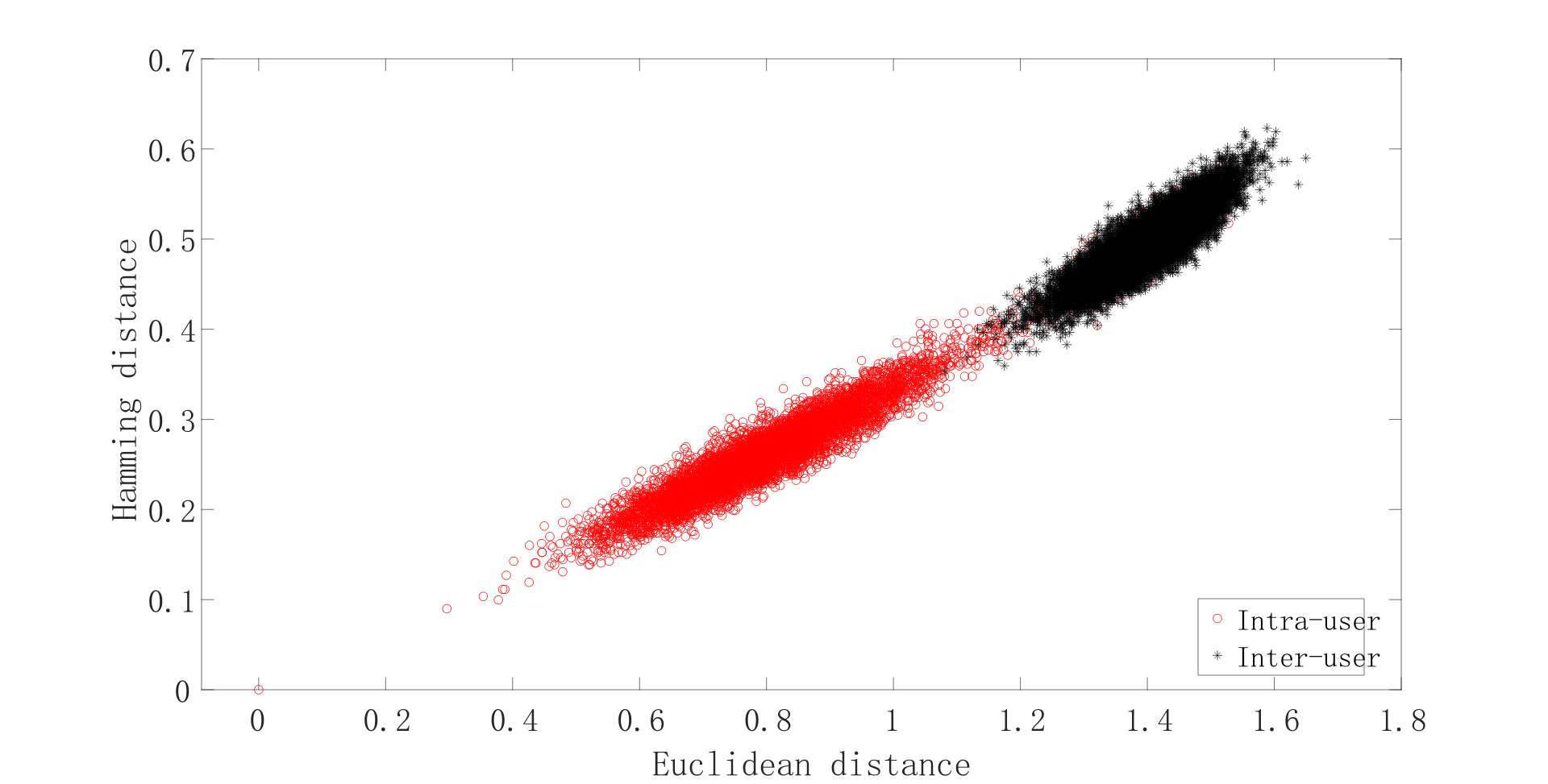}}
\subfloat[IoM]{\includegraphics[width=0.32\linewidth]{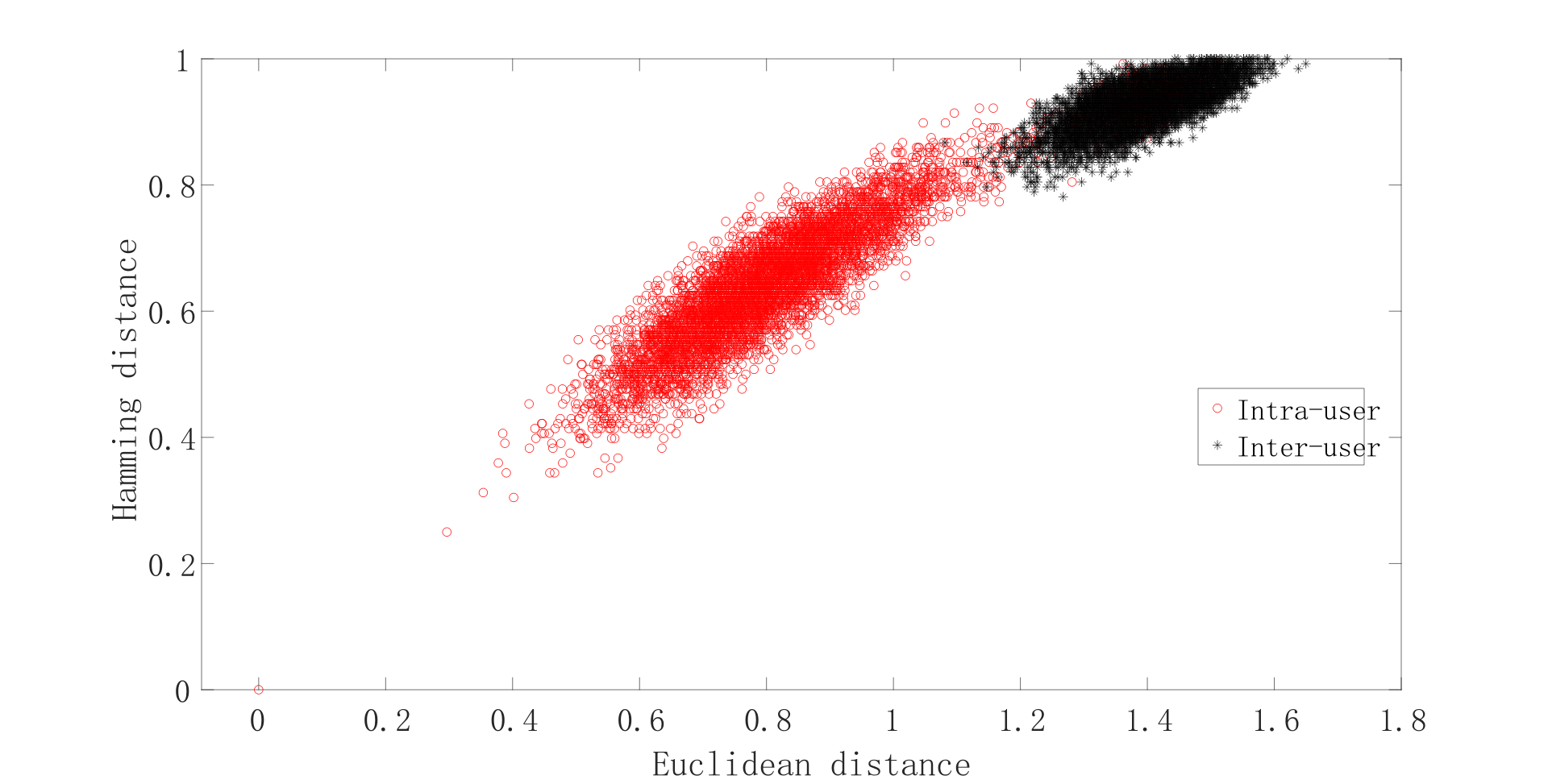}}
\subfloat[2PMCC]{\includegraphics[width=0.32\linewidth]{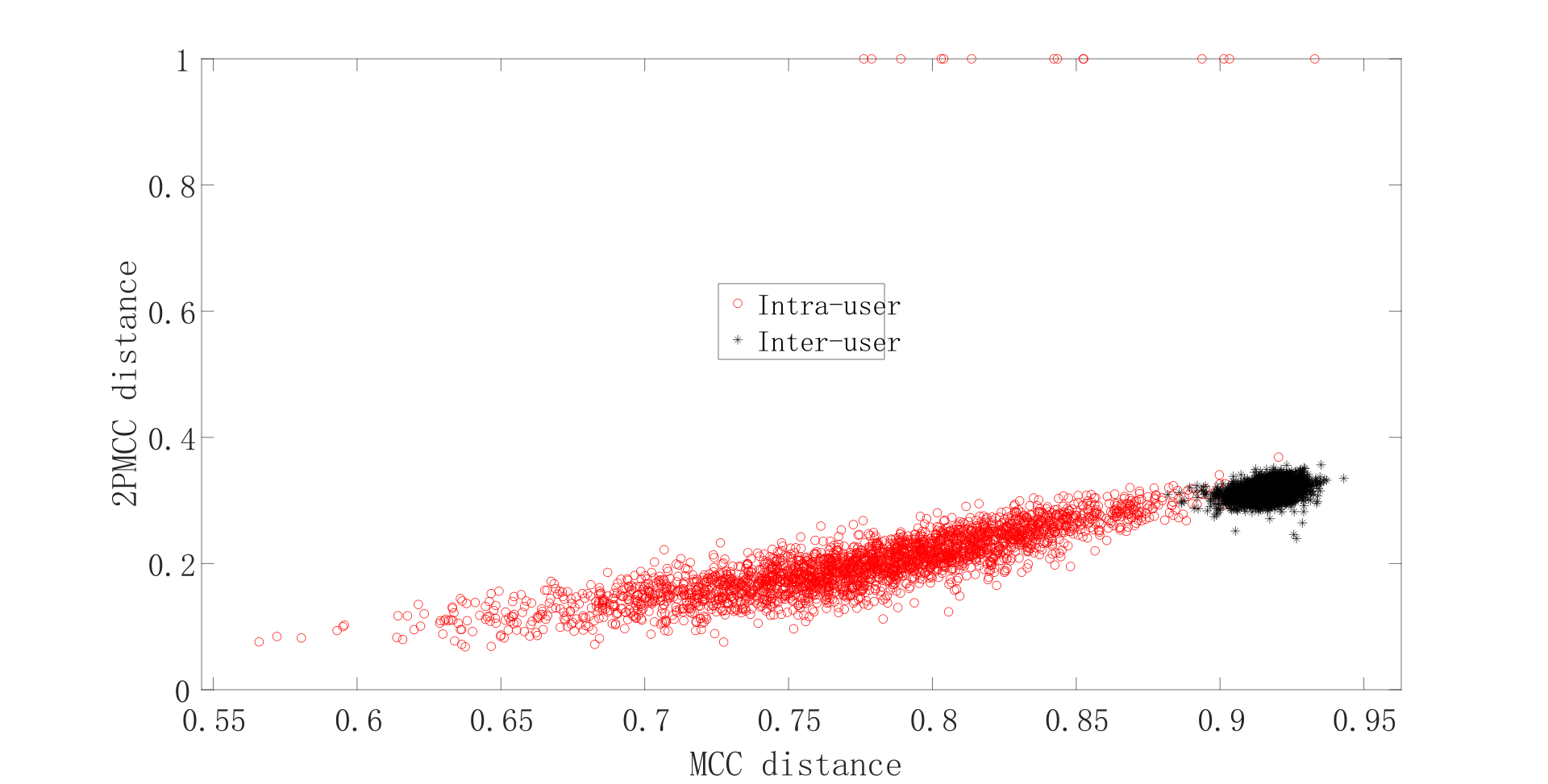}}
\\
\subfloat[NMDSH ($\alpha=0.1$)]{\includegraphics[width=0.32\linewidth]{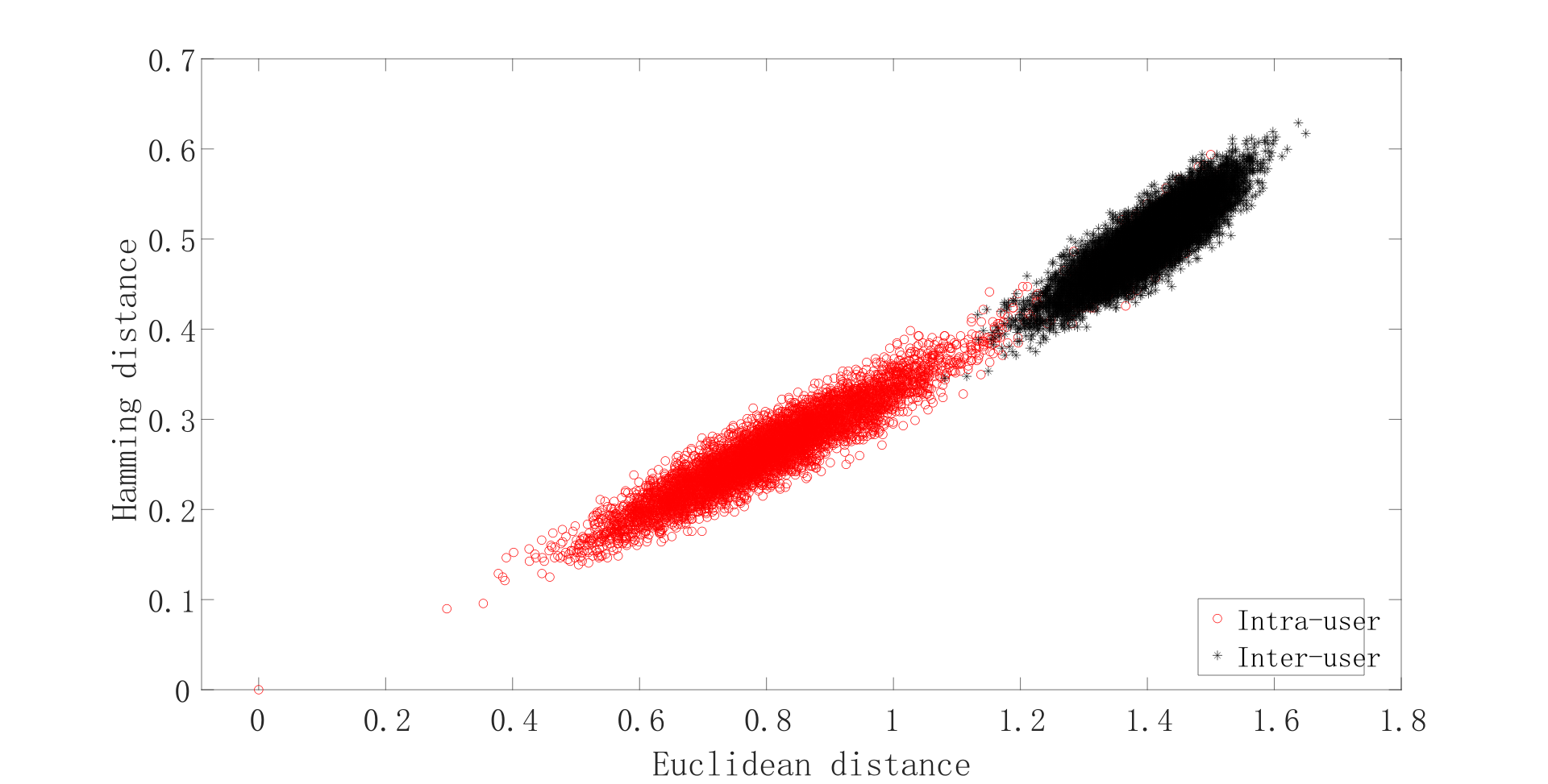}}
\subfloat[NMDSH ($\alpha=0.5$)]{\includegraphics[width=0.32\linewidth]{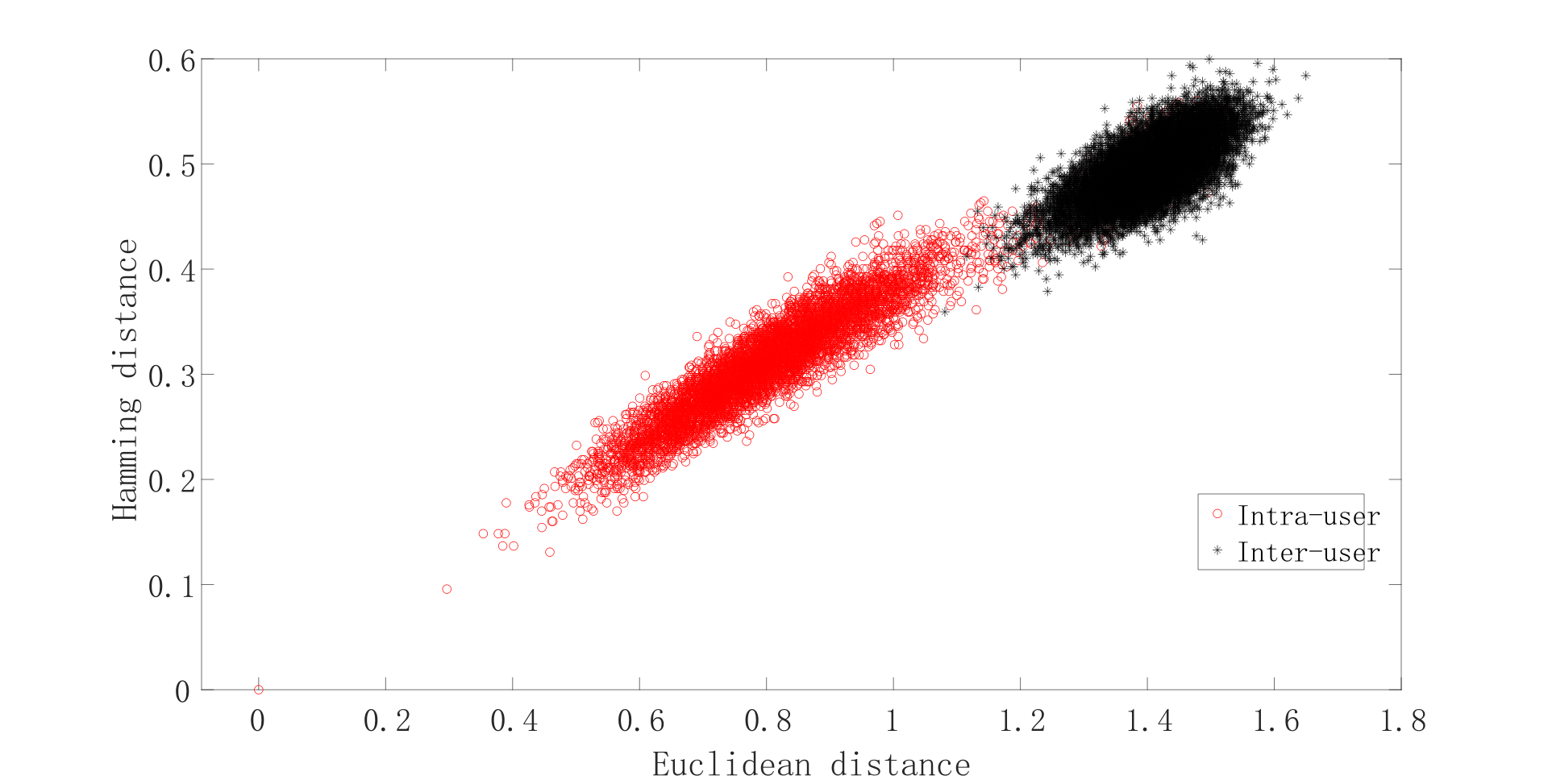}}
\subfloat[NMDSH ($\alpha=0.9$)]{\includegraphics[width=0.32\linewidth]{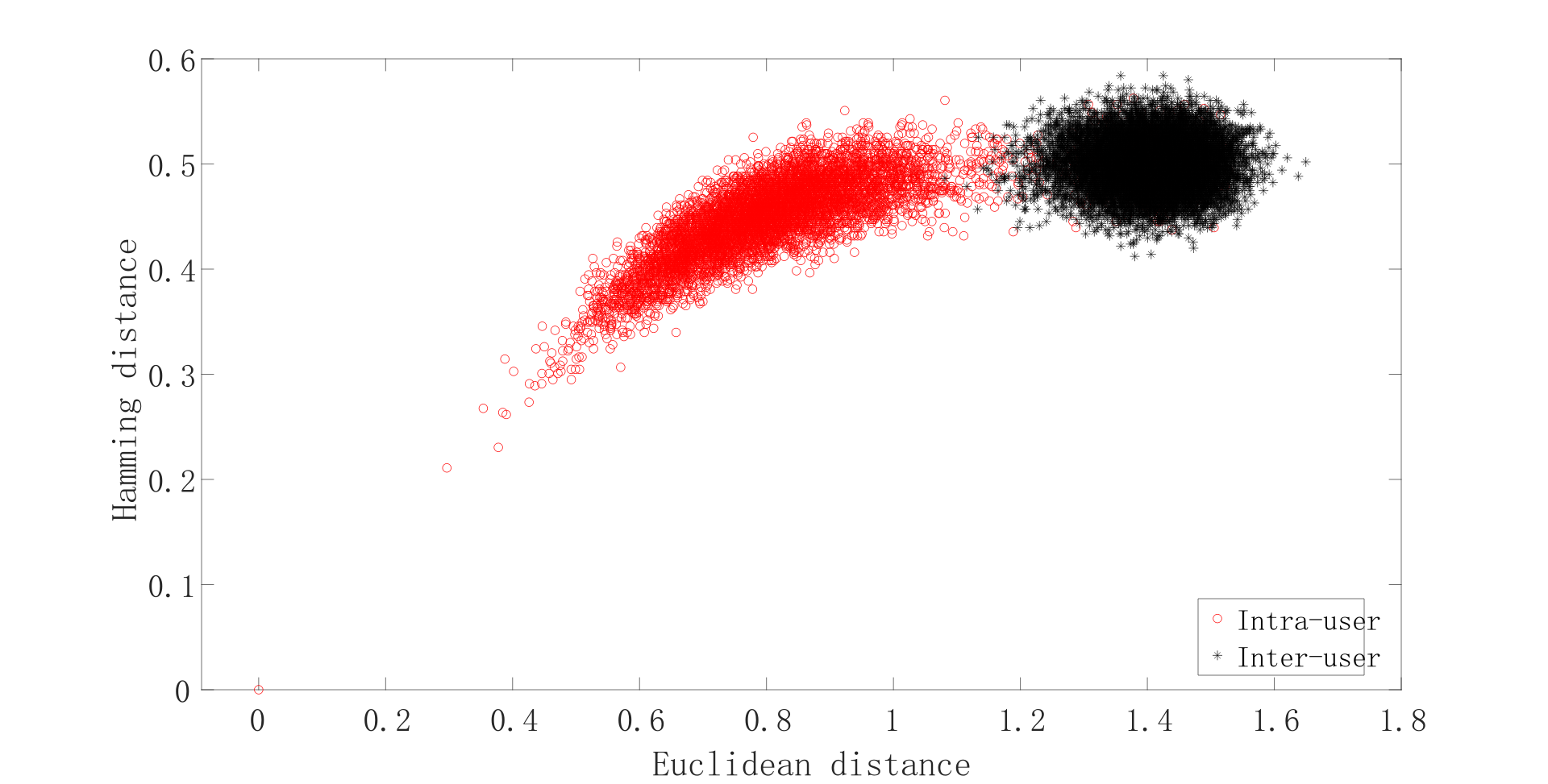}}
\\
\subfloat[IFO]{\includegraphics[width=0.32\linewidth]{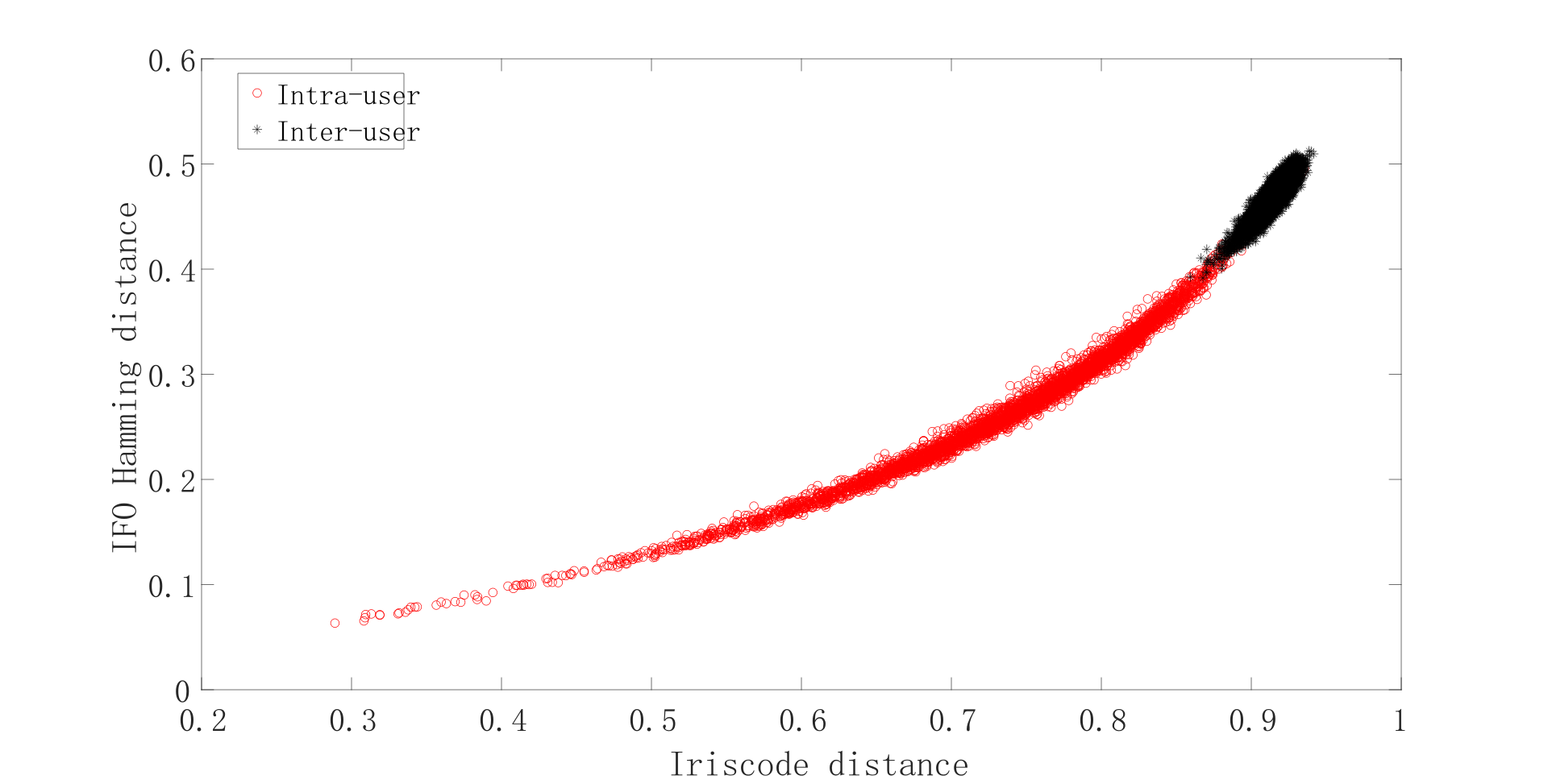}}
\subfloat[Bloom Filter]{\includegraphics[width=0.32\linewidth]{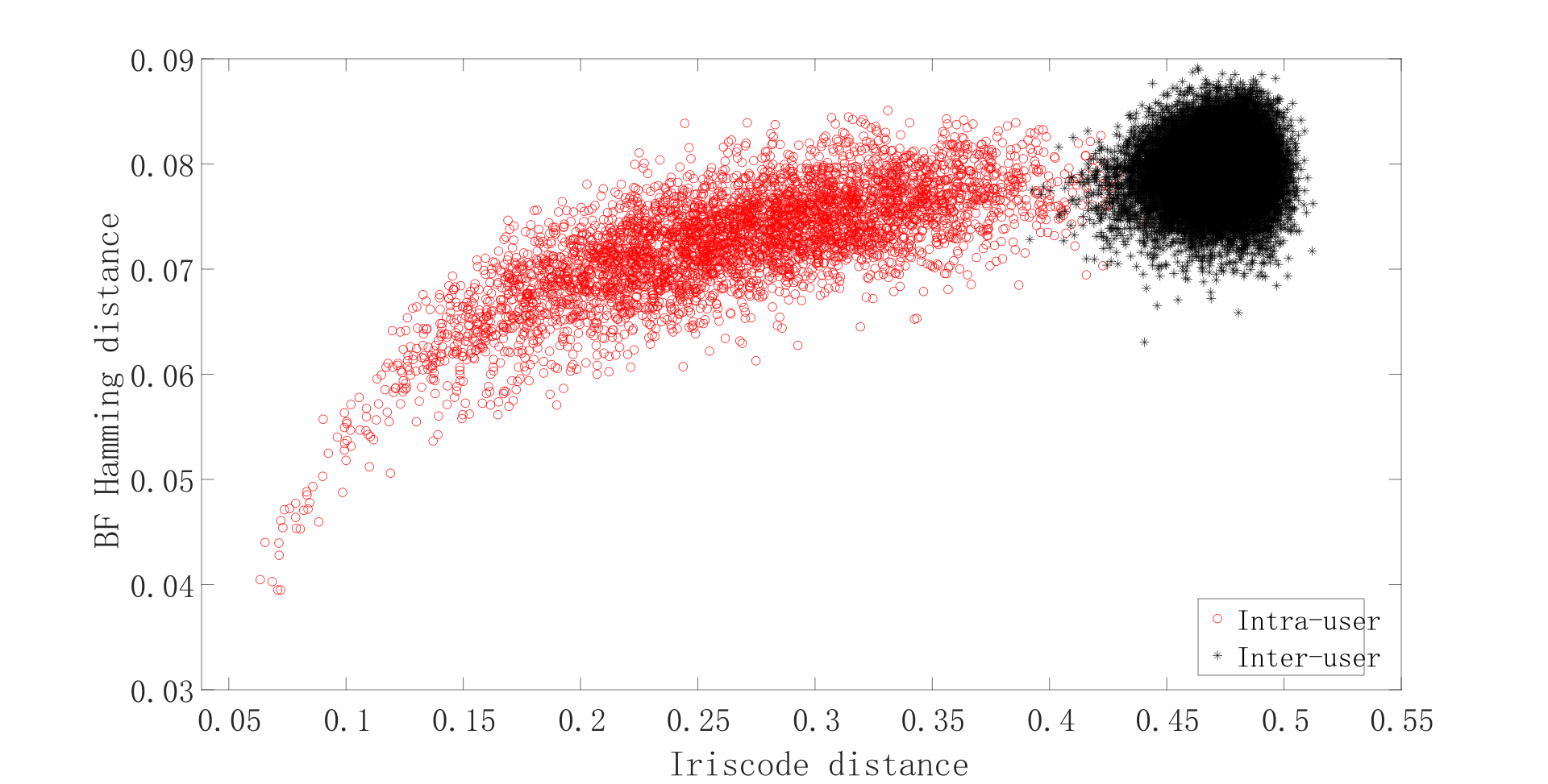}}
\caption{Correlation between original feature space distances and transform space distances ( $l=512$ for BioHashing, IoM, NMDSH). \label{Figure::distanceco}}
\end{figure*}

Fig. \ref{Figure::distanceco} shows the correlation between original Euclidean distances and Hamming distances in BioHashing, IoM, NMDSH ($\alpha=\{0.1,0.5,0.9\}$) on face features. BioHashing, IoM are relatively linear correlated, while NMDSH is non-linear correlated when $\alpha=0.9$.

\section{Quantifying information leakage via the Blahut–Arimoto algorithm\label{appx.quantify}}
Similar to the capacity computation of a communication channel, $\lambda_{max}$ can also be solved using existing optimization algorithms. The Blahut–Arimoto algorithm \cite{blahut1972computation,arimoto1972algorithm} is often applied to compute the information-theoretic capacity of a channel explicitly. This corresponds to the amount of information leaked from $S$ to $T$. The Blahut–Arimoto algorithm aims to find the optimal solution to the convex optimization problem by an iterative process over $\bm{q}$. The initial set of $\bm{q}^0(i)$ for all $i$ is chosen first. A set of conditional probabilities $\bm{\Phi}^{\ell}(i / j)$ with input $s_i$ and output $t_j$ is then computed by
\begin{equation}
\bm{\Phi}^{\ell}(i | j)=\frac{\bm{q}^{\ell}(i) \bm{D}(j | i)}{\sum_{k} \bm{q}^{\ell}(k) \bm{D}(j | k)},\forall(i, j),\label{eq.ba2}
\end{equation}
where $i=1,2,...,I$, $j=1,2,...,J$, and $\ell$ is the iteration index. Then $\bm{q}^{\ell+1}(i)$ is recursively computed by
\begin{equation}
\bm{q}^{(\ell+1)}(i)= \frac{\exp \sum_{i} \bm{D}(j | i) \ln \bm{\Phi}^{\ell}(i | j)}{\sum_k\exp \sum_{i} \bm{D}(j | i) \ln \bm{\Phi}^{\ell}(i | j)}. \label{eq.ba1}
\end{equation}
Then, the leakage $I(\mathcal{S};\mathcal{T})$ can be estimated by
$J(\bm{q}^{\ell+1},\bm{\Phi}^{\ell})$:
\begin{equation}
J(\bm{q}^{\ell+1}, \bm{\Phi}^\ell)=\sum_{i} \sum_{j} \bm{D}(j | i) \bm{q}^{\ell+1}(i) \log \frac{\bm{\Phi}^{\ell}(i | j)}{\bm{q}^{\ell+1}(i)},
\end{equation}
The amount of information leakage at the next iteration is
\begin{equation}
\lambda(\ell+1)= J(\bm{q}^{l+1}, \bm{\Phi}^{\ell}),\label{eq.ba3}
\end{equation}
As shown by the code in Algorithm~2, the process ends when $|\lambda(\ell+1)-\lambda(\ell)|<\delta$.

\begin{algorithm}[t!]
\SetAlgoLined
\begin{flushleft}
\textbf{INPUT:} $\bm{D}(j|i)$ of a given cancelable biometric system\\
\textbf{OUTPUT:} Max leakage $\lambda_{max}$
\end{flushleft}
Initiate $\bm{q}^0$ with uniform distribution, iteration index $\ell=0$\, compute $\lambda(0)$;
Define precision parameter $\delta$\;
\Repeat{$|\lambda(\ell+1)-\lambda(\ell)|<\delta$}{
Compute $\bm{q}^{(\ell+1)}$ according to (\ref{eq.ba1})\;
Compute $\bm{\Phi}^{\ell}(i | j)$ according to (\ref{eq.ba2})\;
Compute $\lambda(\ell+1)$ according to (\ref{eq.ba3})\;
$\ell=\ell+1$\;
}
$\lambda_{max}=\lambda(\ell+1)$\;

\caption{Quantification of the maximum leakage, based on the Blahut–Arimoto algorithm \label{algo.ba}}
\end{algorithm}

It is worth noting that the Blahut–Arimoto algorithm can only be applied to discrete data. Hence, quantization converts a continuous matching score into a discrete score by dividing the scores into small bins with widths $e=0.01$.

Quantifying the information leakage for a given cancelable biometric scheme can be achieved in four steps: 

\begin{enumerate}
\item Collect enough biometric samples $\{\bm{x_1},...,\bm{x_n}\}$, and generate the template $\{\bm{y_1},...,\bm{y_n}\}$ by the given transformation function.
\item Compute the feature space distance and transform space distance pairs $\{(d_{ij},ds_{ij})\ | 1 \leq i,j \leq n\}$, where $d_{ij}=\lVert \bm{x_i}-\bm{x_j} \lVert$ and $ds_{ij}=\lVert \bm{y_i}-\bm{y_j} \lVert$.
\item Denote the minimum and maximum values of $d_{ij}$ by $d_{min}$ and $d_{max}$, respectively, and denote the minimum and maximum values of $ds_{ij}$ by $ds_{min}$ and $ds_{max}$, respectively. Given a point $s$ between $d_{min}$ and $d_{max}$, and a point $t$ between $ds_{min}$ and $ds_{max}$, the transition probability can be estimated as
$\Prob(T=t|S=s) = \frac{\#~of~ pairs~ with~ d_{ij}=s~ and~ ds_{ij}=t}{\#~ of~ pairs~ with~ d_{ij}=s}$, defining $\bm{D}(j|i)$. As infinitely many values exist between these two intervals, they are quantized to obtain a finite number of discrete values.
\item The maximum leakage $\lambda_{\max }$ is computed by algorithm \ref{algo.ba} from the computed $C$ matrix.
\end{enumerate}

\begin{figure*}[t!]
\begin{center}
\imgStubPDFpage{0.40}{page=30,width=1\linewidth,trim=11cm 5cm 12cm 5cm, clip}{pdf/diagram.pdf}{(a) Original} 
\imgStubPDFpage{0.40}{page=31,width=1\linewidth,trim=11cm 5cm 12cm 5cm,clip}{pdf/diagram.pdf}{(b) Reconstructed}
\caption{Reconstructed fingerprint minutiae points of the 100-th user in FVC2002 DB2-A. The approximated minutiae points do not necessarily compare with the original ones, but the generated 2PMCC templates can be matched together. \label{Figure::minutiae}}
\end{center}
\end{figure*}

\begin{figure*}[t!]
\centering
\includegraphics[page=34,width=0.98\linewidth,trim=0cm 11cm 0cm 3cm, clip]{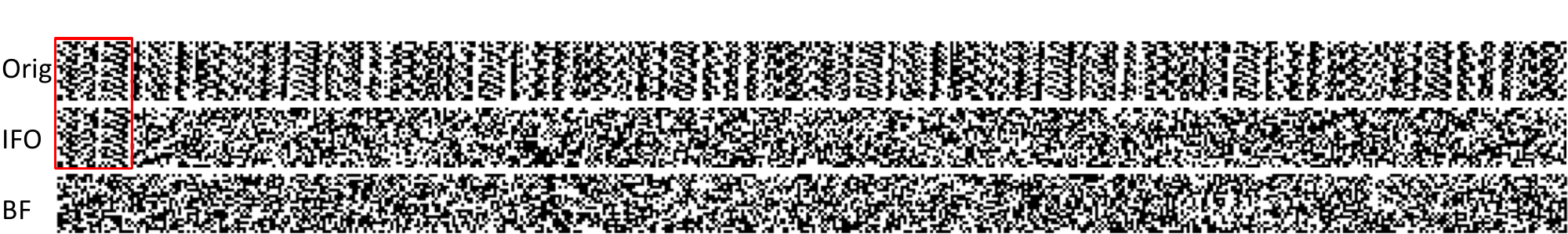} 
\caption{Comparison between the original Iriscode (top), the reconstructed IrisCode from the IFO (middle), and the improved Bloom Filter hashing (bottom). The parametrization for the IFO is $\tau=50, m=800, K=300, P=3$. The parametrization for the improved Bloom Filter is $\omega=10$; block size is $2^{4}$. The red box shows how the IrisCode reconstructed from the IFO partially matches the original IrisCode. \label{Figure::ifobfreconstructiriscode}}
\end{figure*}

\section{Pre-image visualisation}
As the pre-image of fingerprint and iris are minutiae points and iris codes, we visualize the generated pre-images for these two modalities in our experiments. 
In Fig.~\ref{Figure::minutiae}, the original fingerprint and its pre-image recovered from the 2PMCC template are displayed. It can be seen that the pre-image does not need to be similar compared with the original minutiae points to generate a 2PMCC template close to the original one. 
Fig. \ref{Figure::ifobfreconstructiriscode} shows an example of the Iriscode pre-image recovered from the Bloom filtering template. As can be noticed, the reconstructed pre-image is very different from the original one.

\begin{figure*}[t!]
\begin{center}
\imgStubPDFpage{0.32}{page=1,width=1\linewidth,trim=1cm 23.2cm 6cm 1cm,clip}{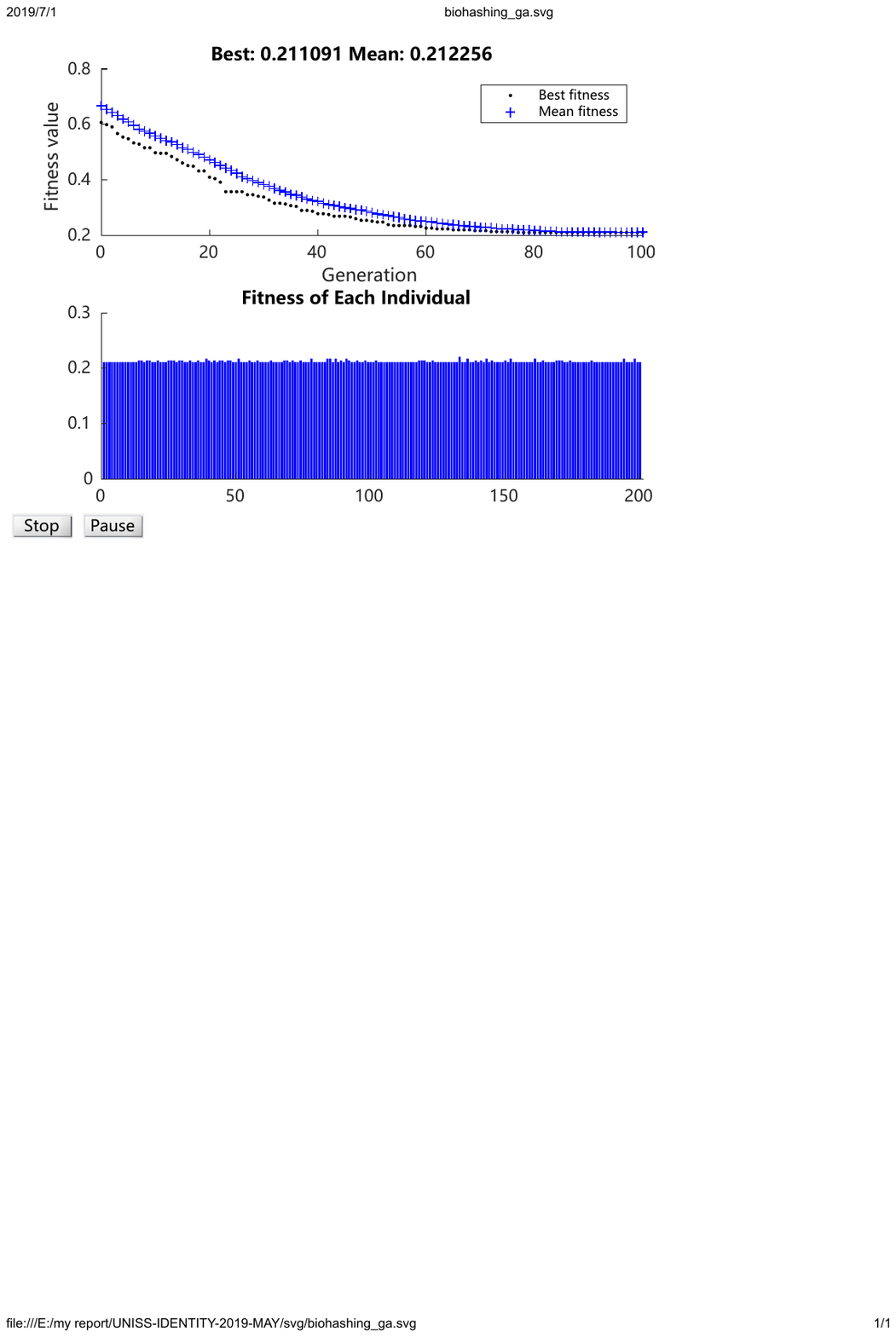}{(a) BioHashing} 
\imgStubPDFpage{0.32}{page=1,width=1\linewidth,trim=1cm 23.2cm 6cm 1cm,clip}{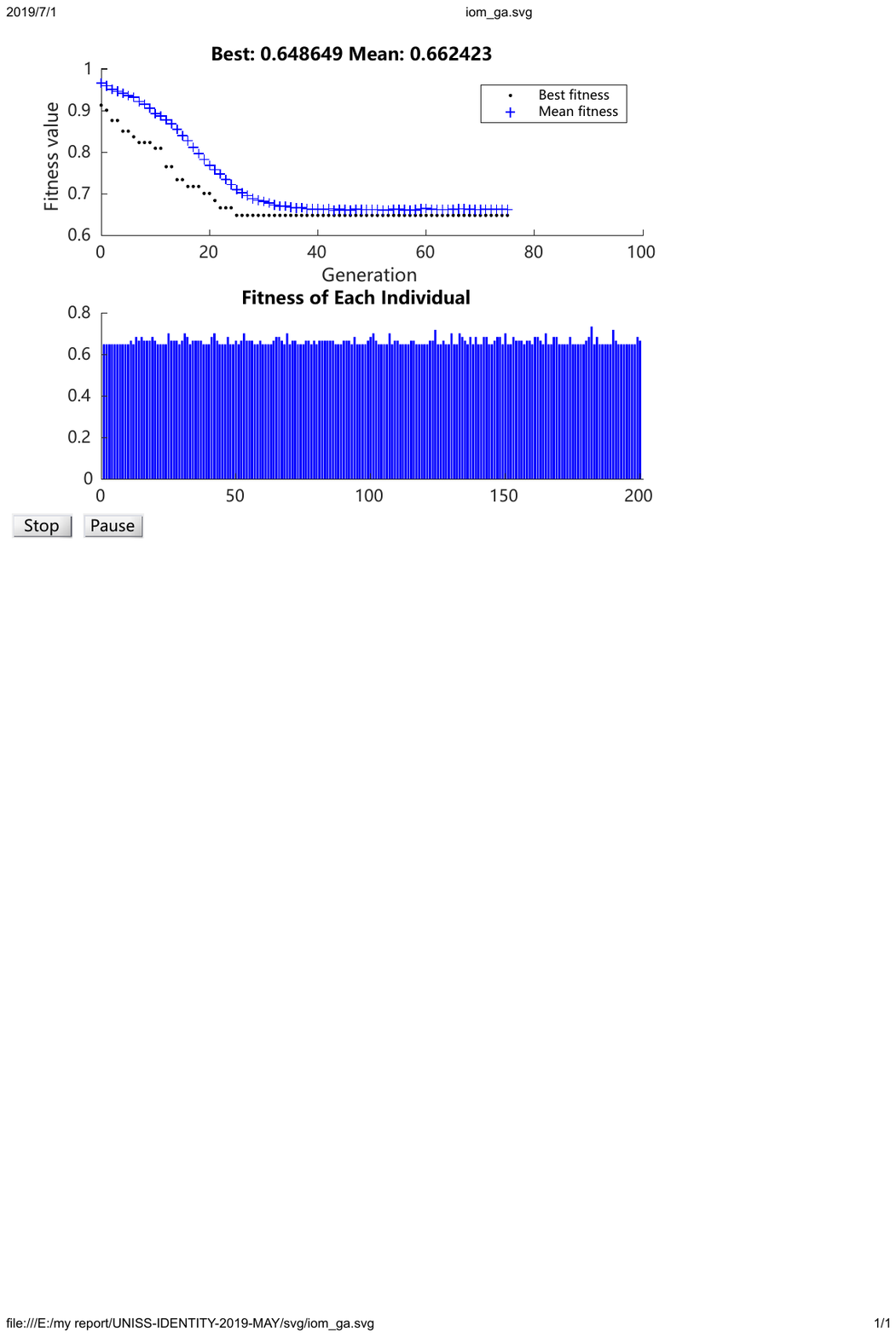}{(b) IoM Hashing}
\imgStubPDFpage{0.32}{page=1,width=1\linewidth,trim=1cm 23.2cm 6cm 1cm,clip}{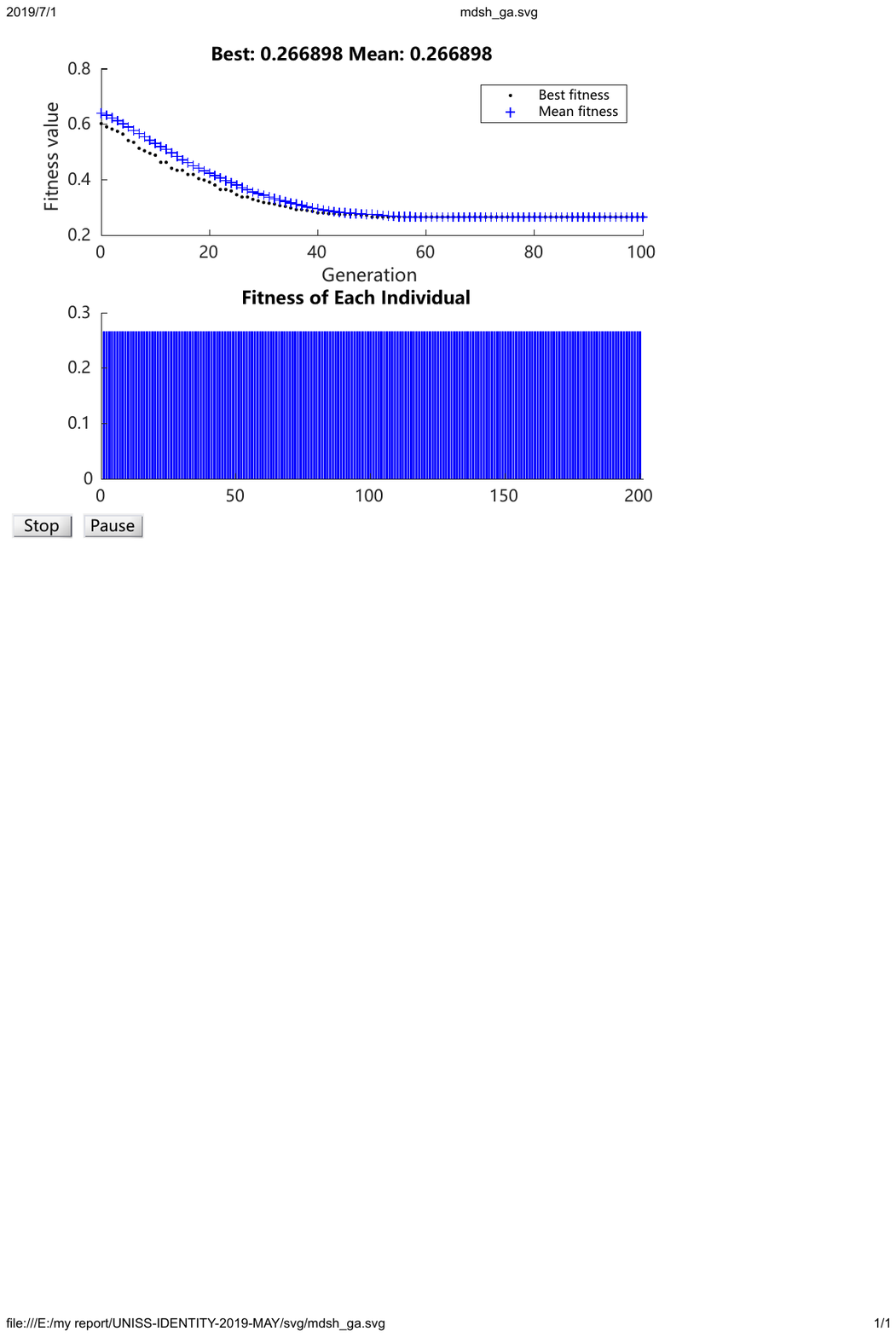}{(c) NMDSH}
\imgStubPDFpage{0.32}{page=1,width=1\linewidth,trim=1cm 23.2cm 6cm 1cm,clip}{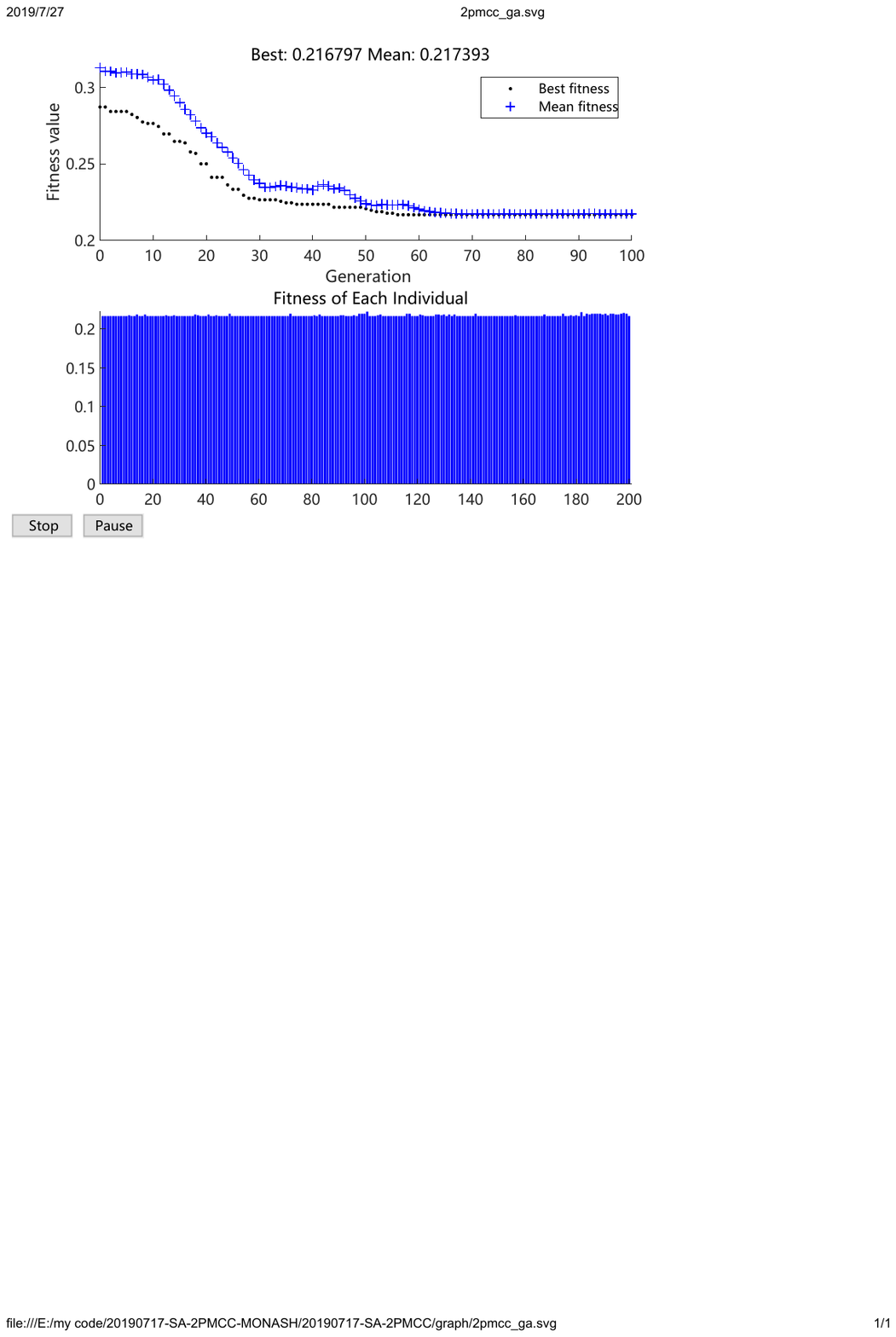}{(d) 2PMCC}
\imgStubPDFpage{0.32}{page=36,width=1\linewidth,trim=9cm 9.5cm 9cm 4cm,clip}{pdf/diagram.pdf}{(e) IFO}
\imgStubPDFpage{0.32}{page=1,width=1\linewidth,trim=1cm 23.2cm 6cm 1cm,clip}{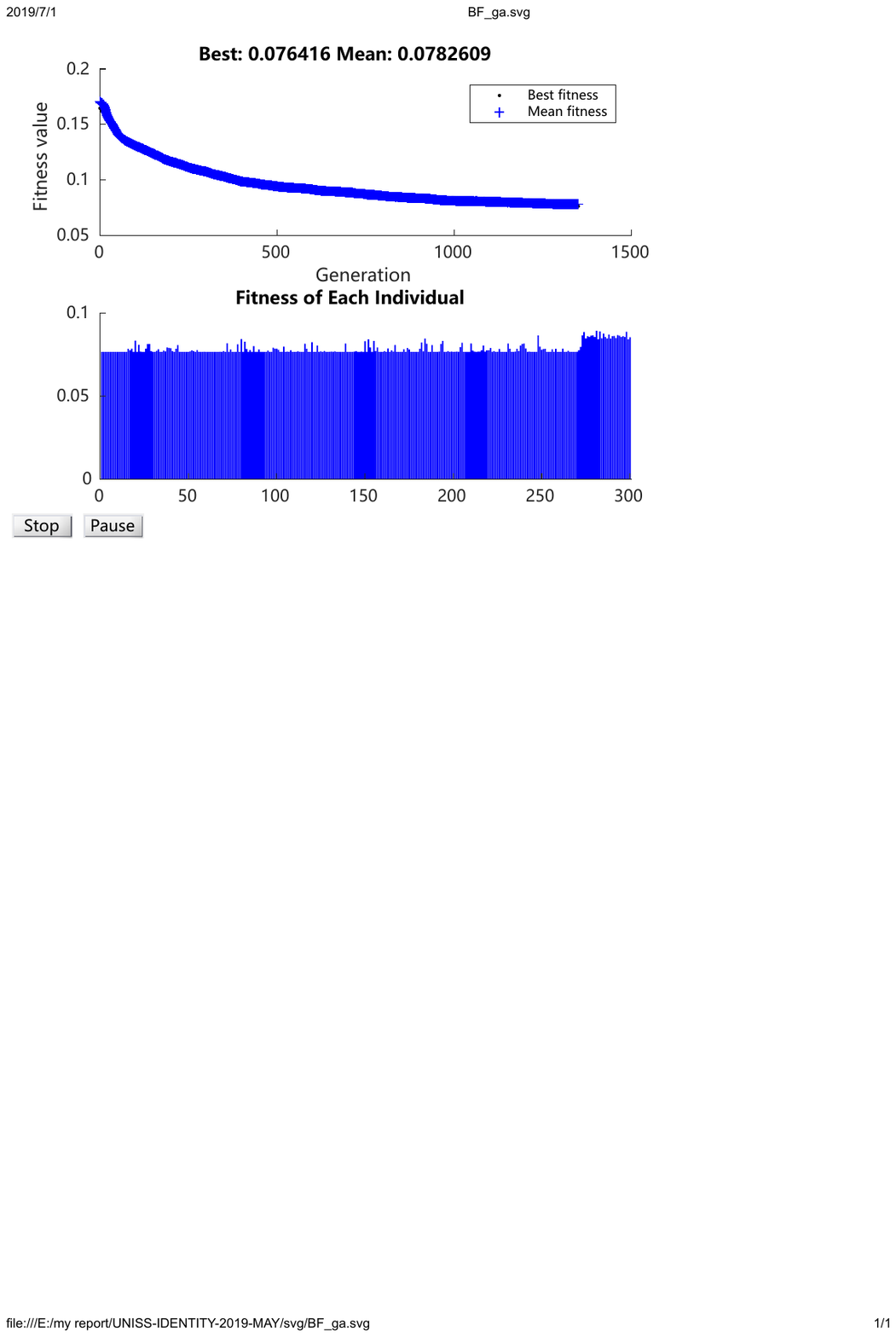}{(f) Bloom filter}
\caption{The objective value against generations. The attack on BioHahing, IoM hashing, NMDSH, and 2PMCC can converge at around 100 generations with a population size of 200. \label{Figure::time1}}
\end{center}
\end{figure*}

\begin{table*}
\centering
\caption{Time cost of the attack for different CB schemes. \label{table::time1}}
\begin{tabular}{ccccc} 
\toprule
CB & Max generations & $l$ & CB execution time & Overall attack time \\ 
\midrule
BioHashing & 100 & 500 & 0.000024 s & 17 s \\
IoM hashing & 100 & 500 & 0.00018 s & 14 s \\
NMDSH & 100 & 256 & 0.041 s & 13 mins \\
2PMCC & 100 & 64 & 0.0025 s & 21 mins \\
IFO & 300 & 800 & 0.0062 s & 9 mins \\
BloomFilter & 1500 & $2^6*2^8$ & 0.0026 s & 33 mins \\
\bottomrule
\end{tabular}
\end{table*}
\section{Time cost of the attack\label{sec.timecostappendix}}
The time cost for the attack depends mainly on the execution time of the cancelable biometric algorithm, the size of the feature, the population size, and the maximum number of generations of the GA. Figure~\ref{Figure::time1} indicates that BioHashing, NMDSH, IoM hashing, and 2PMCC can converge within 100 generations, while Bloom filter can converge around 1000 generations, and IFO can converge within 250 generations. Table~\ref{table::time1} tabulates the specific details of the GA and overall time cost for one template. The result suggests that the time mainly depends on the execution time of the hashing process. However, it is safe to conclude that the pre-image attack is time efficient in real life.

\vfill

\end{document}